\documentclass[final,5p,times,twocolumn]{elsarticle} 

\setcitestyle{authoryear,round,semicolon}

\def\hideInternalStuff{0} 

 \PassOptionsToPackage{hyphens}{url}
 \usepackage{hyperref}
\usepackage{bm}
\usepackage{amsmath,amssymb,amsfonts}
\usepackage{algorithmic}
\usepackage{graphicx}
\graphicspath{{./figures/}
}
\usepackage{textcomp}
\usepackage{}
\usepackage{url}
\usepackage{graphbox}
\usepackage{xcolor}
\usepackage{subcaption}
\usepackage{mathastext}

\newcommand{\todo}[1]{\color{blue}[TODO: {#1}]\normalcolor}
\definecolor{noteToSelfColor}{rgb}{0.6,0.6,0.6}
\newcommand{\noteToSelf}[1]{\color{noteToSelfColor}[{#1}]\normalcolor}
\definecolor{todoDoneColor}{rgb}{0.0,0.6,0.0}
\newcommand{\todoDone}[1]{\color{todoDoneColor}[Done: {#1}]\normalcolor}
\definecolor{unresolvedColor}{rgb}{0.6,0.0,0.0}

\definecolor{discussedWithChiaraColor}{rgb}{1.0,0.65,0.0}
\newcommand{\discussedWithChiara}[1]{\color{discussedWithChiaraColor}[{#1}]\normalcolor}

\definecolor{questionColor}{rgb}{1.0,1.0,0.0}
\newcommand{\quest}[1]{%
  \colorbox{questionColor}{%
    \begin{minipage}[t]{\linewidth}%
        {QUESTION: #1}%
     \end{minipage}%
  }%
}

\definecolor{answerColor}{rgb}{0.0,1.0,0.0}
\newcommand{\answer}[1]{%
  \colorbox{answerColor}{%
    \begin{minipage}[t]{\linewidth}%
        {ANSWER: #1}%
     \end{minipage}%
  }%
}

\definecolor{textSnippetColor}{rgb}{0.6,0.6,0.6}
\newcommand{\textSnippet}[1]{\color{textSnippetColor}[\textit{#1}]\normalcolor}
\definecolor{revisionColor}{rgb}{1.0, 0.57, 0.0}

\newlength{\myWidth}
\newlength{\mySpace}

\usepackage{natbib}

\def\hideInternalStuff{0} 

\def\hideNoteToSelf{1} 
\if\hideInternalStuff1
  \renewcommand{\todo}[1]{}
  \renewcommand{\todoDone}[1]{}
  \renewcommand{\noteToSelf}[1]{}
  \renewcommand{\quest}[1]{}
  \renewcommand{\answer}[1]{}
  \renewcommand{\textSnippet}[1]{}
  
  \renewcommand{\discussedWithChiara}[1]{}
\fi

\if\hideNoteToSelf1
  \renewcommand{\noteToSelf}[1]{}
\fi

\usepackage{makecell}
\usepackage{adjustbox}

\usepackage{xr}         

\newcommand{\suppfigref}[1]{Supplementary Fig.~\ref*{#1}}
\newcommand{\suppsecref}[1]{Supplementary Sec.~\ref*{#1}}

\journal{Human Brain Mapping}

\begin{document}

\begin{frontmatter}

\title{A Contrast-Agnostic Method for Ultra-High Resolution Claustrum Segmentation}

\author[1,2]{Chiara Mauri\corref{cor1}}
  \ead{cmauri@mgh.harvard.edu}
\author[1]{Ryan Fritz}
\author[1]{Jocelyn Mora} 
\author[3,4]{Benjamin Billot}
\author[1,2,3,5]{Juan Eugenio Iglesias}
\author[6,7]{Koen Van Leemput}
\author[1,2]{Jean Augustinack} 
\author[1,2]{Douglas N Greve}

\cortext[cor1]{Corresponding author}

\address[1]{Department of Radiology, Athinoula A. Martinos Center for Biomedical Imaging, Massachusetts General Hospital, Charlestown, MA,
USA}     
\address[2]{Department of Radiology, Harvard Medical School, Boston, MA 
USA}
\address[3]{MIT Computer Science \& Artificial Intelligence Laboratory, Cambridge, MA
USA}  
\address[4]{Epione team, Inria, Sophia Antipolis, France}
\address[5]{UCL Centre for Medical Image Computing, London, United Kingdom
}  
\address[6]{Department of Neuroscience and Biomedical Engineering, Aalto University, Espoo, Finland}
\address[7]{Department of Computer Science, Aalto University, Espoo, Finland}

\begin{abstract}
The claustrum is a band-like gray matter structure located between putamen and insula whose exact functions are still actively researched. Its sheet-like structure makes it barely visible in \textit{in vivo} Magnetic Resonance Imaging (MRI) scans at typical resolutions, and neuroimaging tools for its study, including methods for automatic segmentation, are currently very limited. In this paper, we propose a contrast- and resolution-agnostic method for claustrum segmentation at ultra-high resolution (0.35 mm isotropic); the method is based on the SynthSeg segmentation framework \citep{billot2023synthseg}, which leverages the use of synthetic training intensity images to achieve excellent generalization. In particular, SynthSeg requires only label maps to be trained, since corresponding intensity images are synthesized on the fly with random contrast and resolution. We trained a deep learning network for automatic claustrum segmentation, using claustrum manual labels obtained from 18 ultra-high resolution MRI scans (mostly \textit{ex vivo}). We demonstrated the method to work on these 18 high resolution cases (Dice score = 0.632, mean surface distance = 0.458 mm, and volumetric similarity = 0.867 using 6-fold Cross Validation (CV)), and also on \textit{in vivo} T1-weighted MRI scans at typical resolutions ($\approx$1 mm isotropic). We also demonstrated that the method is robust in a test-retest setting and when applied to multimodal imaging (T2-weighted, Proton Density and quantitative T1 scans). To the best of our knowledge this is the first accurate method for automatic ultra-high resolution claustrum segmentation, which is robust against changes in contrast and resolution. The method is released at \url{https://github.com/chiara-mauri/claustrum_segmentation} and as part of the neuroimaging package Freesurfer \citep{fischl2012freesurfer}.
\end{abstract}


\end{frontmatter}
\section{Introduction}

\noindent
The claustrum is a thin, sheet-like subcortical gray matter structure situated between insular cortex and putamen, enclosed by the white matter of the capsule. It includes a thin dorsal section and a thicker ventral portion characterized by fragmented components, often referred to as ``fingers" or ``puddles" \citep{mathur2014claustrum,coates2024high,remedios2010unimodal}.
Its precise boundaries are not yet fully characterized \citep{mathur2014claustrum}; while earlier studies estimated a volume of approximately 800 mm$^3$ per hemisphere \citep{kapakin2011claustrum,milardi2015cortical}, more recent work using high-resolution imaging has reported significantly larger volumes, ranging from 1,243 mm$^3$ to 2,074 mm$^3$ per hemisphere \citep{calarco2023establishing,coates2024high,kang2020comprehensive}.
Discrepancies in nomenclature also exist. Most studies divide the claustrum into dorsal and ventral portions (e.g., \citet{mathur2014claustrum,coates2024high}), whereas some \citep{kang2020comprehensive,casamitjana2024next} refer to its most inferior portion extending into the temporal lobe as the ``temporal claustrum" - a region considered part of the ventral claustrum in the former classification. In this study, we follow the former dorsal–ventral convention.

The claustrum has been shown to have the highest connectivity in the brain per volume unit \citep{torgerson2015dti}, with vast anatomical connections to almost all cortical areas \citep{baizer2014comparative, coates2023high,crick2005function,smith2020claustrum,dillingham2017claustrum}, including temporal \citep{rodriguez2024functional}, motor \citep{goll2015attention}, somatosensory \citep{goll2015attention}, visual \citep{goll2015attention,rodriguez2024functional,milardi2015cortical}, auditory \citep{goll2015attention,milardi2015cortical}, limbic \citep{goll2015attention,smith2019role,torgerson2015dti}, associative \citep{goll2015attention}, sensorimotor \citep{rodriguez2024functional,milardi2015cortical} and prefrontal cortices \citep{goll2015attention,rodriguez2024functional,milardi2015cortical,brown2017new}, and language areas \citep{rodriguez2024functional}. It has also been shown to be highly interconnected with various subcortical structures \citep{goll2015attention,dillingham2017claustrum}, including basal ganglia and amygdala \citep{rodriguez2024functional,milardi2015cortical}. Besides its wide structural connectivity, claustrum functions have also been object of interest, with hypotheses about its role in consciousness \citep{smythies2012hypotheses,torgerson2015dti,chau2015effect} and conscious percepts \citep{crick2005function}, multisensory integration \citep{mathur2014claustrum,banati2000functional,hadjikhani1998cross,naghavi2007claustrum}, fluency heuristic decisions \citep{volz2010just},
cognitive control \citep{madden2022role,krimmel2019resting,white2018anterior}, task switching \citep{krimmel2019resting,mathur2014claustrum}, saliency detection \citep{mathur2014claustrum,rodriguez2024functional,remedios2014role}, attention \citep{goll2015attention,brown2017new,smith2020claustrum,atlan2018claustrum,mathur2014claustrum} and salience-guided attention \citep{smith2019role,smith2020claustrum,terem2020claustral}. Additionally, other studies suggested that the claustrum may regulate cortical excitability \citep{atilgan2022human}, modulate cortical down-states during sleep \citep{smith2020claustrum}, and it may be involved in the processing of visual sexual stimuli \citep{redoute2000brain} and in mental preparation leading to successful problem solving \citep{tian2011neural}. However, there is still no clear consensus about its functions, with some contradictory results and many open questions \citep{mathur2014claustrum,remedios2010unimodal,atilgan2022human,coates2023high,dillingham2017claustrum,baizer2014comparative}. 

The claustrum has also been suggested to play a role in various neurological disorders \citep{nikolenko2021mystery,benarroch2021role}, such as autism \citep{wegiel2015neuronal,davis2008claustrum}, Parkinson's disease \citep{sitte2017dopamine,sener1998lesions,kalaitzakis2009clinical,arrigo2019claustral}, schizophrenia \citep{cascella2011insula,cascella2014claustrum,bernstein2016bilaterally}, major depressive disorder \citep{bernstein2016bilaterally}, Wilson's disease \citep{sener1993claustrum,sener1998lesions}, Alzheimer's disease \citep{venneri2014claustrum,bruen2008neuroanatomical}, Lewy body dementia \citep{kalaitzakis2009clinical,yamamoto2007correlation}, epilepsy \citep{silva2018claustrum,meletti2015claustrum}, seizures \citep{zhang2001susceptibility,wada1997involvement}, delusional states \citep{patru2015new}, disruption of consciousness \citep{koubeissi2014electrical} and asphyxia \citep{sener1998lesions}. However, these findings need to be corroborated by further analysis, as some of the studies suffer from small sample sizes and several aspects remain unexplored.

Additional investigation of the claustrum and its functions is therefore of great importance, and MRI offers a valuable, non-invasive approach, with excellent soft tissue contrast. However, difficulties arise from the thin shape of the structure, which makes it barely visible in \textit{in vivo} MRI scans at typical resolutions and poses many challenges for both its manual labeling and automatic segmentation. Neuroimaging tools for the study of claustrum are currently very limited. The majority of structural brain atlases for instance do not include claustrum, with the exceptions of a few histological atlases \citep{ding2016comprehensive,mai2015atlas,casamitjana2024next,ewert2018toward,calarco2023establishing,calarco2024cytoarchitectonic}. 
Recently, rigorous protocols for claustrum manual labeling, including both its dorsal and ventral components, have been developed using high-resolution \textit{in vivo} \citep{kang2020comprehensive} and \textit{ex vivo} \citep{coates2024high} MRI images. However, manual labeling is a time-consuming process which requires domain expertise - even more so for a thin and challenging structure like the claustrum - and the protocols developed on high-resolution images might not be applicable at lower resolutions. 
Consequently, the development of a reliable method for automatic claustrum segmentation is essential to advance our understanding of the structure and its functional roles. This method may also be employed for \textit{post hoc} correction of automatic segmentations in the surrounding region, where the putamen is often mislabeled to include parts of the external capsule and the claustrum \citep{perlaki2017comparison,dewey2010reliability}. This is particularly common in Bayesian segmentation methods, which are widely used in neuroimaging, potentially biasing subsequent region of interest (ROI) analyses.

A few approaches have been proposed in the literature for automatic claustrum segmentation: \citet{berman2020automatic} used an intensity-based approach on T1- and T2-weighted \textit{in vivo} scans at 0.7 mm isotropic resolution, with k-means clustering used to separate the claustrum from white matter, in a region of interest detected through anatomical landmarks. \citet{brun2022automatic} adopted a single-atlas segmentation approach: It created a brain atlas of claustrum at 0.5 mm isotropic, by registering together 7T T1-weighted \textit{in vivo} scans of multiple subjects and then manually labeling the claustrum in the resulting template. The atlas was then deformed into subject spaces to yield the claustrum segmentation on new cases. Similarly, \citet{coates2024high} and \citet{coates2023high} manually labeled the claustrum on a single ultra-high resolution \textit{ex vivo} MRI image (0.1 mm isotropic), and then nonlinearly registered it on unseen \textit{in vivo} scans to produce automatic segmentations. Finally, \citet{albishri2022unet} and \citet{li2021automated} trained neural networks based on 2D U-Net architectures \citep{ronneberger2015u} to segment claustrum on T1- and/or T2-weighted \textit{in vivo} scans at 0.7 mm isotropic and 1 mm isotropic resolution respectively, while \citet{neubauer2022efficient} extended the model from \citet{li2021automated} to neonatal brains using transfer learning. However, these approaches suffer from some shortcomings: \citet{berman2020automatic} achieves low performances and includes only the dorsal portion of the claustrum, \citet{brun2022automatic} and \citet{coates2024high} do not quantify segmentation accuracy, and \citet{albishri2022unet} and \citet{li2021automated} are difficult to generalize to unseen datasets. In particular, the latter is a limitation of many supervised Convolutional Neural Networks (CNNs): Their ability to generalize across datasets is hampered by the huge variability in acquisition protocols and sequences, often necessitating fine-tuning or re-training of the models for applicability to unseen domains.

In this paper, we propose a novel approach to claustrum segmentation which addresses these shortcomings: We manually labeled claustrum on ultra-high resolution (0.1-0.25 mm isotropic) \textit{ex vivo} and \textit{in vivo} images where claustrum is clearly visible. Then, as \textit{ex vivo} and \textit{in vivo} have different contrasts, we trained a contrast- and resolution-agnostic deep learning network to automatically segment the claustrum. Because of the contrast- and resolution-independent nature of the network, we can also apply this method to \textit{in vivo} datasets at standard ($\approx$1 mm isotropic) resolution of any contrast, which is a scenario of interest in many applications.
The proposed method is based on SynthSeg \citep{billot2023synthseg}, a segmentation framework that only requires a set of label maps to be trained; no real intensity scans are needed since all training intensity images are synthesized from the labels. In particular, a 3D U-Net \citep{ronneberger2015u} is trained on heavily augmented versions of the provided labels and on corresponding intensity images that are synthesized on-the-fly with varying contrast and resolution, making the automatic segmentation insensitive to these variations. The use of synthetic images has been widely leveraged in various neuroimaging applications \citep{billot2023synthseg,hoffmann2021synthmorph,iglesias2021joint,iglesias2023synthsr,hoopes2022synthstrip}, and it allows the network to have a rich training environment, overlooking the idiosyncrasies of a specific training dataset and yielding state-of-the-art generalization \citep{billot2023synthseg,billot2023robust}. 

The use of this contrast- and resolution-insensitive method yields the following advantages: (1) During training, it enables leveraging of ultra-high-resolution labels from \textit{ex vivo} scans, which capture both dorsal and ventral claustrum components, while subsequently applying the method to segment the entire claustrum in standard-resolution \textit{in vivo} images.
(2) It allows the method to achieve state-of-the-art generalization across different \textit{in vivo} datasets of the same modality, regardless of the specific scanner and acquisition protocol used. (3) It enables claustrum segmentation in \textit{in vivo} images across a wide range of contrasts and resolutions, including T1-weighted, T2-weighted, proton density, and quantitative T1 scans. To the best of our knowledge this is the first method for automatic high-resolution claustrum segmentation that yields accurate segmentation across all of claustrum and displays excellent generalization. The method is released at \url{https://github.com/chiara-mauri/claustrum_segmentation}, as well as in the neuroimaging software FreeSurfer \citep{fischl2012freesurfer}.

\section{Data}\label{sec:Data}

\begin{figure}[t!]
\centering
\includegraphics[trim={0cm 0cm 0cm 0cm}, align = c, clip=true,width=\linewidth]
{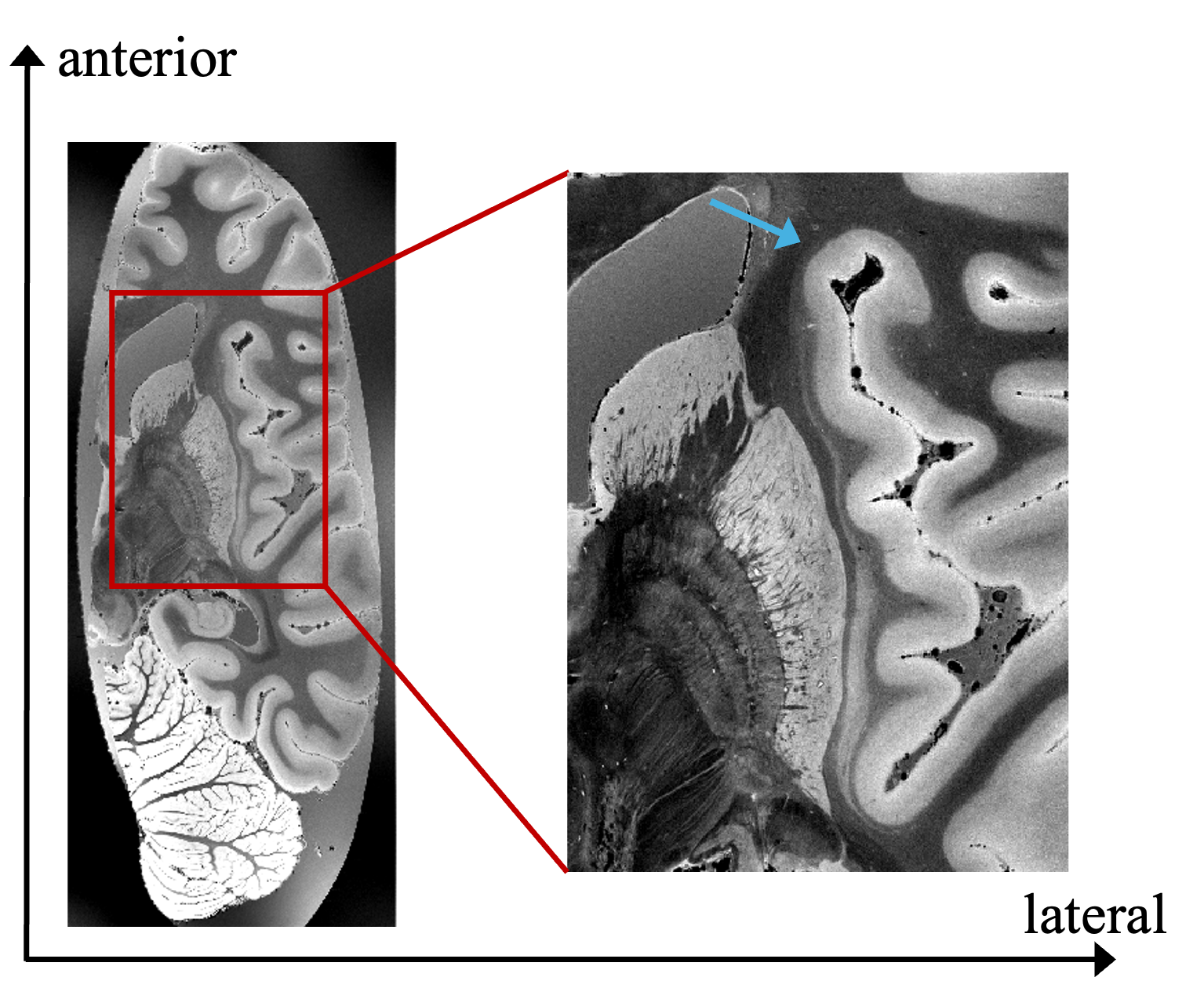}
\caption{Example of an \textit{ex vivo} hemisphere (case 14, left hemisphere, axial view, slice 789, voxel size 0.12 mm), where a cropped region around claustrum is highlighted. Note that the contrast between claustrum and white matter faints in the anterior end of claustrum (blue arrow), raising challenges in the delineation of the structure.}
\label{fig:full_hemi}
\end{figure}

{
\renewcommand\arraystretch{1.5}
\addtolength{\tabcolsep}{+0.4mm}
\begin{table*}[t!]
\tiny
\begin{center}
\resizebox{1\linewidth}{!}{%
\begin{tabular}{|c|c|c|c|c|c|c|c|c|c|}
\hline
Sample & \textit{In vivo}/\textit{ex vivo} & Age 
& Sex & Field of view (FoV) & PMI (hours) & Brain weights (grams) &   MRI resolution (mm, isotropic) \\ \hline
1 & \textit{ex vivo} & 78 & M & left hemisphere & 24 & 1,320 &  0.12 \\ \hline
 2 & \textit{ex vivo} & 75 & F & left hemisphere & 15 & 1,140  & 0.12 \\ \hline
 \makecell {3 \citep{costantini2023cellular} }& \textit{ex vivo} & 79 & M& left hemisphere & 15 & 1,200  & 0.15 \\ \hline
 4 & \textit{ex vivo} & 70 & F & left hemisphere & 23 & 1,103  & 0.12 \\ \hline
 5 & \textit{ex vivo} & 73 & F & right hemisphere & 23 & 1,500  & 0.15 \\ \hline
 6 & \textit{ex vivo}& 62 & M & left hemisphere & 16 & 1,350 &  0.15 \\ \hline
 7 & \textit{ex vivo}& 60 & F & left hemisphere & 2 & 1,380 &  0.15 \\ \hline
 8 &\textit{ex vivo} & 75 & M  & right hemisphere & 24 & 1,310  & 0.12\\ \hline
 9 & \textit{ex vivo}& 60&	M&	left hemisphere&	21	&1,310	& 0.12  \\ \hline
 10 & \textit{ex vivo} & 81&	M	&left hemisphere &	6&	1,270 &	0.12 \\ \hline
 11 & \textit{ex vivo} & 61&M &	right hemisphere &	23	& 1,310		& 0.12 \\ \hline
 12 & \textit{ex vivo}& 52	&F&	left hemisphere &	15	&1,250&	0.12\\ \hline
\makecell{ 13 
\citep{edlow20197}} & \textit{ex vivo}& 58&	F	& whole brain	&14	&1,210		& 0.10  \\ \hline
 14 & \textit{ex vivo}&57&	F	& left hemisphere &	24&	1,215	& 0.12 \\ \hline
 15 &\textit{ex vivo} & 48 & M &  right hemisphere  &? & ?& 0.12  \\ \hline
 \makecell{16 
 \citep{lusebrink2017t1}} & \textit{in vivo} & 34 & M & whole brain & - & -  & 0.25 \\ \hline
\end{tabular}
}
\caption{
Demographics of the ultra high-resolution MRI scans used for manual labeling of the claustrum. None of the \textit{ex vivo} cases had neurological disorders prior to death, and the \textit{in vivo} scan is from a healthy subject. Some information about sample 15 were not available. All samples, except for cases 13 and 16, have been scanned \textit{in situ}.  PMI = postmortem interval.
}
\label{tab:data}
\end{center}
\end{table*}
}
\noindent
To develop the claustrum segmentation method, we used ultra-high resolution MRI scans comprising 15 \textit{ex vivo} samples and one \textit{in vivo} T1-weighted scan, as detailed in Table \ref{tab:data}. In total, 18 hemispheres (12 left, 6 right) were included, with isotropic resolution ranging from 0.1 to 0.25 mm. An example of \textit{ex vivo} hemisphere, with the claustrum region highlighted, is shown in Fig.~\ref{fig:full_hemi}. These data were used to obtain claustrum manual labels and subsequently train a SynthSeg segmentation method, as will be explained in Sec.~\ref{sec:method}. 

We also considered several \textit{in vivo} datasets acquired at typical \textit{in vivo} resolutions, to test the segmentation method in this scenario. In particular, we employed:
\begin{list}{-}{\leftmargin=1em \itemindent=0em}
    \item The IXI dataset,
    which is composed of T1-weighted scans of 581 healthy subjects spanning the whole adulthood (20-86 years old), with resolution $0.94 \times 0.94 \times 1.20$ mm. 
     This dataset comprises scans from three different sites, acquired with a Philips 3T scanner, a Philips 1.5T scanner, and a GE 1.5T scanner.
   
    \item The Miriad dataset \citep{malone2013miriad}, which contains repeated T1-weighted scans of 46 subjects with mild–moderate Alzheimer's disease (AD) (mean age 69.4 $\pm$ 7.1 years) and of 23 healthy controls (mean age 69.7 $\pm$ 7.2 years), acquired by the same radiographer with the same scanner and sequences. The data resolution is $0.94 \times 0.94 \times 1.50$ mm. For each subject, we used the first two scans, which were taken on average 16 days apart from each other.
    \item The FreeSurfer Maintenance (FSM) dataset \citep{ds004958:1.0.0}, which is composed of multimodal images of 39 subjects, with 1 mm isotropic voxel size. Specifically, all subjects in the dataset have a T1-weighted MPRAGE scan, while 31 subjects also have a T2-weighted scan.
    For 36 subjects, a Proton Density (PD)-weighted (M0 image) and a quantitative T1 (qT1) scans were also computed from an MP2RAGE, via steady-state Bloch equations \citep{marques2010mp2rage}. The same method was also used to synthesize MPRAGE images (1 mm isotropic) with different inversion times (TI), based on PD and qT1 scans. Each of these modalities presents a different contrast, enabling the evaluation of the segmentation method across diverse imaging conditions.
\end{list}

\section{Method}\label{sec:method}

\subsection{Claustrum manual labeling protocol}\label{sec:labeling}
\noindent
The claustrum was manually annotated in all hemispheres listed in Table~\ref{tab:data}. Two raters contributed sequentially: Rater 1 (J.M.) first labeled seven hemispheres (cases 4, 13, 14, 15, and 16), while rater 2 (R.F.) completed the remaining cases, some of which had been briefly initiated by rater 1. To assess inter-rater variability, rater 2 independently re-labeled all seven cases previously annotated by rater 1. These alternative labels were used solely for inter-rater comparison and were not merged with the original annotations.

Both raters labeled claustrum in the coronal view on every 5th slice using the FreeView software - part of the FreeSurfer software package \citep{fischl2012freesurfer} - with the voxel edit tool (brush size of 1 or 2 voxels). For labeling larger areas, they delineated an initial outline, followed by filling the structure, while in narrower regions labeling was performed voxel by voxel. Additionally, the raters were instructed to optimize the contrast in Freeview to better distinguish the claustrum from adjacent structures, especially for samples with subtle differences in signal intensity.

\begin{figure*}[t!]
\centering
\setlength{\myWidth}{0.24\linewidth}
\setlength{\mySpace}{-3mm}
\begin{tabular}{ccc}
\centering
\vspace{5mm}

\begin{subfigure}[t]{\myWidth}
\includegraphics[trim={0cm 0cm 0cm 0cm}, clip=true,width=\linewidth,align=c]
{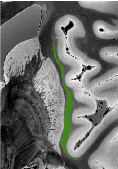}
\caption{Axial view (slice 789)}
\label{subfig:label_ax}
\end{subfigure} &

\hspace{\mySpace}
\begin{subfigure}[t]{\myWidth}
\includegraphics[trim={0cm 0cm 0cm 0cm}, clip=true,width=\linewidth,align=c]
{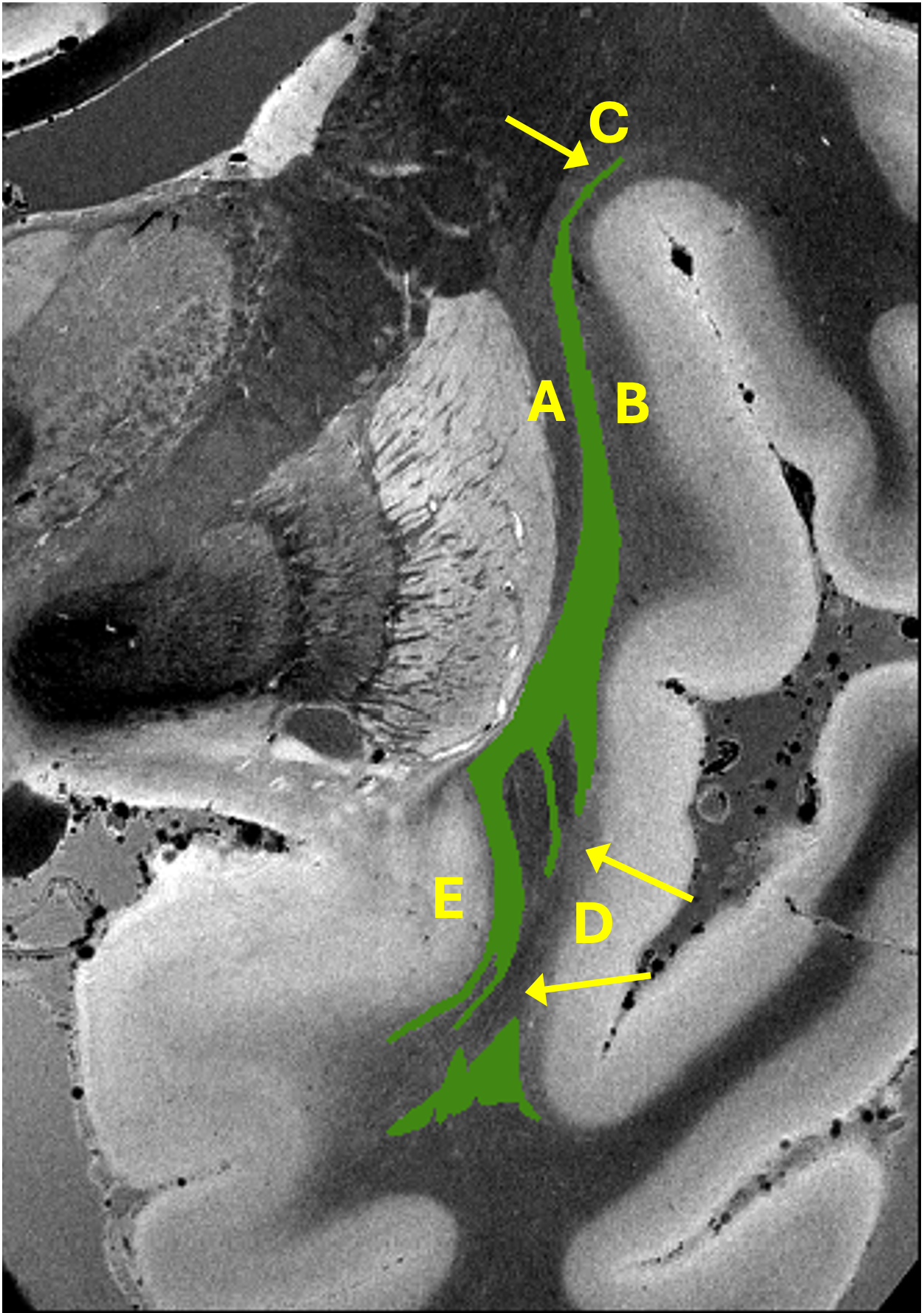}
\caption{Coronal view (slice 720)}
\label{subfig:label_cor}
\end{subfigure} 
 &

\hspace{\mySpace}
\begin{subfigure}[t]{0.25\linewidth}
\includegraphics[trim={0cm 0cm 0cm 0cm}, clip=true,width=\linewidth,align=c]
{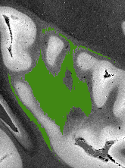}
\caption{Sagittal view (slice 345)}
\label{subfig:label_sag}
\end{subfigure}
\end{tabular}

\begin{tabular}{cc}
\centering

\begin{subfigure}[t]{0.4\linewidth}
\includegraphics[trim={0cm 0cm 0cm 0cm}, clip=true,width=\linewidth,align=c]%
{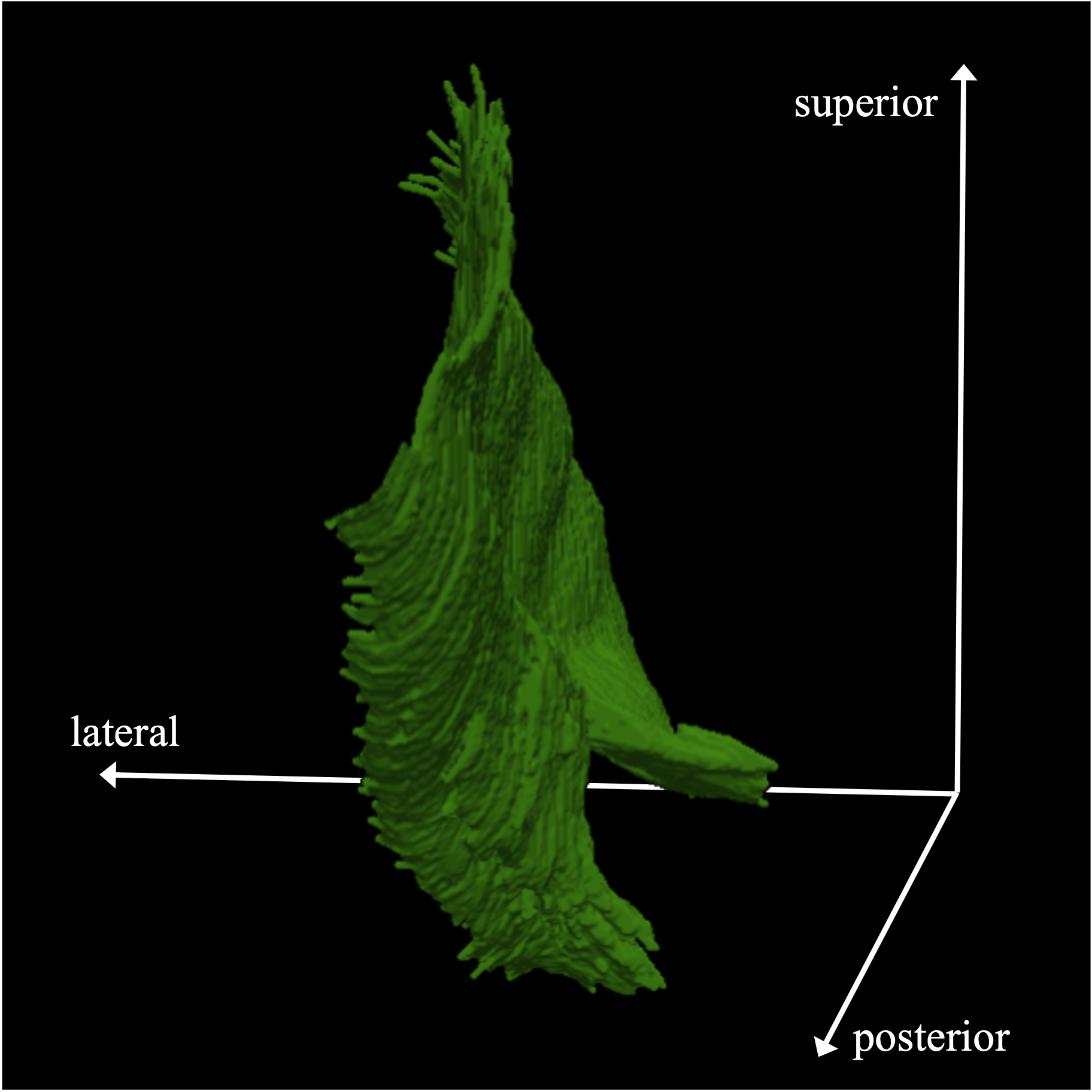} 
\caption{3D view}
\label{subfig:label_3D1}
\end{subfigure}
 &

\hspace{-4mm}
\begin{subfigure}[t]{0.4\linewidth}
\includegraphics[trim={0cm 0cm 0cm 0cm}, clip=true,width=\linewidth,align=c]%
{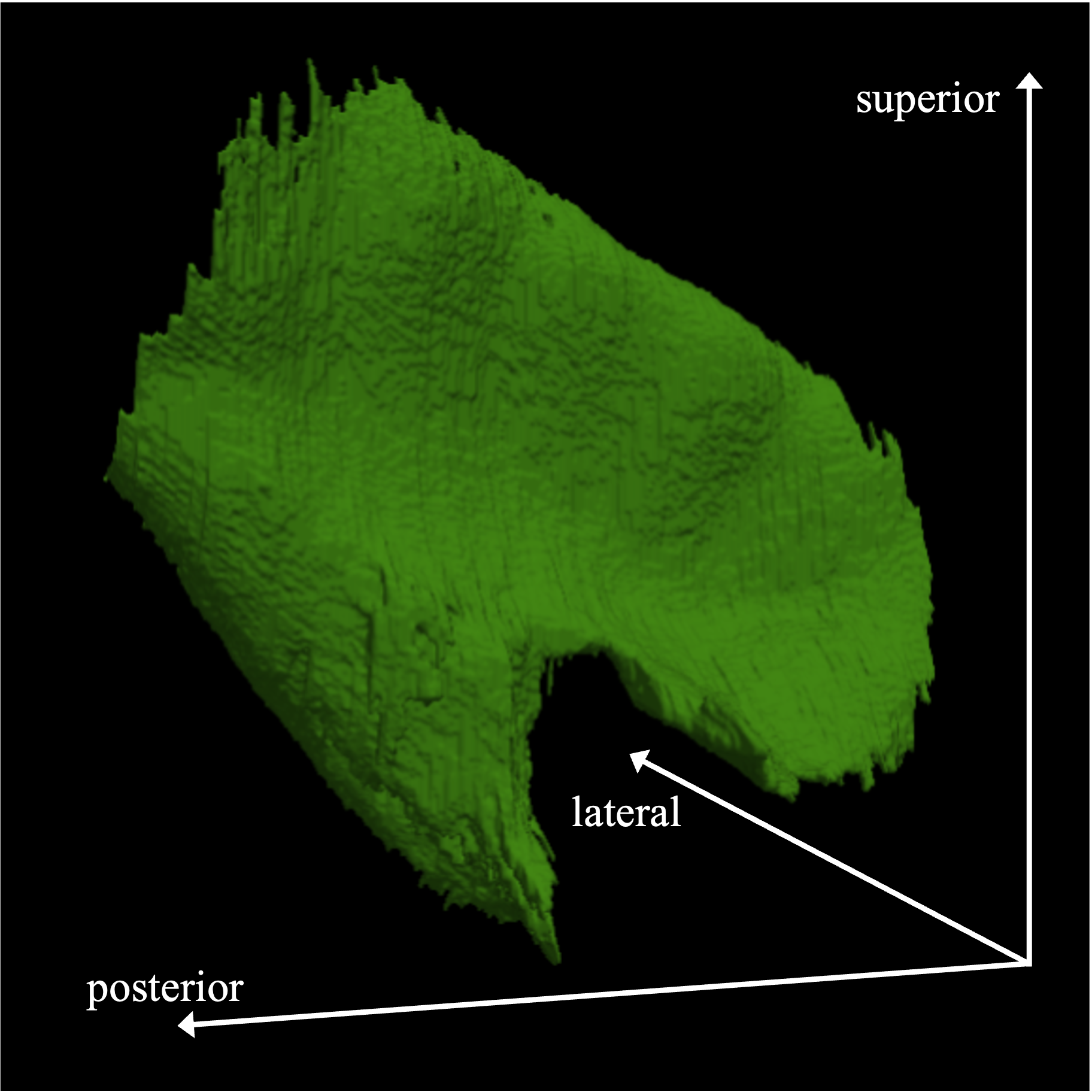} 
\caption{3D view}
\label{subfig:label_3D2}
\end{subfigure} 
\end{tabular}
\caption{(a-b-c): Different views of an \textit{ex vivo} hemisphere (case 14, left hemisphere) cropped around claustrum, with overimposed manual label. The coronal view (b) shows both dorsal and ventral portions of the claustrum, and was used for manual tracing. The dorsal part is thinner, enclosed between the external (A) and the extreme (B) capsules, and ultimately wraps around the superior cortical gyri (C). The ventral portion is wider, displays the characteristic ``fingers" (D), and closely follows the contour of the ventral amygdala (E).
 (d-e): 3D rendering of the claustrum manual label of the same case, shown from two different angles. Note that the structure has only one connected component and that the fragmented look of the ventral ``fingers" appears only in 2D views.}
\label{fig:manual_label}
\end{figure*}

Both dorsal and ventral portions of the claustrum were labeled. Tracing started by locating the superior (or dorsal) portion of the claustrum, which is an extremely thin but tall area of gray matter residing within cortical white matter, broadly located between the putamen medially and the insula laterally in its midsection. Two narrow isthmuses of white matter occur on either side of the dorsal claustrum, the external capsule medially (Fig.~\ref{subfig:label_cor}, A) and the extreme capsule laterally (Fig.~\ref{subfig:label_cor}, B) at its midsection. In the superior direction, the claustrum continues as a narrow strip, extends beyond its midway neighbors (putamen and insula), and ultimately curves toward and hugs the superior cortical gyri (Fig.~\ref{subfig:label_cor}, C), so tracing followed accordingly. The superior curve of the claustrum points laterally in respective frontal or parietal lobes.

\begin{figure}[t!]
\centering
\setlength{\myWidth}{0.38\linewidth}
\setlength{\mySpace}{-3mm}
\centering
\begin{tabular}{cc}
\vspace{3mm}
\begin{subfigure}[t]{\myWidth}
\includegraphics[trim={0cm 0cm 0cm 0cm}, clip=true,width=\myWidth,align=c]
{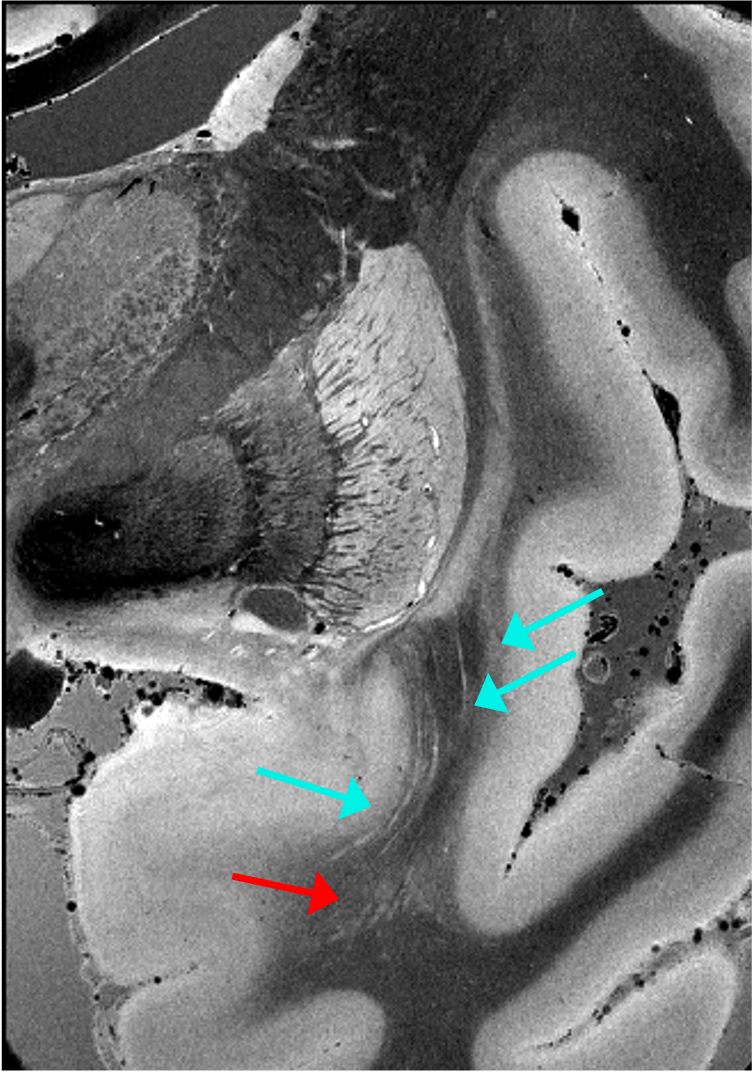}
\caption{Coronal view (slice 720)}
\end{subfigure}
&
\hspace{\mySpace}
\begin{subfigure}[t]{\myWidth}
\includegraphics[trim={0cm 0cm 0cm 0cm}, clip=true,width=\myWidth,align=c]
{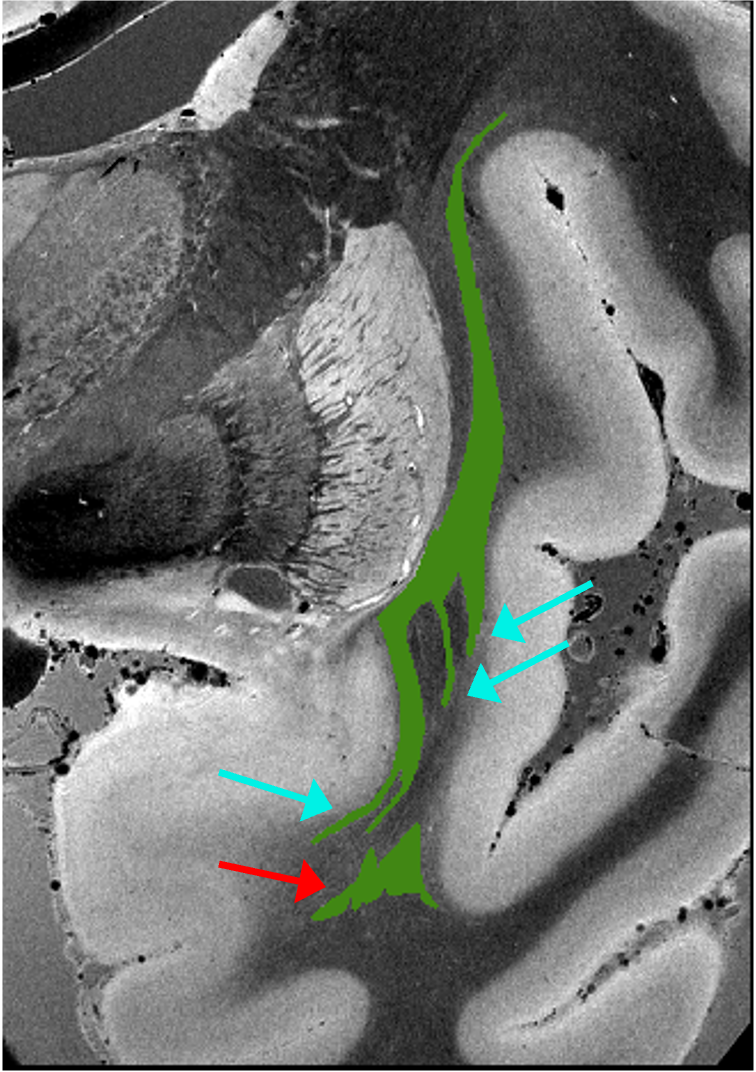}
\caption{Coronal view (slice 720) with manual label}
\end{subfigure} \\
\begin{subfigure}[t]{\myWidth}
\includegraphics[trim={0cm 0cm 0cm 0cm}, clip=true,width=\myWidth,align=c]
{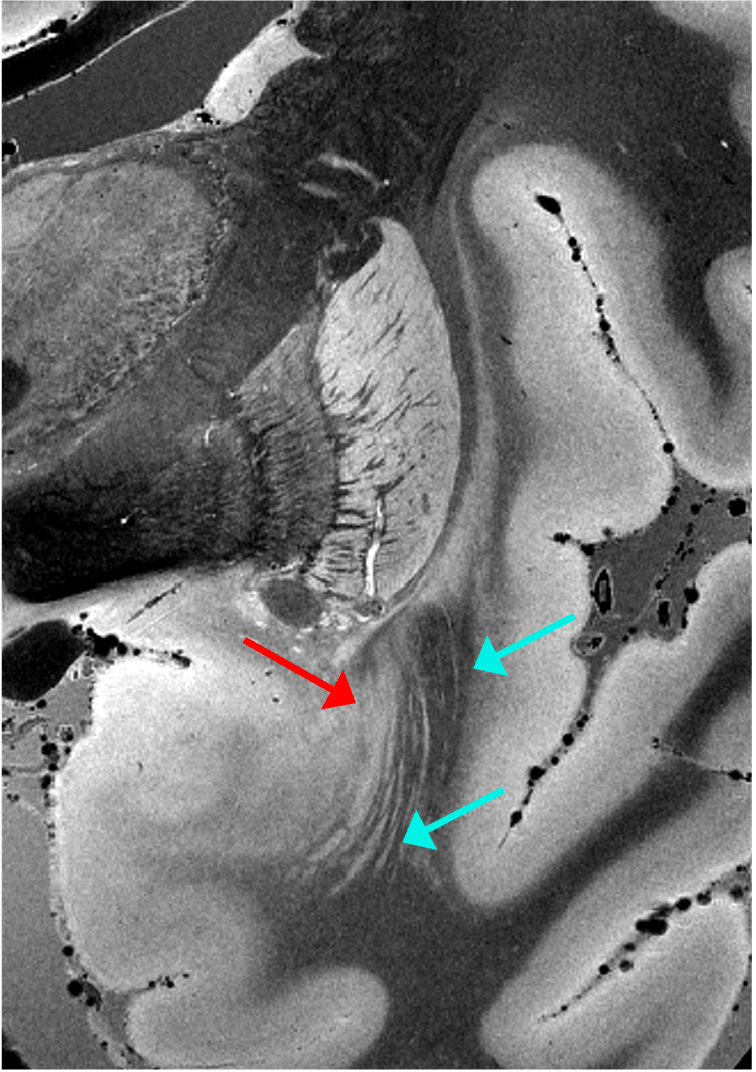}
\caption{Coronal view (slice 730)}
\end{subfigure}
 &
 \hspace{\mySpace}
 \begin{subfigure}[t]{\myWidth}
\includegraphics[trim={0cm 0cm 0cm 0cm}, clip=true,width=\myWidth,align=c]
{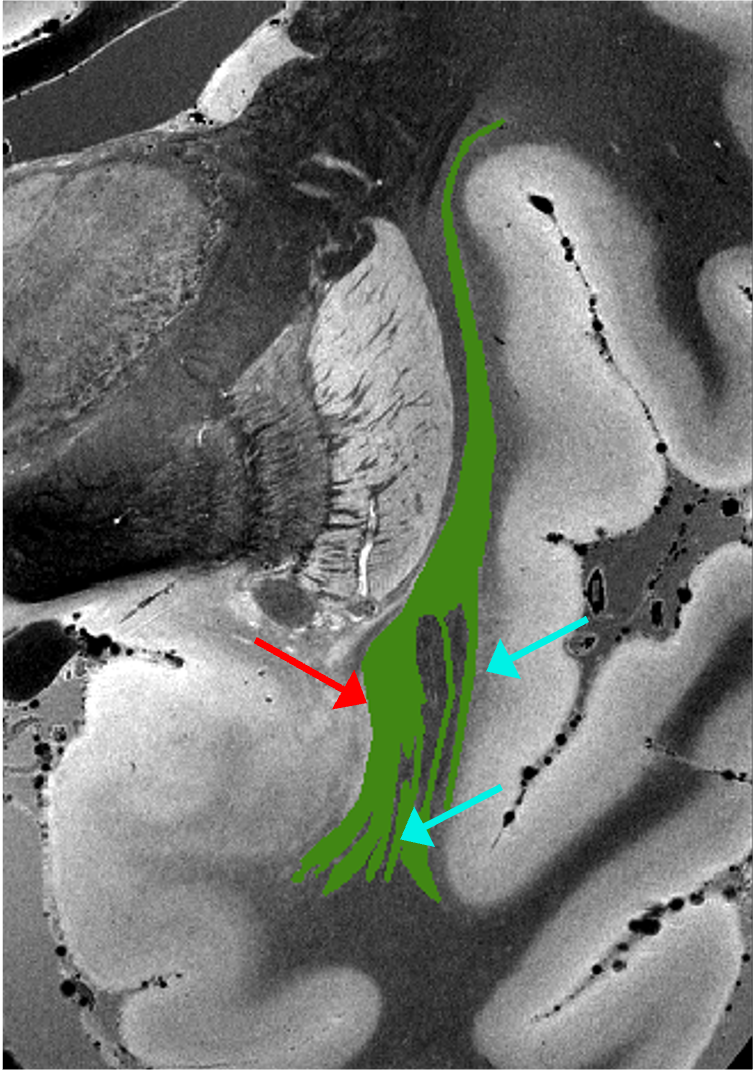}
\caption{Coronal view (slice 730) with manual label}
\end{subfigure}
\end{tabular}
\caption{Two coronal views of an \textit{ev vivo} hemisphere (case 14, left hemisphere) cropped around claustrum, with overimposed manual label. Light-blue arrows point to ventral ``fingers" that have been annotated individually, while red arrows highlight regions where they were labeled as a whole, due to voxel size limitations.}
\label{fig:labels_fingers}
\end{figure}

Tracing then proceeded towards the inferior (or ventral) claustrum, which fans out into divided sections at the temporal stem, showing a fan-like structure (or ``fingers"; Fig.~\ref{subfig:label_cor}, D). 
The ventral claustrum occupies a greater territory than the superior portion (which does not expand out at the corona radiata); It is not restrained by the narrow zone at the midsection and expands ventrally near the temporal stem white matter. When feasible, individual ventral ``fingers" were delineated separately, whereas they were labeled as a single structure if their spacing was smaller than the voxel size (see Fig.~\ref{fig:labels_fingers}). Special care was taken in the mid-ventral region to prevent the claustrum label from extending into the ventral amygdala (Fig.~\ref{subfig:label_cor}, E).

This protocol was used to manually label claustrum in every 5th coronal slice. In the remaining slices, labels were generated with SmartInterpol \citep{atzeni2018probabilistic}, an automated method that exploits a combination of label fusion and convolutional neural networks to transfer the existing labels to the slices without annotations. SmartInterpol has the advantage of reducing the burden of manual tracing, while providing labels that vary smoothly across slices (see \suppsecref{supp-sup_sec:SI} for a visual example). The strategy of labeling every 5th slice was adopted after testing intervals of 2, 4, 5, and 10 slices and visually inspecting the resulting final labels (results not shown).

After applying SmartInterpol, the segmentation in each coronal slice was cross-checked against adjacent slices, and axial as well as 3D views were also inspected (shown in Figs.~\ref{subfig:label_ax} and \ref{subfig:label_3D1}- \ref{subfig:label_3D2} respectively). Manual adjustments were made when necessary to ensure label accuracy and consistency across slices, alternating as needed between axial, coronal, and 3D views. The axial view was particularly useful for detecting minor underlabeling of the dorsal claustrum, especially in its most superior region.

\subsection{Whole field of view (FoV) labels}\label{sec:preprocessing}
\begin{figure*}[t!]
\centering
\setlength{\myWidth}{\linewidth}
       \includegraphics[trim={0cm 0cm 0cm 0cm}, clip=true,width=\myWidth]
      {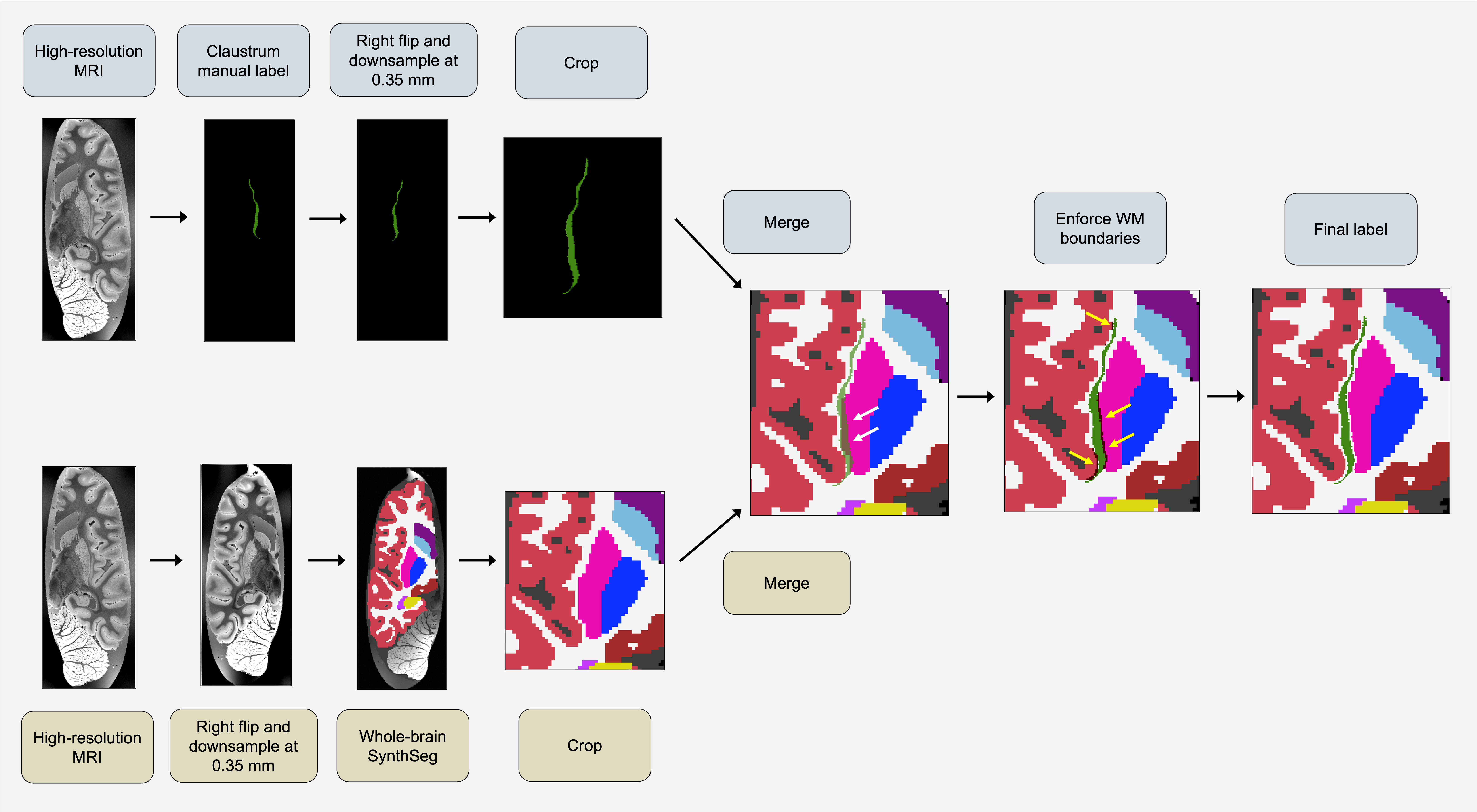}
\caption{Overview of the pipeline used to create training labels. Case 14 (left hemisphere) is shown. In the merged figure, white arrows highlight regions where whole-brain SynthSeg incorrectly labeled the claustrum as putamen. The claustrum is shown with partial transparency to make these errors visible. In the subsequent figure, yellow arrows indicate voxels that were manually reassigned to white matter (WM); these appear as black regions in the image.
}
\label{fig:overview_train_labels}
\end{figure*}

\begin{figure*}[t!]
\centering
\setlength{\myWidth}{0.2\linewidth}
      \begin{tabular}{ccc}
       \includegraphics[trim={0cm 0cm 0cm 0cm}, clip=true,width=\myWidth]
      {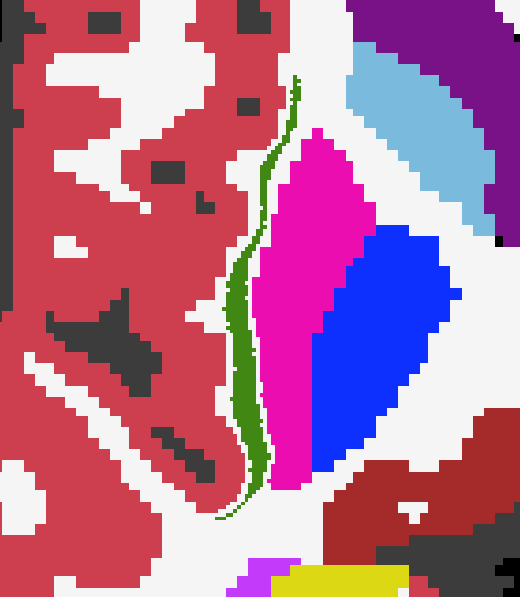} &
\includegraphics[trim={0cm 0cm 0cm 0cm}, clip=true,width=\myWidth]{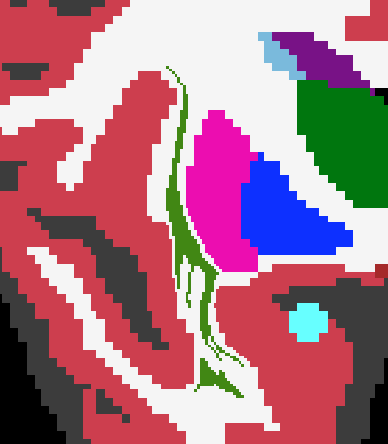} &
\includegraphics[trim={0cm 0cm 0cm 0cm}, clip=true,width=0.23\linewidth]{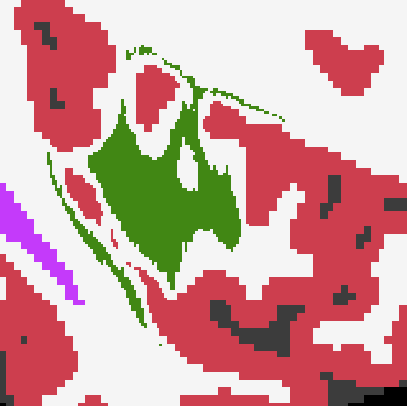} 
\end{tabular}
\caption{
Example of axial (left, slice number 270), coronal (middle, slice number 247), and sagittal (right, slice number 118) view of a label map (case 14, left hemisphere flipped on the right side), obtained by overimposing the claustrum manual label to the automatic segmentation of surrounding structures, which were used to train the SynthSeg segmentation model. The claustrum label is at 0.35 mm resolution, whereas the other structures are output by the whole-brain SynthSeg at 1 mm. The slices shown here correspond to those in Fig.~\ref{fig:manual_label}, given the claustrum voxel sizes of 0.35 mm vs 0.12 mm respectively.
}
\label{fig:labels}
\end{figure*}

\noindent
After obtaining claustrum manual annotations as described in Sec.~\ref{sec:labeling}, the left hemisphere labels were left-right reversed so that all labels appeared on the right hemisphere. All label volumes where cropped to a FoV of 56 mm per dimension and downsampled to 0.35 mm isotropic voxel size (160 voxels per dimension); this is necessary to accommodate GPU memory during training. SynthSeg training requires dense labels where the whole FoV is annotated, not just the claustrum, so we used the default SynthSeg whole-brain network \citep{billot2023synthseg} to segment the surrounding structures - e.g. putamen, cortex, white matter - from the MRI scans (all on the right hemisphere, downsampled to 0.35 mm isotropic), and then cropped the resulting segmentations with the same FoV as the manual labels. The claustrum manual labels were then superimposed on the SynthSeg segmentations, mostly overwriting voxels that have been labeled by SynthSeg as white matter or putamen. Additionally, all voxels adjacent to the claustrum were forced to be white matter in the final labels, to ensure that white matter appeared between the claustrum and other structures, such as putamen and insula. These final labels will be used for training the segmentation model. Fig.~\ref{fig:overview_train_labels} provides an overview of the training labels creation pipeline, while Fig.~\ref{fig:labels} presents an example of a final label, including both claustrum and surrounding structures.

\begin{figure*}[t!]
\centering
\setlength{\myWidth}{0.15\linewidth}
\setlength{\mySpace}{-3mm}
\begin{tabular}{cccccc}
 \hspace{\mySpace}
      \includegraphics[trim={0cm 0cm 0cm 0cm}, clip=true,width=\myWidth]
      {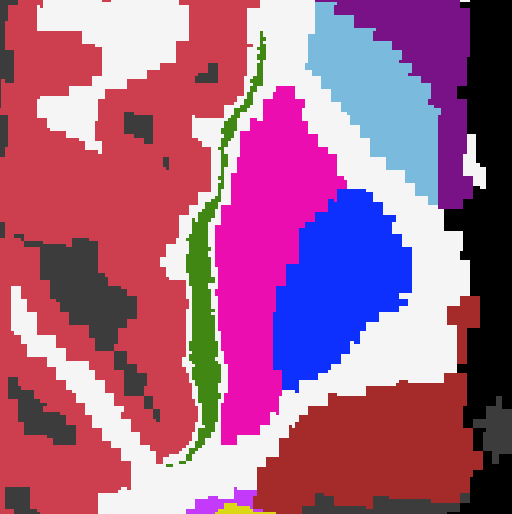} &
  \hspace{\mySpace}
\includegraphics[trim={0cm 0cm 0cm 0cm}, clip=true,width=\myWidth]      {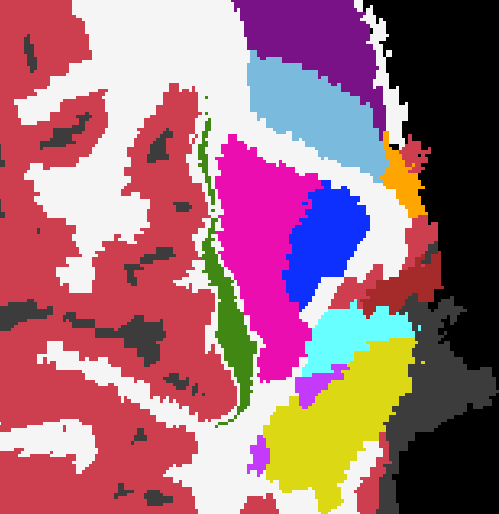} &
   \hspace{\mySpace}
   \includegraphics[trim={0cm 0cm 0cm 0cm}, clip=true,width=\myWidth]
          {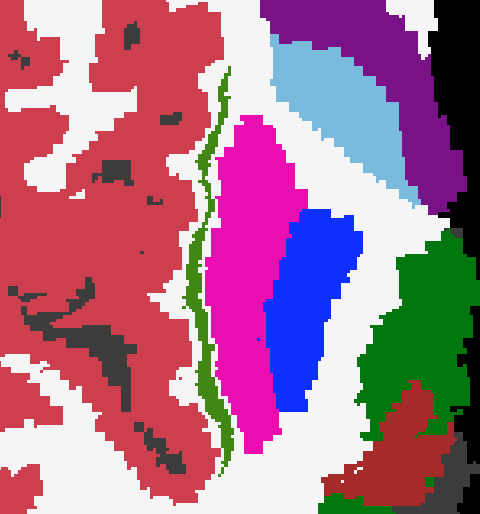} &
\includegraphics[trim={0cm 0cm 0cm 0cm}, clip=true,width=\myWidth]      {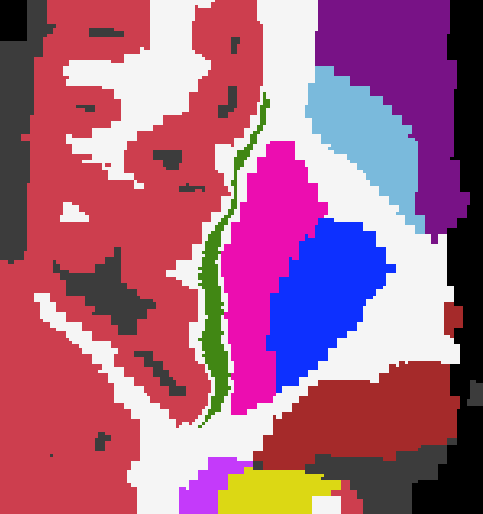} &
  \hspace{\mySpace}
   \includegraphics[trim={0cm 0cm 0cm 0cm}, clip=true,width=\myWidth]
{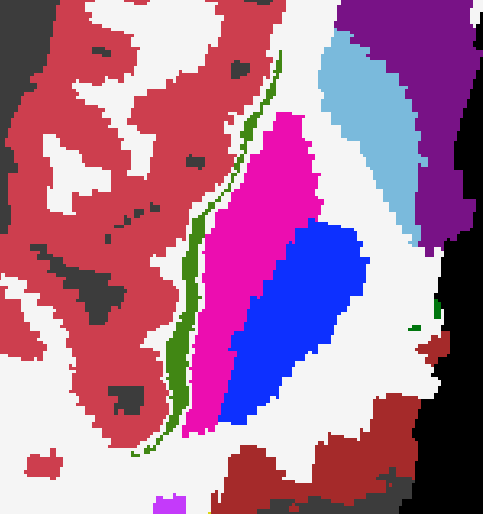} &
     \hspace{\mySpace}
\includegraphics[trim={0cm 0cm 0cm 0cm}, clip=true,width=\myWidth]      {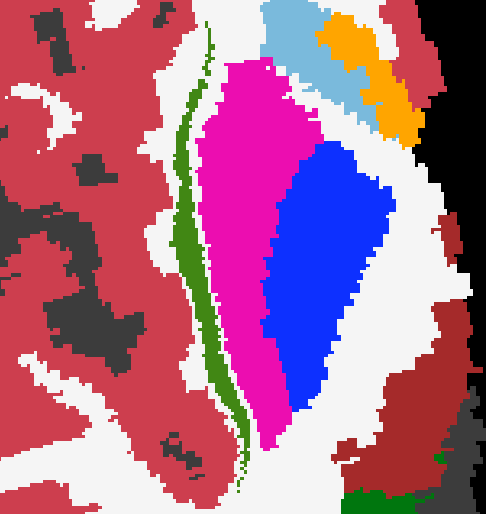}  \\
 \hspace{\mySpace}
      \includegraphics[trim={0cm 0cm 0cm 0cm}, clip=true,width=\myWidth]
      {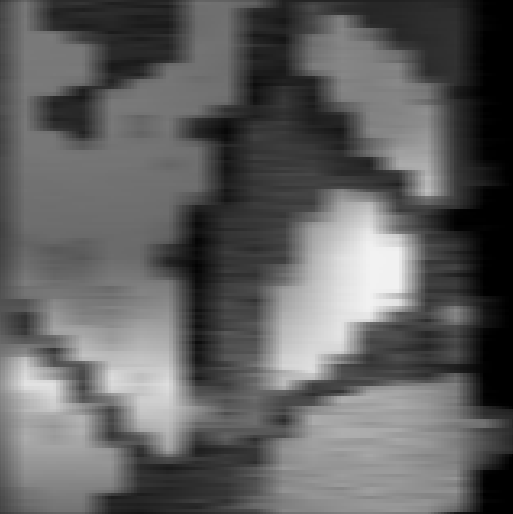} &
  \hspace{\mySpace}
\includegraphics[trim={0cm 0cm 0cm 0cm}, clip=true,width=\myWidth]      {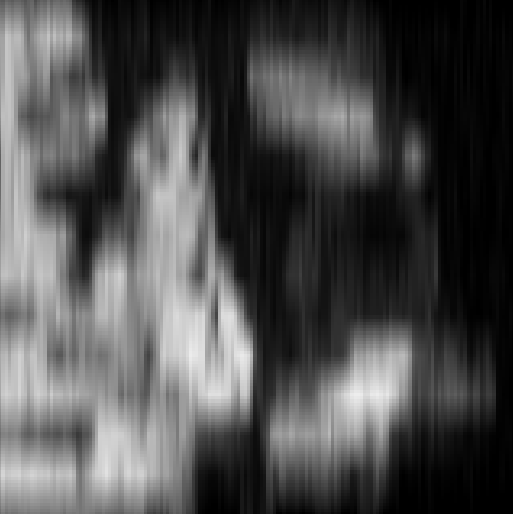} &
   \hspace{\mySpace}
   \includegraphics[trim={0cm 0cm 0cm 0cm}, clip=true,width=\myWidth]
          {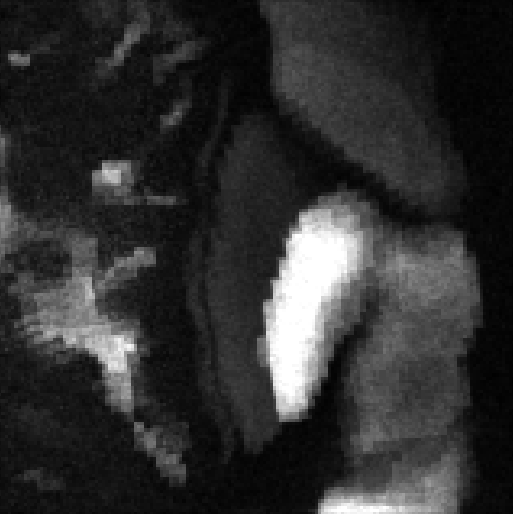} &
\includegraphics[trim={0cm 0cm 0cm 0cm}, clip=true,width=\myWidth]      {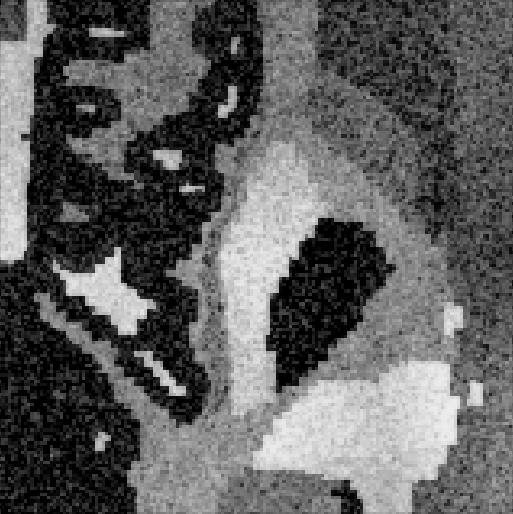} &
  \hspace{\mySpace}
   \includegraphics[trim={0cm 0cm 0cm 0cm}, clip=true,width=\myWidth]
{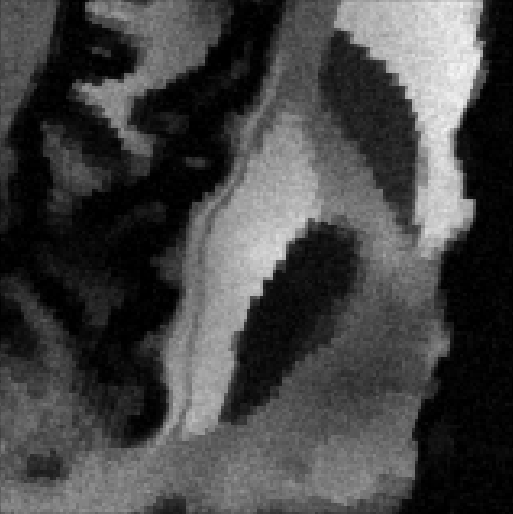} &
     \hspace{\mySpace}
\includegraphics[trim={0cm 0cm 0cm 0cm}, clip=true,width=\myWidth]      {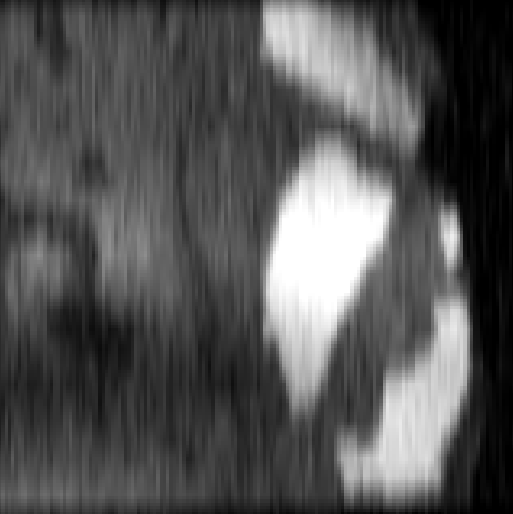}  \\
\end{tabular}
\caption{Top: Example of augmented labels derived from the label shown in Fig.~\ref{fig:labels} (case 14, axial view, slice number 270).
Bottom: Corresponding synthetic images generated with random contrast.
}
\label{fig:synthetic_images}
\end{figure*}

\subsection{Training the automatic segmentation method}\label{sec:training}
\noindent
We trained SynthSeg\footnote{We used the training code available at \url{https://github.com/BBillot/SynthSeg}.} on the labels described in Sec.~\ref{sec:preprocessing} to segment the claustrum and other structures in the ROI (we will show only claustrum segmentation in the remainder). In particular, SynthSeg performs a heavy spatial augmentation of the provided labels with both affine and nonlinear diffeomorphic transformations, to enrich the training environment. Synthetic intensity images are then generated, conditioned on the given labels, by sampling random intensities via a Gaussian Mixture Model. A set of common image augmentations is also applied in the following order: bias field, intensity normalization, gamma transform, Gaussian smoothing and downsampling (to simulate different lower resolutions). Downsampling is performed either isotropically or anisotropically, with the resolution randomly sampled from the uniform distributions $\mathcal{U}(0.35, 3.5)$ and $\mathcal{U}(0.35, 5)$ respectively. The synthetic intensity images are then upsampled back to the labels resolution (0.35 mm isotropic in our case, compared to 1 mm isotropic in the default SynthSeg). This ensures that the network learns to produce sharp, high-resolution segmentations, regardless of the simulated voxel size. A U-Net model \citep{ronneberger2015u} is trained on these label-intensity pairs (see Fig.~\ref{fig:synthetic_images} and \suppfigref{supp-sup_fig:synthetic_images} for examples), with both label augmentation and synthetic intensity image generation performed on the fly to increase variability during training. Note that, since all intensity images are synthesized with random contrast, the presence of a single \textit{in vivo} case in the training set does not influence the learning process. This distinction between \textit{in vivo} and \textit{ex vivo} cases only becomes relevant at test time, when the trained model is applied to real images.

We first performed some pilot experiments to determine the optimal configuration of architecture and synthesis hyperparameters for claustrum segmentation (the configuration adopted in \citet{billot2023synthseg} was optimized for whole-brain segmentation at 1 mm). Compared to the baseline configuration in \citet{billot2023synthseg}, we tweaked different architecture characteristics (such as kernel size, number of levels, number of features, learning rate, batch size, weights attributed in the loss to different structures), and some SynthSeg-specific features (such as grouping of structures in the Gaussian Mixture Model, resolution of synthetic data and hyperparameters of the prior for the amount of rotation and deformation in the labels augmentation, for the Gaussian Mixture Model, and for the gamma transform in the image synthesis). For each modification to the baseline setting, we trained a model for 100 epochs on 12 cases, and we tested it on 3 validation cases. We found that none of these changes yielded a significant improvement in performance (results not shown). We therefore selected the same architecture as in \citet{billot2023synthseg}, given by a 3D U-Net with 5 levels, 2 convolutions per level with kernel size 3x3x3, soft Dice Loss, a learning rate of $10^{-4}$, and a batch size of 1, and the default synthesis parameters, except for one modification in the image synthesis: We added random voxel-wise Gaussian noise when generating synthetic image intensities, since we found that it crucially increased the robustness of the method (see~\ref{sec:appendix}).

With this selected configuration, we then trained the model using 6-fold Cross-Validation (CV) to assess performance: We split the data into a training set of 12 subjects, a validation and a test set of 3 images each, and we repeated the procedure 6 times so that every subject appeared in the test set exactly once. For each split, we trained the model for 100 epochs (equivalent to 500,000 training steps), with synthetic images generated on the fly, independently of the fold. The validation set was used to determine the best epoch - i.e., the epoch that yielded the best validation Dice score (see \ref{sec:test_metrics}) for the claustrum - which was then used to make predictions on the test subjects. We highlight that, whereas the Dice loss used during training was computed on synthetic images and encompassed all structures within the ROI, the validation Dice score used for epoch selection was computed on real images instead, and only for the claustrum.

To apply the method to a new image (for validation and testing in the CV and in all experiments in the remainder), we used the following procedure. We first needed a final step in the pre-processing of the 18 high-resolution dataset: We non-linearly registered each high-resolution intensity image to MNI152\footnote{We used the ICBM 2009c Nonlinear Asymmetric T1-weighted template, with 1 mm isotropic resolution.} using the contrast-insensitive SynthMorph \citep{hoffmann2021synthmorph}, and then mapped the claustrum manual labels for all 18 cases in MNI152 space using this transform (resampling them at 0.35 mm isotropic). We then created a simple prior probabilistic atlas in MNI152 space (shown in \suppsecref{supp-sup_sec:atlas}), by averaging all warped manual labels, and we then left-right flipped it to obtain a probabilistic atlas for the left claustrum as well. These atlases were used to crop a new image into the proper FoV around the claustrum (they are not otherwise used in the segmentation): Given a new test image, we first register it to MNI152, then map the probabilistic atlases into the native space and threshold them at 0.001. Once there, we crop a cube of size 60 mm per dimension around the claustrum prior on each hemisphere and upsample to 0.35 mm isotropic to create images of the appropriate size to feed into the model (note that these FoVs are in the native space of the input image). 
 The network is then applied to the cropped volumes to yield a segmentation of claustrum in the (0.35 mm isotropic) native space. Note that before feeding the left volume into the network, we left-right reverse it, since the model has been trained on right hemispheres. The resulting segmentation is then flipped back into the native orientation.
The registration to establish the FoV again uses SynthMorph so as to maintain contrast-independence. By default, the registration is affine; non-linear registration can be used but it increases the processing time. The purpose of the registration step is just to establish a FoV in the location likely to be around the claustrum and otherwise plays no role in the segmentation itself, so an affine transformation is accurate enough.

Using this method to apply the model to the real intensity images, we assessed CV performance using Dice score as primary metric. To contextualize these results, we did the following: (1) We computed the Dice score between the labels obtained by the two raters on the seven shared samples (described in Sec.~\ref{sec:labeling}) as a measure of inter-rater variability. For four of these samples, the coronal slices manually labeled before applying SmartInterpol were identical for both raters. For these cases we also computed the Dice score on these slices only.
(2) We applied the pre-trained method from \citet{albishri2022unet}\footnote{Available at \url{https://github.com/AhmedAlbishri/AM-UNET}.} to our 0.25 mm isotropic \textit{in vivo} labeled case (sample 16 in Table~\ref{tab:data}, downsampled at 0.7 mm isotropic to match the training resolution of their model) and computed the resulting overlap with ground truth using Dice score. We also assessed CV performance in terms of other test metrics (intersection over union, true positive rate, false discovery rate, robust Hausdorff distance, mean surface distance, and volumetric similarity; see \ref{sec:test_metrics}). We also compared automatic segmentations and manual labels in terms of claustrum volume and its spatial extensions along the x, y, and z axes, defined as the length of edges of the smallest bounding box that encloses the whole claustrum. Finally, we evaluated the impact on CV performance of including a single \textit{in vivo} case in the dataset, and of using labels from one rater versus the other for the samples labeled by both (details in \suppsecref{supp-sup_sec:dryad} and \suppsecref{supp-sup_sec:comparison_train_labels} respectively).

To evaluate the performance of the method on lower-resolution data as well, we downsampled all 18 hemispheres to voxel sizes ranging from 0.4 mm to 3.5 mm isotropic. For each voxel size and CV fold, we tested the model previously trained on high-resolution data on the corresponding downsampled test subjects. We then computed the average Dice score across folds, for each resolution. 
It is important to note that the models remain unchanged, with the lower resolution only impacting the test phase. This simulation experiment enables us to evaluate how well our method, trained at 0.35 mm resolution, captures fine claustrum details when applied to lower-resolution data.

After assessing performance with CV, we trained a final model on all the 18 available cases for 100 epochs (500,000 training steps). We also developed a procedure to assess segmentation quality in absence of ground truth, and used it to choose an epoch of the final model (based on \textit{in vivo} data), as described in the section below.

\subsection{
Quality assessment of segmentations without ground truth
}\label{sec:epoch_selection}
\noindent
The nonlinear registration of the manual labels in MNI space described in Sec.~\ref{sec:training} (which was used to create an atlas of claustrum needed to crop a new intensity image) served two additional purposes: (1) We can compute Dice statistics between any two pairs of manual labels after nonlinear registration; this allows us to understand the variability of the Dice coefficient between manual labels, contingent upon the accuracy of the nonlinear registration. (2) It allows us to compute Dice scores between the automatic segmentation on a new image and the 18 manual labels as a measure of quality (once that image has been nonlinearly registered to MNI152). Note that this nonlinear registration of the new image is only performed as part of the evaluation; it is not needed to do the segmentation.

To evaluate the quality of an automatic segmentation, we therefore nonlinearly register it to MNI152 space using SynthMorph, with trilinear interpolation for resampling. We then compute the Dice score between the registered segmentation and each of the 18 manual labels in MNI space. The \textit{maximum} Dice score across these comparisons is used as our quality control (QC) measure. This QC score can be automatically generated, enabling the assessment of segmentation quality on very large datasets without having to examine each and every image. Only segmentations with low QC scores are in fact flagged for visual review in subject space. The rationale behind this procedure is the following: A high QC score indicates that the segmentation closely resembles at least one manual label after alignment to MNI space, and is thus likely to be accurate. In contrast, a very low QC score signals a notable deviation from all manual labels in MNI space, which could result from a segmentation failure or a registration error. Importantly, the latter case does not compromise our QC procedure, since the segmentation will be inspected \textit{in subject space}, and will therefore pass the visual assessment, if accurate.
In theory, this procedure could be applied to every analysis, but the nonlinear registration from the test subject space to MNI space does substantially increase the execution time, so we only use it for our evaluation of segmentation quality. 

We emphasize that the QC score should not be interpreted as a direct measure of segmentation accuracy. In addition to the possibility of assigning low QC scores to high-quality segmentations due to registration inaccuracies, even a perfect segmentation will not achieve a QC score of 1. In fact, the QC score does not express similarity to the segmentation ground truth (which is assumed to be unavailable), but to a set of reference annotations, rather indicating if the segmentation displays plausible anatomy (contingent upon registration accuracy).
 To define a possible range of good QC scores, we computed QC metrics for the 18 high-resolution manual labels. For each manual case, the QC score was calculated as the maximum value derived from pairwise Dice comparisons with other manual labels, yielding a reference distribution of 18 scores. Given that the manual labels are accurate in the subject space, and their alignment to the MNI152 space was visually validated, this reference distribution outlines the expected range of satisfactory QC scores, while concurrently expressing the intrinsic inter-subject variability of the claustrum.

Finally, we highlight that the probabilistic atlas in MNI space described in Sec.~\ref{sec:training} is solely used for cropping new test images around the claustrum and otherwise plays no role in segmentation or evaluation. This evaluation of segmentation quality relies instead on the individual manual labels in MNI space, which were used to compute the QC scores.

As mentioned in the previous section, we trained the final model on all manual labels for 100 epochs, but we still need to pick the best one from those 100. We based the epoch selection on T1-weighted \textit{in vivo} images, in particular on a set of 20 subjects from the IXI dataset. On this validation set, we ran the automatic segmentation for each epoch and used the nonlinear registration to MNI152 to compute the QC metric as discussed above. We then chose the epoch that had the highest mean QC score on the validation subjects, and used the corresponding model as final segmentation method for the experiments described in the next section.

\section{Experiments on \textit{in vivo} data}\label{sec:experiments}
\noindent
To assess robustness, we applied the final segmentation model to all the 581 \textit{in vivo} images from the IXI dataset, computing the QC score for every subject, as explained in Sec.~\ref{sec:epoch_selection}. We then visually inspected (in subject space) the segmentations with the 50 lowest QC scores to investigate \textit{potentially} problematic segmentations.
We also used claustrum volume in the segmentation as indicator of success or failure and looked at the subjects with the 20 smallest and the 20 largest volumes in the dataset, as additional pool of potentially suspicious cases.
We then compared the distribution of the QC score and claustrum volumes obtained on the IXI subjects to the corresponding ones obtained on the manual labels.
Additionally, we assessed the effect on segmentation of magnetic field strength. A total of 396 subjects were scanned at 1.5T across two sites, while 185 subjects were scanned at 3T. In the absence of ground truth, we compared QC metrics and claustrum volumes between the two groups.

In order to assess test-retest robustness of the proposed method, we applied it to repeated scans from the Miriad dataset: We rigidly registered the two time points to a subject-specific template (using \citet{reuter2012within}) - as typically done in longitudinal analysis to avoid bias towards one specific time point - then applied the method to both images (using the same FoV), and computed the Dice score between the two segmentations to quantify the stability of the method.
We also compared claustrum volumes obtained for AD subjects vs. healthy controls.

To test the generalization ability of the method to different modalities, we applied it to multimodal images from the FSM dataset. For each subject, we applied the method to the T1-weighted scan as baseline, and to the T2-weighted, Proton Density and quantitative T1 scans (when available), using the same FoV. We then computed the Dice score between the obtained segmentations and the one computed on the T1-weighted scan, as a measure of robustness. Additionally, for each subject for whom they were available, we applied the method to all synthetic images (using the same FoV) and then computed the Dice score between each segmentation and the one obtained on the image with TI=1,000, which was set as reference. This experiment allows us to test the robustness of the method with respect to different contrasts in a controlled setting as well as its ability to handle different TIs that may be encountered in acquisition of real images.

\section{Results}

{
\renewcommand\arraystretch{1.3}
\addtolength{\tabcolsep}{+0.4mm}
\begin{table*}[t!]
\tiny
\begin{center}
\resizebox{1\textwidth}{!}{%
\begin{tabular}{|c|c|c|c|c|c|c|}
\hline
Dice score &  IoU & TPR & FDR & 
 HD & MSD & VS
 \\ \hline
0.632 $\pm$ 0.061 &   0.465 $\pm$ 0.066  & 0.676 $\pm$ 0.100 & 0.381 $\pm$ 0.119 & 1.824 $\pm$ 0.585 mm &0.458 $\pm$ 0.124 mm &  0.867 $\pm$ 0.092
\\ \hline
\end{tabular}
}
\caption{
Results of 6-fold CV on the 18 high-resolution cases. For each test metric, mean and standard deviation across the 18 cases are reported. Abbreviations: Intersection over union (IoU), True positive rate (TPR), False discovery rate (FDR), Robust Hausdorff distance (HD), Mean surface distance (MSD), Volumetric similarity (VS).}
\label{tab:CV_results}
\end{center}
\end{table*}
}
\noindent
The CV results on the 18 high-resolution cases are shown in Table~\ref{tab:CV_results}, reporting an average Dice score of 0.632 $\pm$ 0.061. This compares to an inter-rater variability Dice score of 0.805 $\pm$ 0.018 across the seven samples labeled by both raters. In the four cases where the manually annotated slices were identical, the inter-rater Dice score was 0.836 $\pm$ 0.018 when computed on those slice alone, compared to 0.812 $\pm$ 0.027 when computed on the whole labels.
Additionally, applying the method of \citet{albishri2022unet} to our \textit{in vivo} labeled case resulted in a Dice score of 0.299. \suppsecref{supp-sup_sec:curves}, \suppsecref{supp-sup_sec:dryad} and \suppsecref{supp-sup_sec:comparison_train_labels} provide additional details on CV results, including training and validation curves across all folds, CV performance excluding the \textit{in vivo} case, and the impact on results of using labels from one rater versus the other for the seven samples labeled by both, respectively.

\begin{figure*}[t!]
\centering
\begin{tabular}{cc}
\centering
\begin{subfigure}[t]{0.44\linewidth}
    \includegraphics[trim={0cm 0 0cm 0}, clip=true, width=\linewidth, align=t]{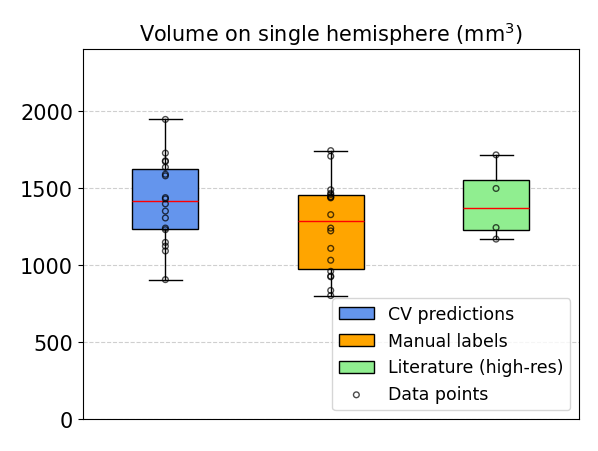} 
    \vspace{-2mm}
    \caption{}
    \label{subfig:volumes_CV}
    \end{subfigure}
    &
    \begin{subfigure}[t]{0.425\linewidth}
    \includegraphics[trim={0cm 0 0cm 0}, clip=true, width=\linewidth,align=t]{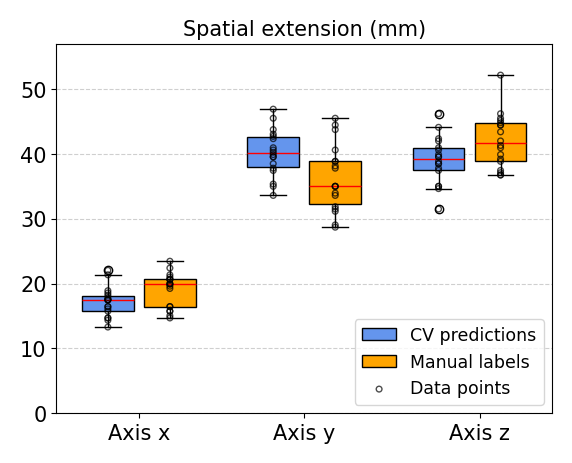} 
    \vspace{-3mm}
    \caption{}
      \label{subfig:xyz_CV}
\end{subfigure} 
\end{tabular}

\begin{subfigure}[t]{\linewidth} 
\footnotesize
\begin{center}
    \centering
    \renewcommand\arraystretch{2.3}
\addtolength{\tabcolsep}{+0.5mm}
    \begin{tabular}{|c|c|c|c|c|}
         \hline
         \makecell{Study} & Space & Claustrum volume & Data modality & \makecell{Voxel size (isotropic)} \\ \hline
          \makecell{\citet{kang2020comprehensive}} & native & 3430 mm$^3$ (bilateral) & \textit{in vivo} T1-weighted & 0.7 mm \\
          \hline
          \makecell{\citet{calarco2023establishing}} & native & 1243 mm$^3$ (left) & histology & 0.1 mm  \\ \hline
          \makecell{\citet{coates2024high} (case 13)} & MNI152 & \makecell[c]{2074.41 mm$^3$ (right)\\ 1736.24 mm$^3$ (left)} & \textit{ex vivo} MRI & 0.1 mm  \\ 
          \hline
          \makecell{\citet{coates2024high} (case 13)} & native & \makecell[c]{1497.0 mm$^3$ (right) \\ 1168.2 mm$^3$ (left)} & \textit{ex vivo} MRI & 0.1 mm  \\ 
          \hline
          \makecell{Our manual label (case 13)} & native & \makecell[c]{1468.9 mm$^3$ (right)\\ 1438.1 mm$^3$ (left)} & \textit{ex vivo} MRI & 0.1 mm  \\
          \hline
    \end{tabular}
    \vspace{1mm}
    \caption{}
    \label{subfig:literature_volumes}
    \end{center}
\end{subfigure}
\vspace{3mm}
\caption{
(a) Distribution of claustrum volumes on a single hemisphere in the 18 automatic segmentations obtained with CV, and in the corresponding manual labels. As a reference, we also display values reported in the literature from other studies using high resolution imaging or histology (\citet{kang2020comprehensive,calarco2023establishing,coates2024high}; details provided in Fig.~\ref{subfig:literature_volumes}). In all boxplots, the red line represents the median, the box spans the inter-quartile range (first to third quartiles), and the whiskers extend to the farthest data point within 1.5 times the inter-quartile range.
(b): Claustrum spatial extension along the x, y, and z axes (defined by the smallest bounding box enclosing the claustrum) for both CV predictions and manual labels. RAS orientation is used: x represents the right-left, y the anterior-posterior, and z the superior-inferior direction. Average x, y, z extensions for CV predictions:  17.15 $\pm$ 2.20 mm, 40.15 $\pm$ 3.47 mm, 39.03 $\pm$ 3.48 mm. Average x, y, z extensions for manual labels: 19.06 $\pm$ 2.59 mm, 36.26 $\pm$ 4.95 mm, 42.02 $\pm$ 4.00 mm. Note that while the comparison of spatial extensions between automatic and manual segmentations is meaningful, their absolute values are instead influenced by variations in head orientation across cases.
(c) Claustrum volumes reported in previous studies using high-resolution imaging or histology. All volumes are derived from a single case. Since \citet{coates2024high} provides volume estimates for case 13 in MNI152 space, for comparison we report in the table the volume of their manual label warped in native space, and the volume of our manual label for the same case too. Note that we included their volumes in native space - and not in MNI152 - in the distribution of volumes from the literature displayed in Fig.~\ref{subfig:volumes_CV}.
}
\label{fig:volumes_xyz_CV}
\end{figure*}

Fig.~\ref{subfig:volumes_CV} shows the distribution of claustrum volumes in the CV automatic segmentations, compared to the ones in the manual labels. We obtained an average of 1,415.35 $\pm$ 260.66 mm$^3$ in the automatic vs 1,253.05 $\pm$ 283.79 mm$^3$ in the manual segmentations (a difference of 162 mm$^3$, around 13\%). For comparison, we also display the distribution of claustrum volumes reported in the literature by studies on high resolution imaging or histology \citep{calarco2023establishing,coates2024high,kang2020comprehensive}, whose details are reported in Fig.~\ref{subfig:literature_volumes}.
We note that \citet{coates2024high} provides volume estimates from manually labeling case 13 \citep{edlow20197} in MNI152 space, whereas all other reported volumes, including ours, are in native space. To ensure comparability, we non-linearly registered their manual labels\footnote{Available at \url{https://osf.io/tbjv4/}} into native space using the inverse of the previously computed SynthMorph warp, and then used the volumes in native space - rather than MNI152 - in the literature analysis. The resulting average hemispheric volume from the literature is 1,405.80 $\pm$ 216.15 mm$^3$, displaying a 12\% difference from our manual labels.
Additionally, Fig.~\ref{subfig:xyz_CV} shows the claustrum spatial extension along the x, y, and z axes, where we observe a relative difference between extensions in CV predictions and manual labels of 9.08\%, 11.92\%, and 6.80\%, in the right-left, anterior-posterior, and superior-inferior directions, respectively.

\begin{figure*}[t!]
\centering
\setlength{\myWidth}{0.3\linewidth}
\begin{tabular}{ccc}
    \begin{subfigure}[b]{\myWidth}
        \centering
        \includegraphics[trim={0cm 0cm 0cm 0cm}, clip=true, width=\linewidth]{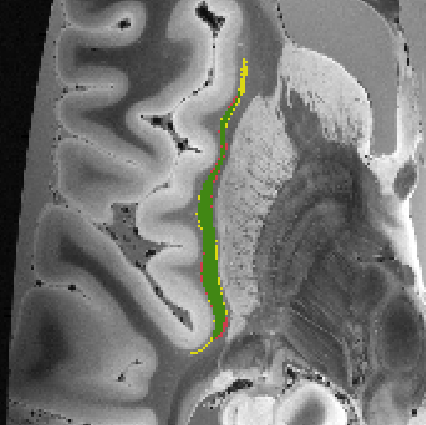}
        \caption{Case 14, axial view, slice 270, CV fold 1}
        \label{fig:sub1}
    \end{subfigure} &
    \begin{subfigure}[b]{\myWidth}
        \centering
        \includegraphics[trim={0cm 0cm 0cm 0cm}, clip=true, width=\linewidth]{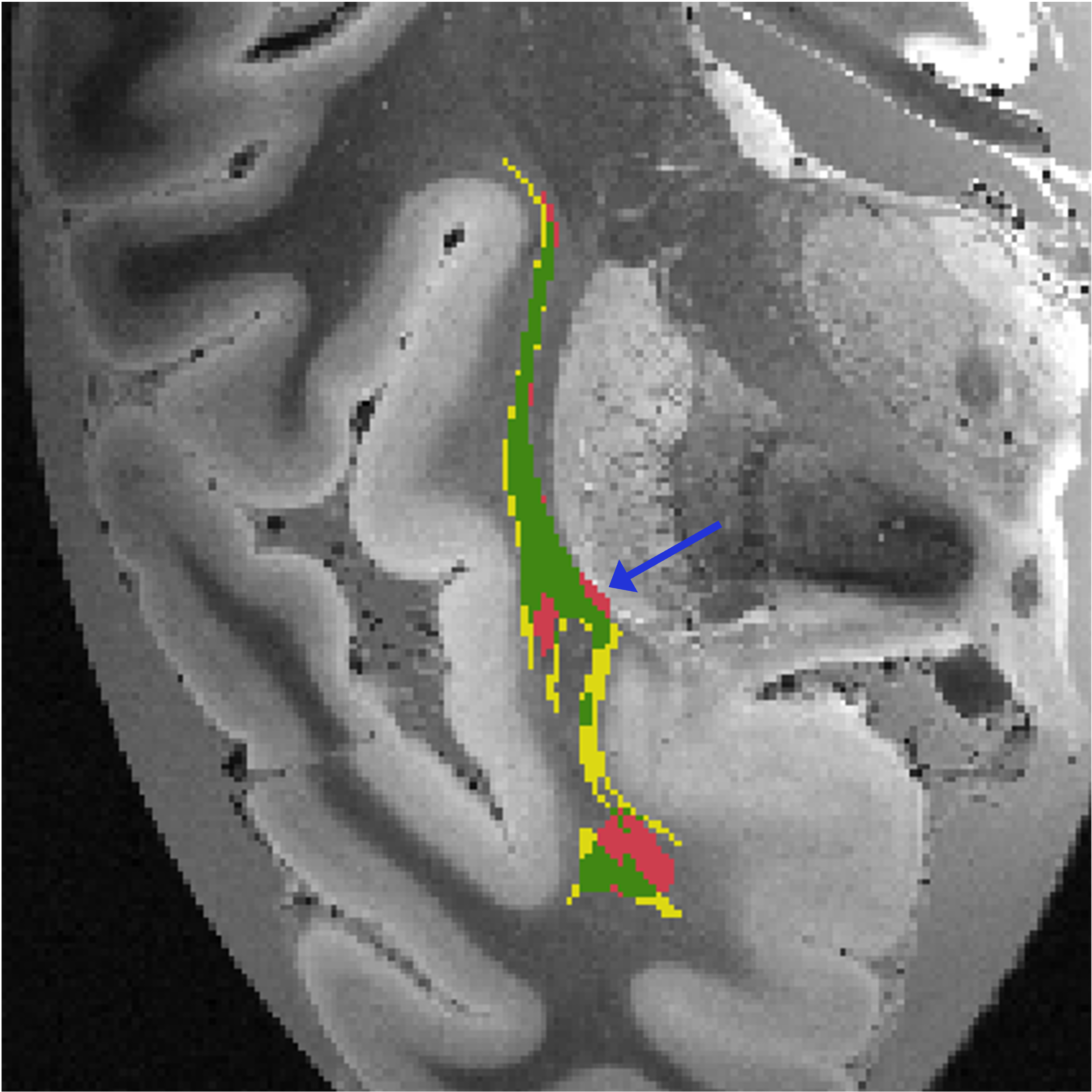}
        \caption{Case 14, coronal view, slice 247, CV fold 1}
        \label{fig:sub2}
    \end{subfigure} 
    &
     \begin{subfigure}[b]{\myWidth}
        \centering
        \includegraphics[trim={0cm 0cm 0cm 0cm}, clip=true, width=\linewidth]{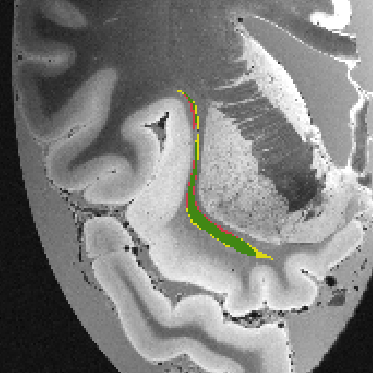}
        \caption{Case 14, coronal view, slice 220, CV fold 1}
        \label{fig:sub3}
    \end{subfigure} 
     \vspace{3mm}
    \\
    \begin{subfigure}[b]{\myWidth}
        \centering
        \includegraphics[trim={0cm 0cm 0cm 0cm}, clip=true, width=\linewidth]{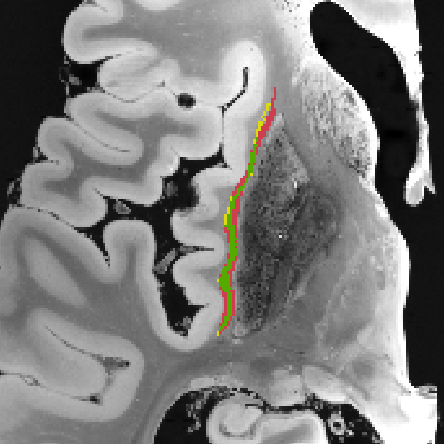}
        \caption{Case 2, axial view, slice 304, CV fold 2}
        \label{fig:sub4}
    \end{subfigure} 
     &
    \begin{subfigure}[b]{\myWidth}
        \centering
        \includegraphics[trim={0cm 0cm 0cm 0cm}, clip=true, width=\linewidth]{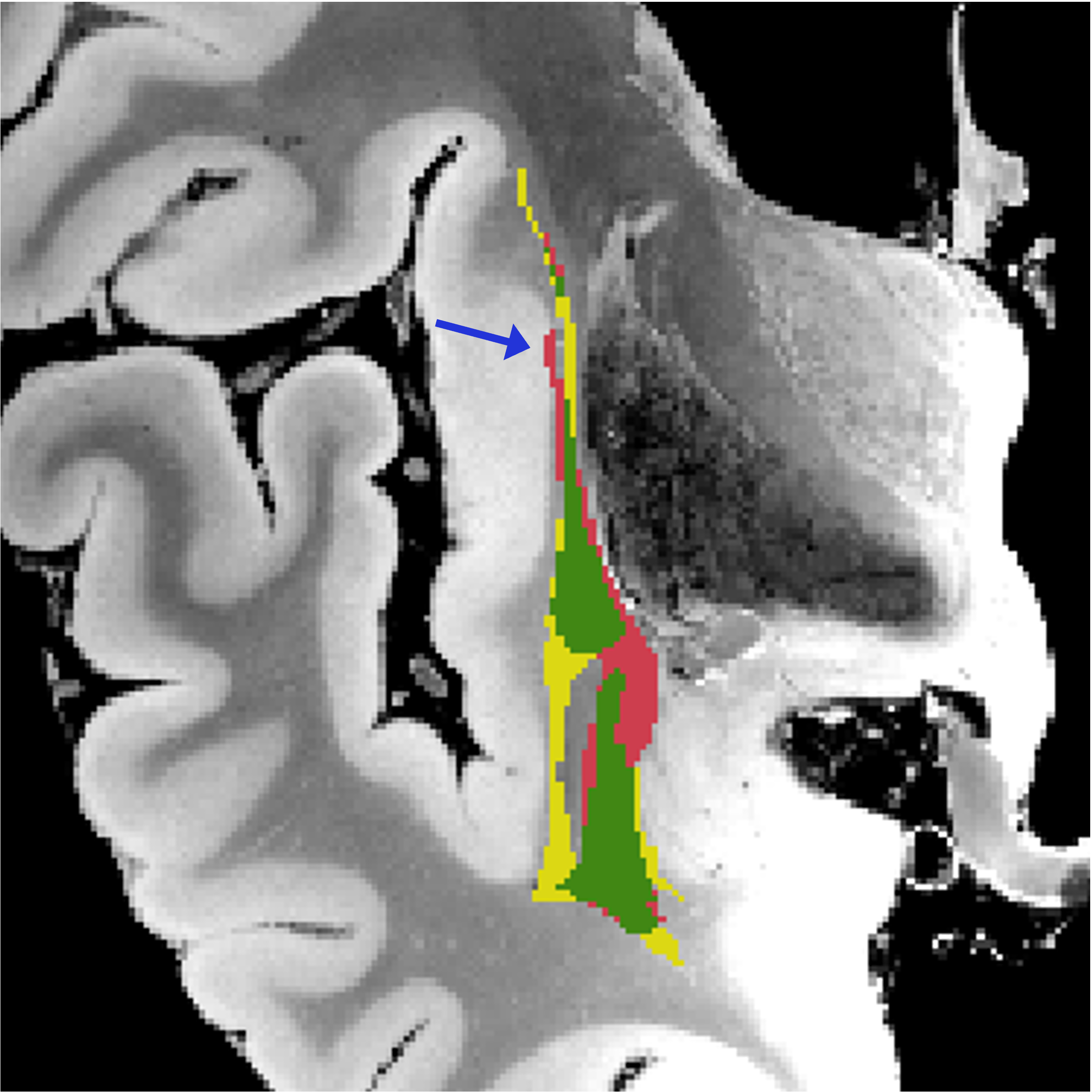}
        \caption{Case 2, coronal view, slice 265, CV fold 2}
        \label{fig:sub5}
    \end{subfigure} 
     &
    \begin{subfigure}[b]{\myWidth}
        \centering
        \includegraphics[trim={0cm 0cm 0cm 0cm}, clip=true, width=\linewidth]{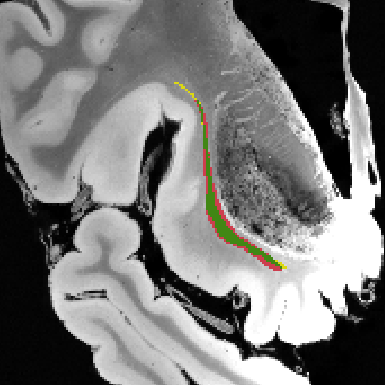}
        \caption{Case 2, coronal view, slice 239, CV fold 2}
        \label{fig:sub6}
    \end{subfigure}
\end{tabular}
\caption{
Axial and coronal views of predictions from the proposed method on two test subjects, using two different CV folds. Green indicates true positives, red false positives, and yellow false negatives. Arrows highlight false positives where the claustrum segmentation overlaps with putamen (due to hyper-intensities, probably Virchow-Robin spaces) or cortex.
Statistics for case 14: Dice = 0.737, TPR = 0.700, FDR =  0.222, HD = 1.443 mm, MSD = 0.332 mm, VS = 0.962, predicted volume = 1,350.03 mm$^3$, real volume = 1,457.60 mm$^3$.
This case shows performance above average and has been selected to allow comparison to Fig.~\ref{fig:manual_label} and Fig.~\ref{fig:labels}.
Statistics for case 2: Dice = 0.616, TPR = 0.700, FDR = 0.450, HD = 1.262 mm, MSD = 0.370 mm, VS = 0.854, predicted volume = 1,239.84 mm$^3$, real volume = 923.6 mm$^3$.
This case has been selected to illustrate examples of overlabeling errors.
}
\label{fig:seg_cv}
\end{figure*}

An example of claustrum segmentation obtained with the proposed method on two test subjects in the CV procedure is shown in Fig.~\ref{fig:seg_cv}. The case in the top row shows in general a good agreement between ground truth and prediction (Dice score = 0.737, predicted volume 1,350 mm$^3$ vs. real volume of 1,457 mm$^3$). Some boundary errors are present, mostly false negatives, particularly at the anterior/posterior and superior/inferior ends, where the claustrum is extremely thin. The bottom row example shows increased false positives along the boundaries, with a very thin ground truth in the dorsal area (Dice score = 0.616, predicted volume 1,239 mm$^3$ vs. real volume of 923 mm$^3$).
Both examples also show segmentation errors in the ventral region, where some ``fingers” are under- or over-segmented.

\begin{figure}[t!]
\centering
\begin{tabular}{c}
      \includegraphics[trim={0cm 0 0cm 0}, clip=true,width=0.95\linewidth]{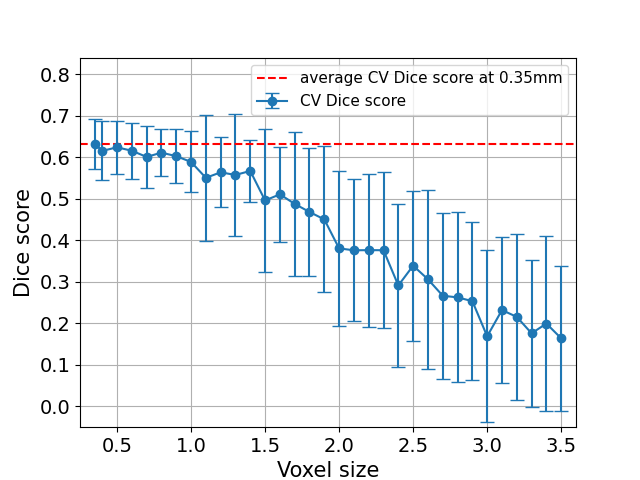}
\end{tabular}
\caption{
Dice scores obtained by CV models trained on high-resolution data and tested on downsampled images. For each voxel size, the dot represents the mean CV Dice score, while the whiskers extend to one standard deviation away from the average.
}
\label{fig:dice_ds}
\end{figure}

Fig.~\ref{fig:dice_ds} shows the Dice score obtained by the CV models trained on high-resolution data and tested on downsampled images. From 0.35 mm to around 1.4 mm, the CV Dice score declines only marginally. Beyond this point, the performance degradation becomes more pronounced.

When we then trained the model on all the 18 high-resolution cases, training took around 7 days on an RTX 8000 GPU. The procedure for identifying the optimal epoch based on the QC score as described in Sec.~\ref{sec:epoch_selection} selected the model of the 19th epoch, which yielded an average QC metric of 0.580 $\pm$ 0.060 on the IXI subset used for validation.

\begin{figure*}[t!]
\centering

\begin{subfigure}[b]{\textwidth}
\begin{tabular}{cccc}
\vspace{3mm}
      \includegraphics[align=c,trim={3cm 8.7cm 3.2cm 5.7cm}, clip=true,width=0.23\linewidth]{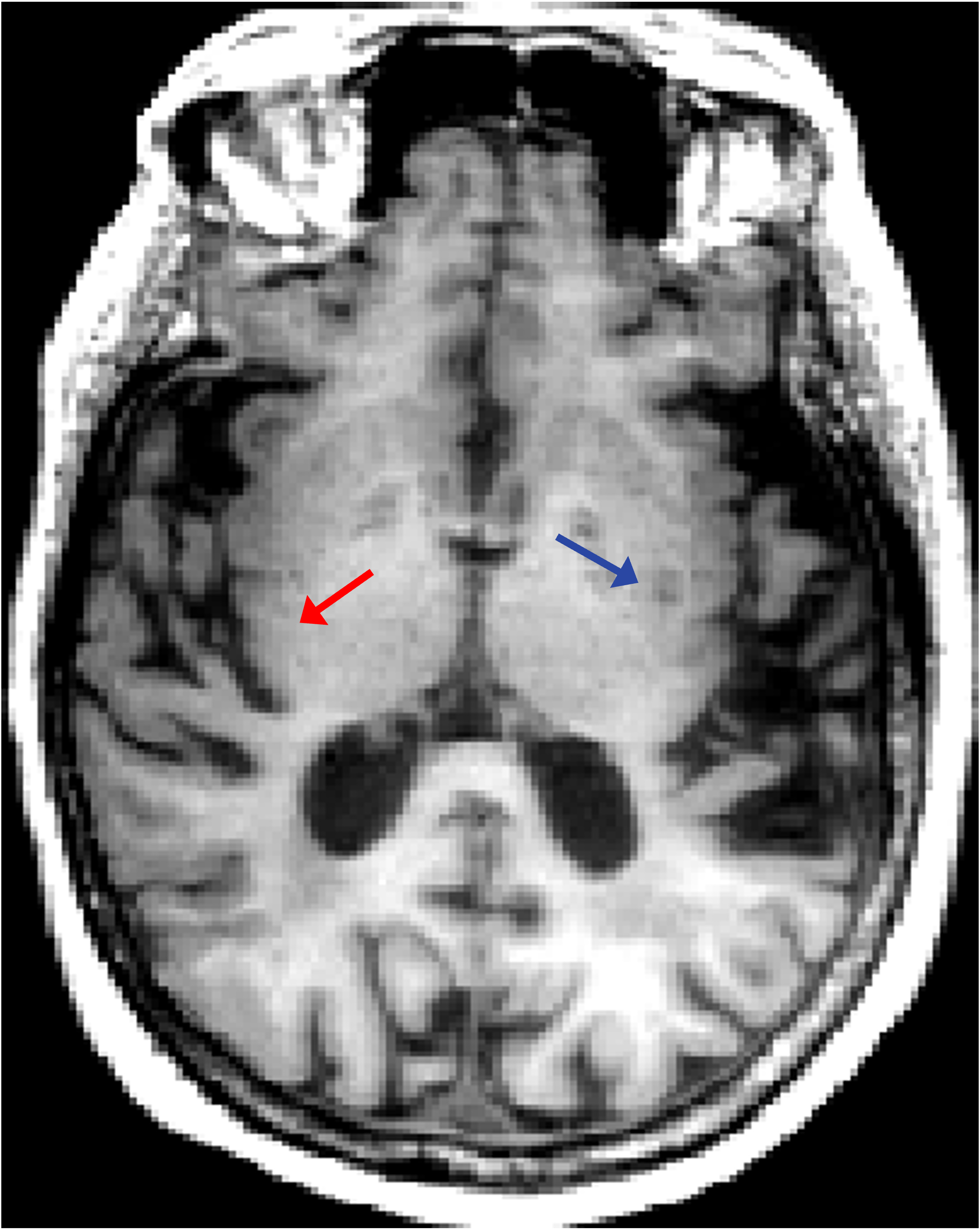} &
      {\setlength{\fboxrule}{2pt}
     \hspace{-5mm}
      \includegraphics[align=c,trim={3cm 8.7cm 3.2cm 5.7cm}, clip=true,width=0.23\linewidth]
    {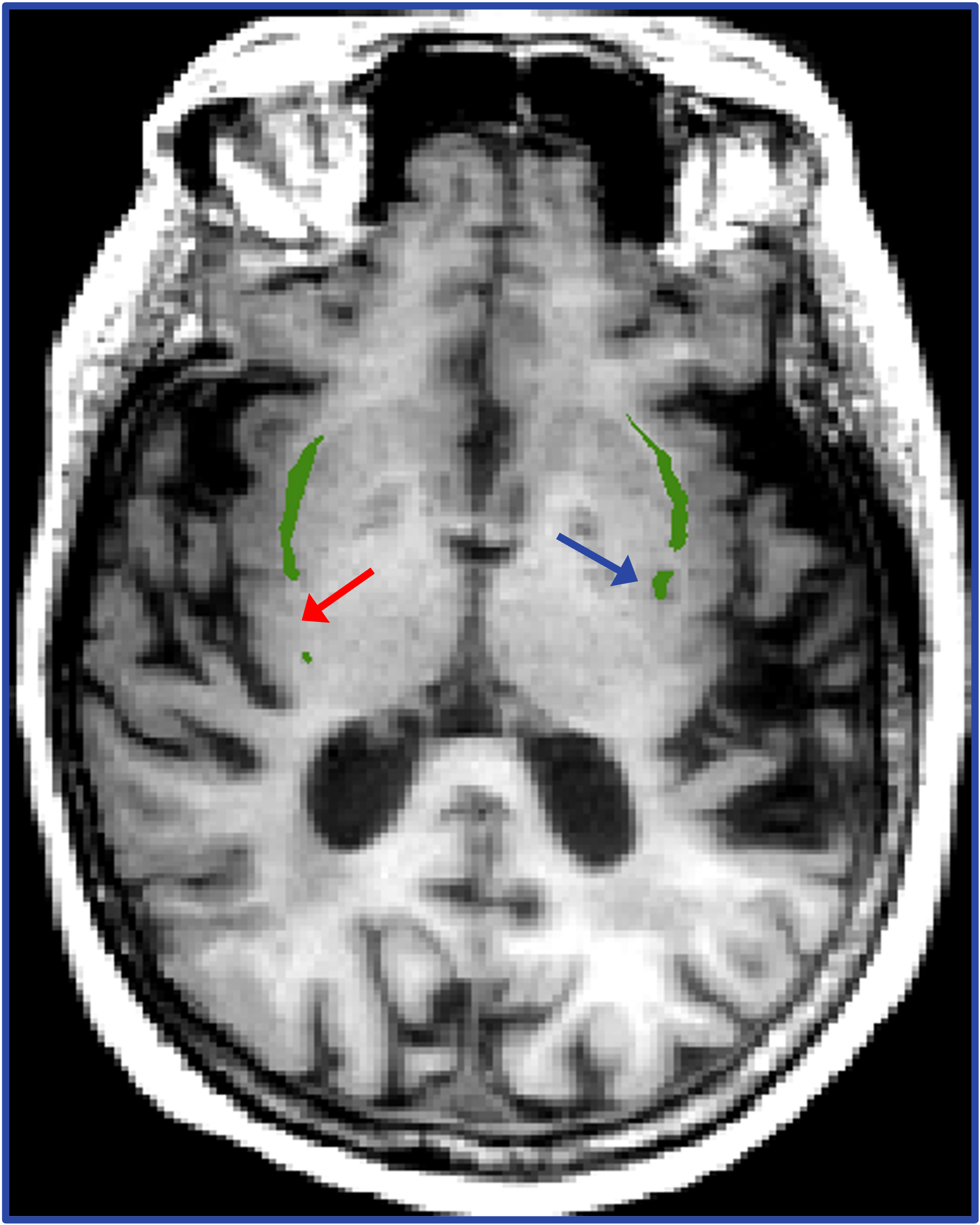}
      }&
      \includegraphics[align=c,trim={3.2cm 10.6cm 3.2cm 5.8cm}, clip=true,width=0.24\linewidth]{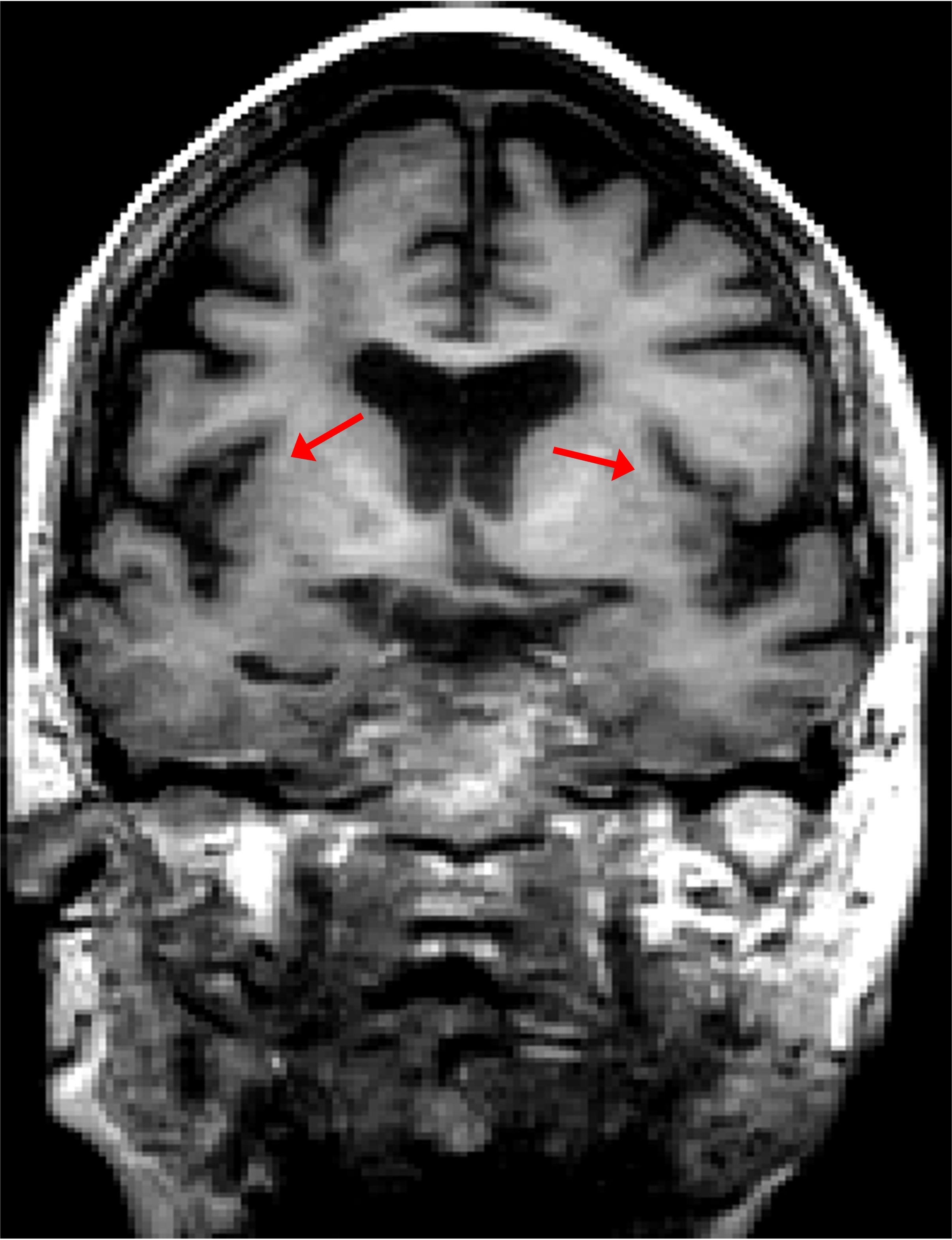} &
      {\setlength{\fboxrule}{2pt}
     \hspace{-5mm}
      \includegraphics[align=c,trim={3.2cm 10.6cm 3.2cm 5.8cm}, clip=true,width=0.24\linewidth]
    {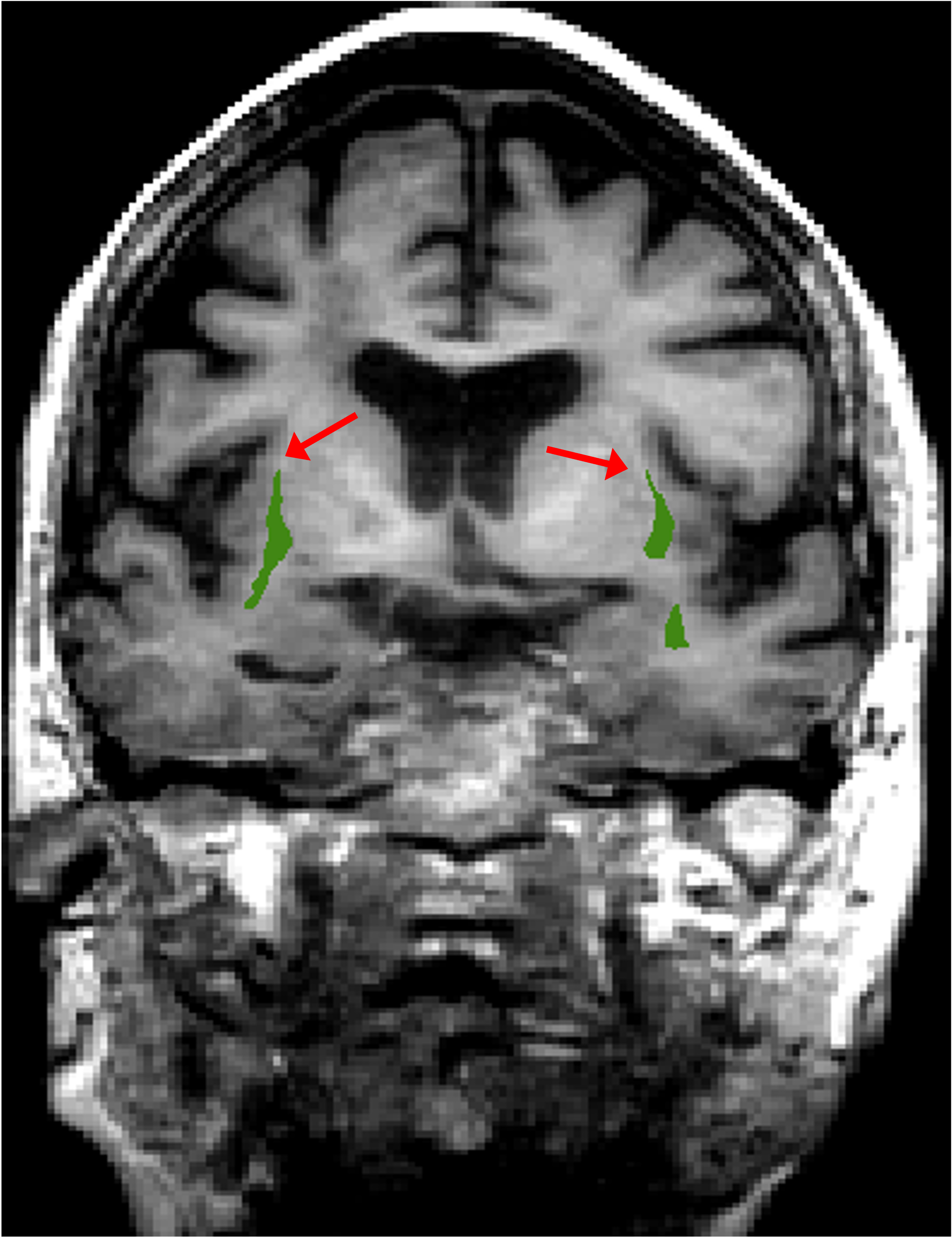}
      }\\
      
      \includegraphics[align=c,trim={4cm 7.4cm 3.8cm 7cm}, clip=true,width=0.23\linewidth]{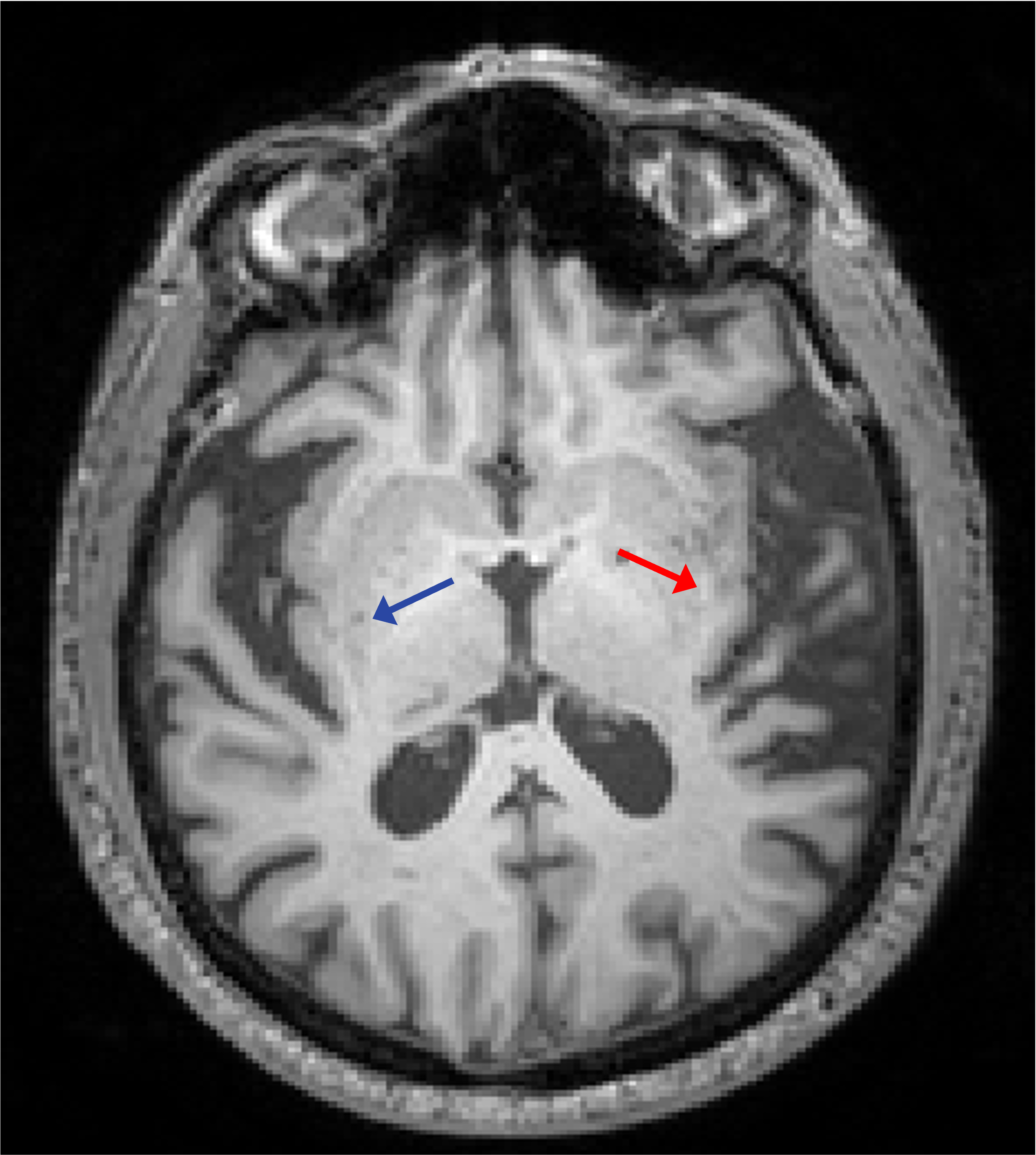} &
      {\setlength{\fboxrule}{2pt}
     \hspace{-5mm}
      \includegraphics[align=c,trim={4cm 7.4cm 3.8cm 7cm}, clip=true,width=0.23\linewidth]
    {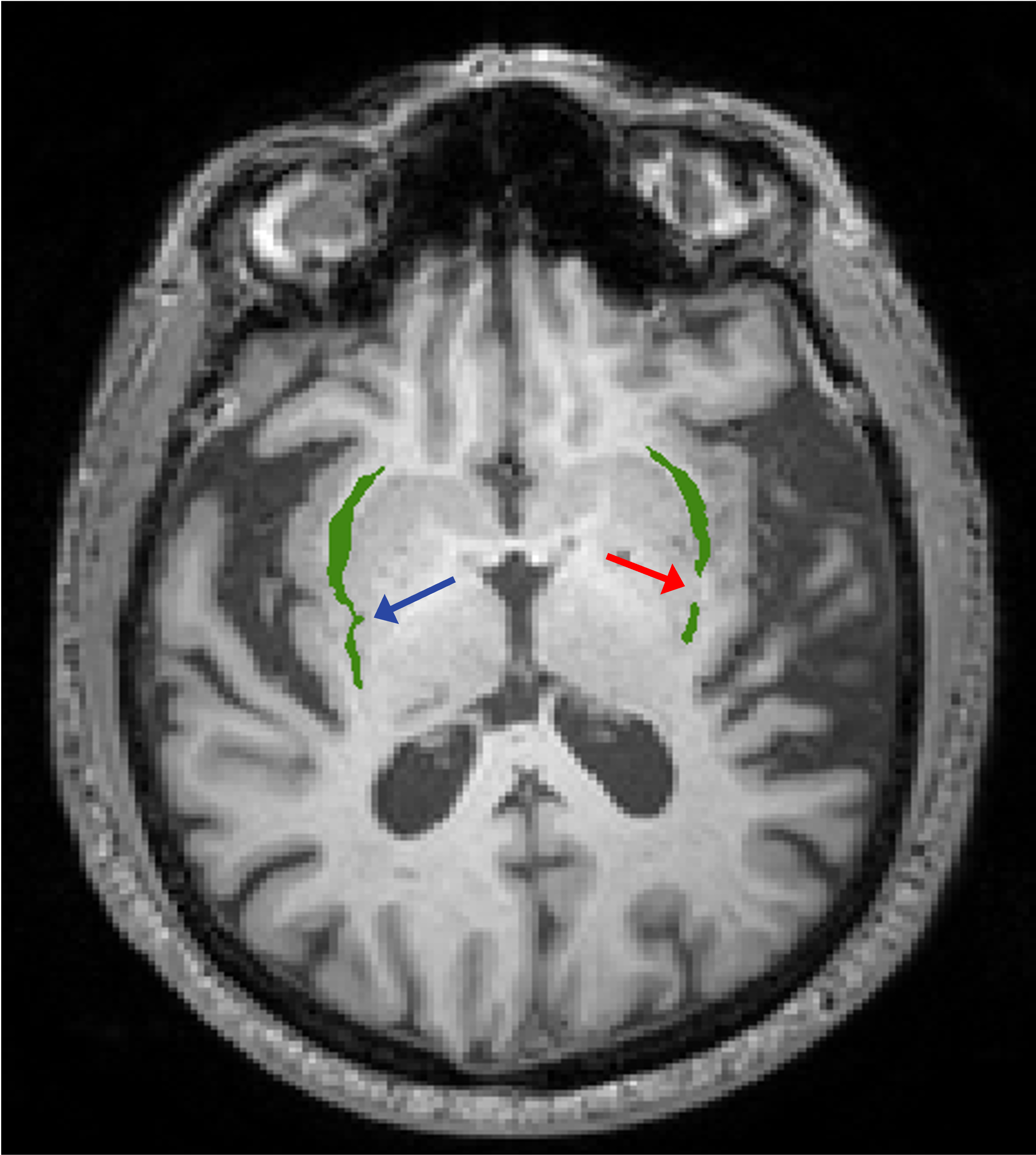}
      } &
      \includegraphics[align=c,trim={4.8cm 10.6cm 4.8cm 5.8cm}, clip=true,width=0.239\linewidth]{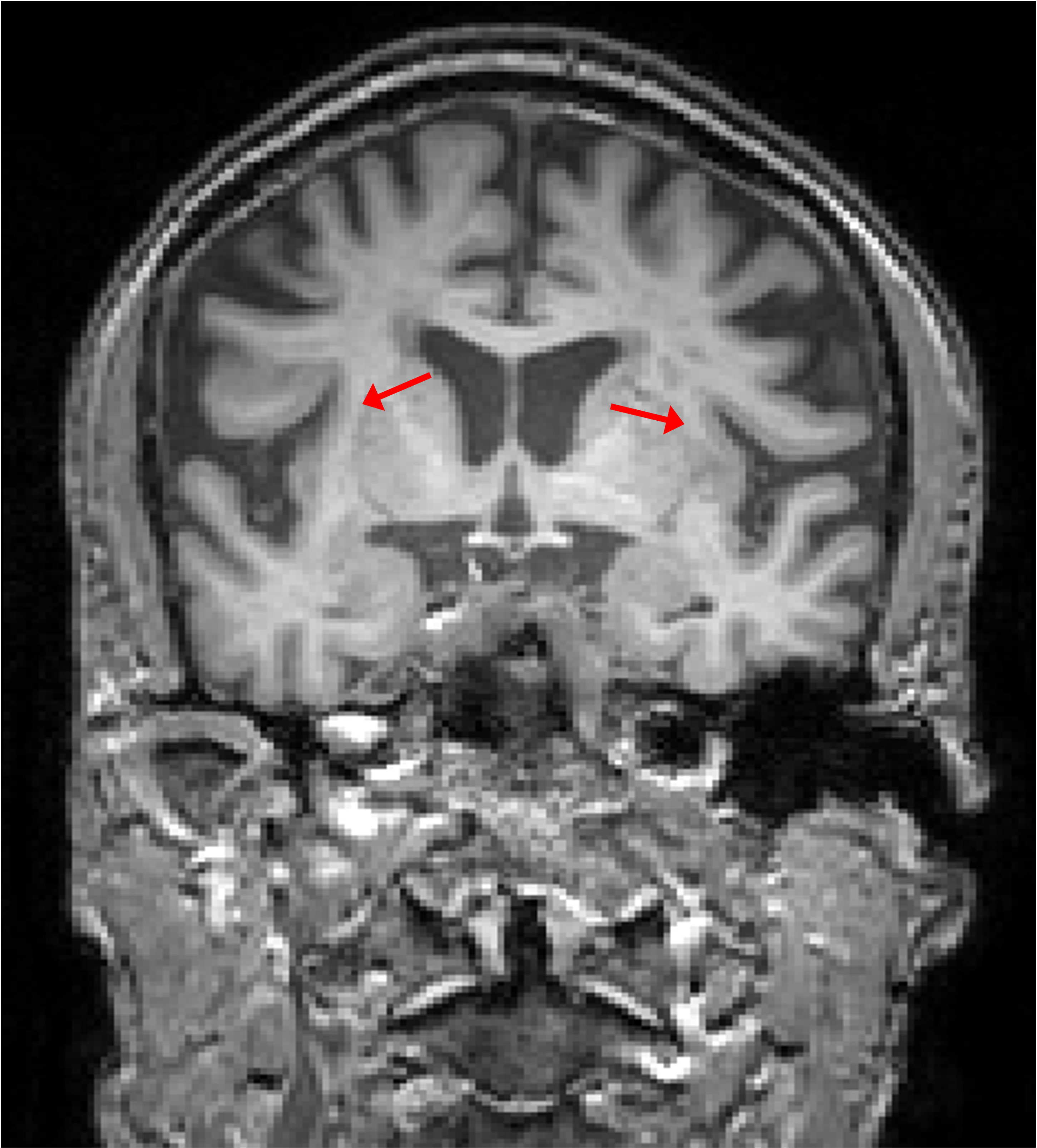} &
      {\setlength{\fboxrule}{2pt}
     \hspace{-5mm}
      \includegraphics[align=c,trim={4.8cm 10.6cm 4.8cm 5.8cm}, clip=true,width=0.239\linewidth]
    {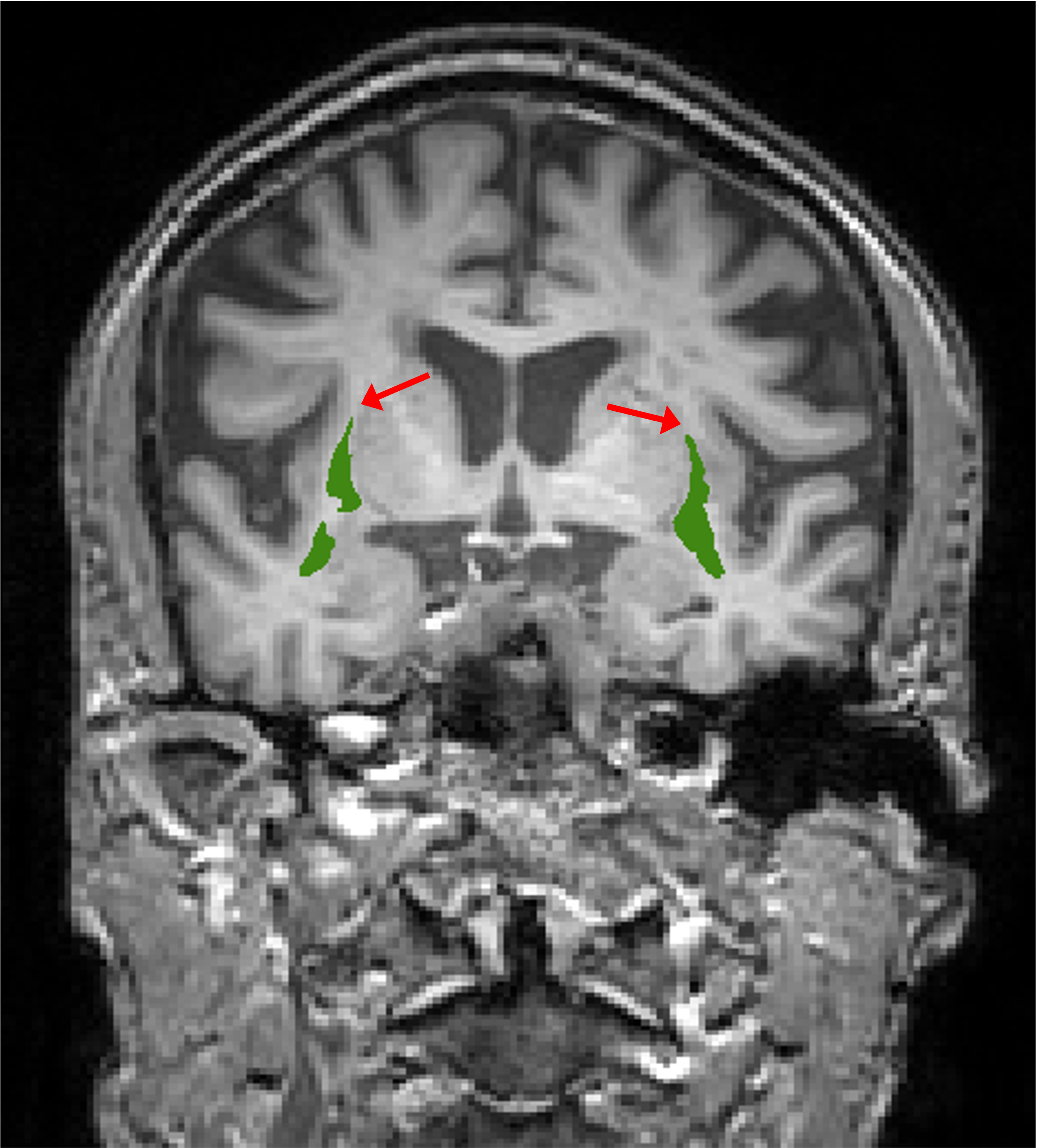}
      }
      \end{tabular}
      \caption{Axial (left) and coronal (right) views of the subjects in the IXI dataset with the two lowest QC scores. Blue arrows indicate regions where the claustrum segmentation overlaps with the putamen, while red arrows mark areas of incomplete segmentation.
      First row: 
      subject ID: 464, slices number: 158 (axial), 91 (coronal). 
      QC score averaged across hemispheres = 0.362. 
      Second raw: 
    subject ID: 257, slices number: 157 (axial), 128 (coronal).
      QC score averaged across hemispheres
      = 0.415.
      }
     \end{subfigure}\vfill
         \vspace{5mm}
      \begin{subfigure}[b]{\textwidth}
      \begin{tabular}{cccc}
      \vspace{3mm}
 \includegraphics[align=c,trim={2cm 5.7cm 2cm 3.5cm}, clip=true,width=0.23\linewidth]{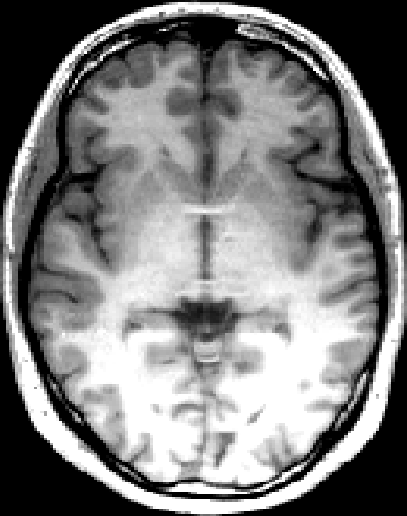} &
      {\setlength{\fboxrule}{2pt}
     \hspace{-5mm}
      \includegraphics[align=c,trim={2cm 5.7cm 2cm 3.5cm}, clip=true,width=0.23\linewidth]
    {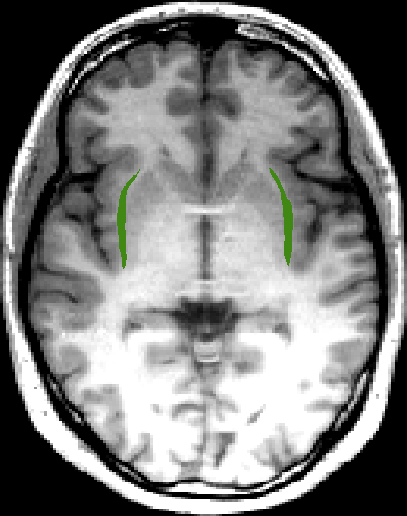}
      } &
    \includegraphics[align=c,trim={2cm 4.4cm 2cm 3.8cm}, clip=true,width=0.24\linewidth]
    {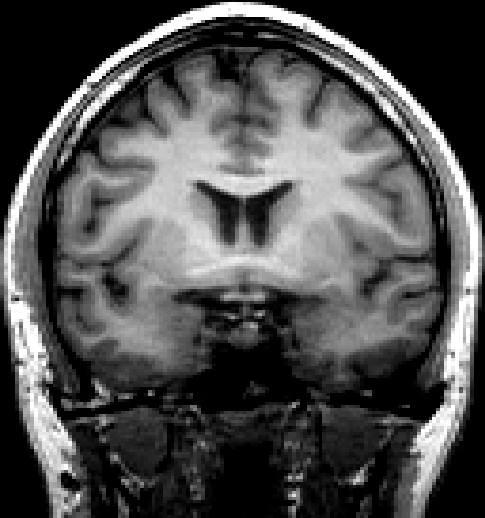} &
      {\setlength{\fboxrule}{2pt}
     \hspace{-5mm}
      \includegraphics[align=c,trim={2cm 4.4cm 2cm 3.8cm}, clip=true,width=0.24\linewidth]
   {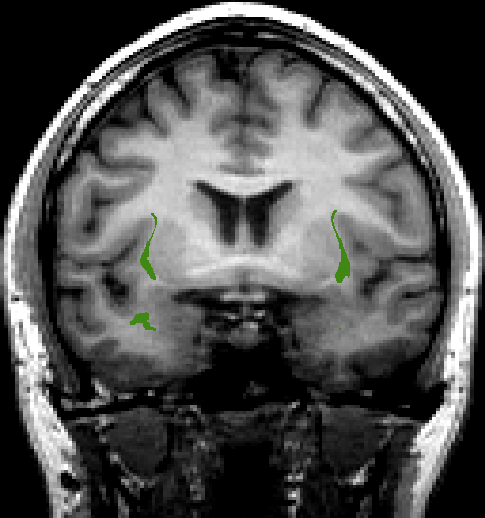}
      }\\
     \includegraphics[align=c,trim={2cm 5.7cm 2cm 3.5cm}, clip=true,width=0.23\linewidth]{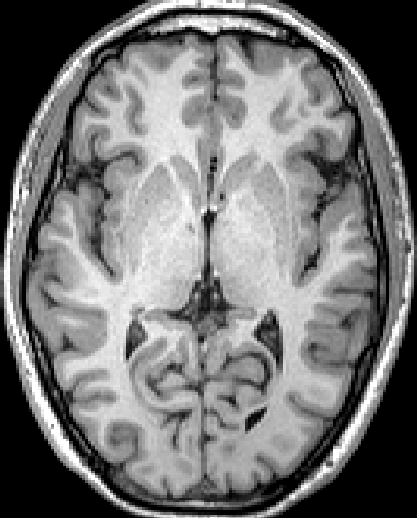} &
      {\setlength{\fboxrule}{2pt}
     \hspace{-5mm}
      \includegraphics[align=c,trim={2cm 5.7cm 2cm 3.5cm}, clip=true,width=0.23\linewidth]
    {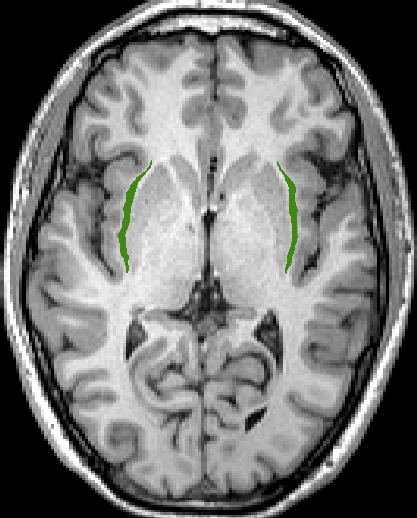}
      } &
    \includegraphics[align=c,trim={2cm 5.3cm 2cm 3.8cm}, clip=true,width=0.24\linewidth]
    {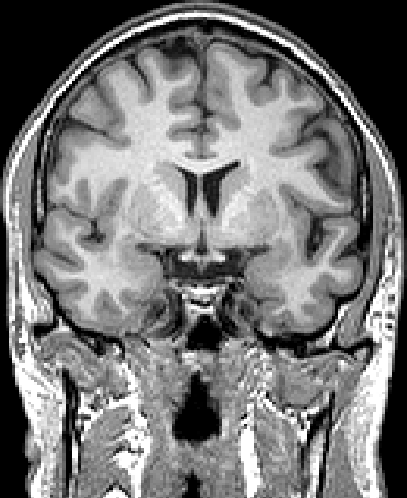} &
      {\setlength{\fboxrule}{2pt}
     \hspace{-5mm}
      \includegraphics[align=c,trim={2cm 5.3cm 2cm 3.8cm}, clip=true,width=0.24\linewidth]
     {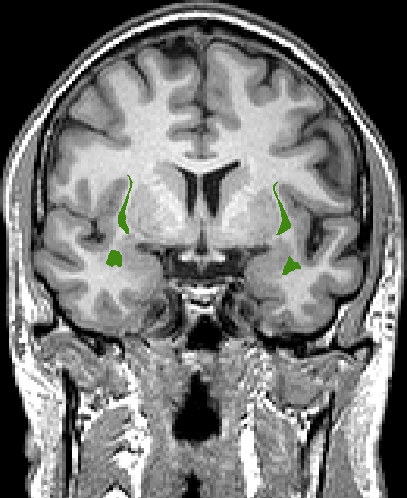}
      }\\
\end{tabular}
  \caption{Axial (left) and coronal (right) views of the subjects in the IXI dataset with the two highest QC scores.
  First row: subject ID: 303, slices number: 136 (axial), 98 (coronal). QC averaged across hemispheres = 0.638.
  Second row: subject ID: 179, slices number: 137 (axial), 108 (coronal). QC score averaged across hemispheres = 0.639.}
\end{subfigure}
\caption{
Claustrum segmentation on subjects from the IXI dataset with the two lowest and the two highest QC score (averaged across hemispheres). Slices were selected from similar locations as in Fig.~\ref{fig:manual_label} and Fig.~\ref{fig:labels} to facilitate a visual comparison.
}
\label{fig:IXI_subjects}
\end{figure*}

After applying this final segmentation method to all IXI images and visually inspecting the cases with the 50 lowest QC scores, and with the 20 smallest and 20 largest volumes, we concluded that the method showed some segmentations errors in around half of the inspected volumes, but it never severely failed. The most common error was to label some part of the putamen as claustrum, especially in its posterior part, or in presence of darker voxels within the putamen, as is shown in Fig.~\ref{fig:IXI_subjects}. These voxels are likely Vichow-Robins spaces or some kind of lesion. Since the set of examined subjects is expected to include the worst segmentations, we can conclude that the method never critically failed when applied to over 500 subjects from the IXI dataset. Fig.~\ref{fig:IXI_subjects} shows the segmentations obtained for the subjects with the two lowest and the two highest QC scores. We observe that the segmentations with the two lowest scores appear incomplete (red arrows) and overlapping with the putamen at times (blue arrows), especially in presence of darker voxels. On the other end, the cases with the two highest scores display segmentations that are anatomically accurate and complete, with the claustrum wrapping around the cortex superiorly (in the coronal views). A 3D view of claustrum segmentation on a IXI subject with QC score close to the dataset average is shown in \suppfigref{supp-sup_fig:IXI3D}.

\begin{figure}[t!]
\centering
\begin{tabular}{c}
      \includegraphics[trim={0cm 0 0cm 0}, clip=true,width=0.83\linewidth]{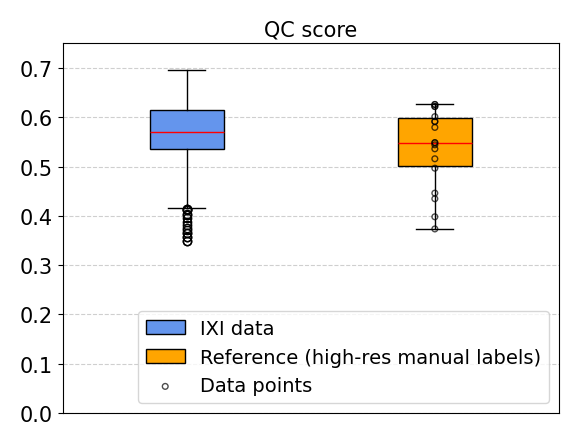}  \\
     \vspace{4mm}
     \hspace{-3mm}
      \includegraphics[trim={0cm 0 0cm 0}, clip=true,width=0.83\linewidth]
      {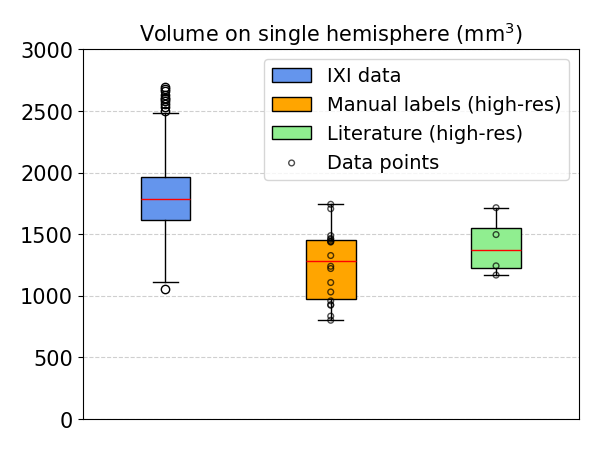}
\end{tabular}
\caption{
Top: Distribution of QC scores on the IXI subjects, computed as explained in Sec.~\ref{sec:epoch_selection}, together with the reference distribution for QC scores, computed on the 18 high-resolution manual labels. Bottom: Distribution of claustrum volumes on a single hemisphere on the automatic segmentations on IXI subjects, and in the 18 high-resolution manual labels. As a reference, we also display values reported in the literature from other studies using high resolution imaging or histology \citep{coates2024high,calarco2023establishing,kang2020comprehensive}. 
In all boxplots, the red line indicates the median, the box spans the first to third quartiles, and the whiskers extend to the farthest data point within 1.5 times the inter-quartile range. For clarity, IXI data points are displayed only when they fall outside the whiskers (i.e., as outliers).
}
\label{fig:IXI_dices_volumes}
\end{figure}

Additionally, Fig.~\ref{fig:IXI_dices_volumes} shows the distribution of the QC score and claustrum volumes obtained on the IXI subjects, together with the corresponding distributions computed on the manual labels. We obtained an average QC score of 0.570 $\pm$ 0.060 for the IXI dataset and 0.539 $\pm$ 0.078 for the reference distribution computed on the manual labels. The IXI QC metrics stratified by field strength were on average 0.570 $\pm$ 0.045 on 1.5T and 0.570 $\pm$ 0.047 on 3T scans 
(pvalue = 0.741 with linear regression adjusted for age).
Regarding claustrum volumes, we obtained a mean of 1,793.92 $\pm$ 259.16 mm$^3$ on the IXI subjects, which compares to 1,253.05 $\pm$ 283.79 mm$^3$ on the manual labels, and 1,405.80 $\pm$ 216.15 mm$^3$ from previous studies in the literature \citep{coates2024high,calarco2023establishing,kang2020comprehensive}.
Average IXI volumes stratified by field strength were 1,788.18 $\pm$ 252.44 mm$^3$ on 1.5T scans vs. 1,806.23 $\pm$ 239.80 mm$^3$ on 3T (pvalue = 0.593 with linear regression adjusted for age).

\begin{figure}[t!]
\centering
\setlength{\myWidth}{0.45\linewidth}
\setlength{\mySpace}{-3mm}
\begin{tabular}{cc}
    \hspace{\mySpace}
\begin{subfigure}{\myWidth}
      \includegraphics[trim={11cm 7.5cm 10cm 3cm}, clip=true,width=\linewidth]{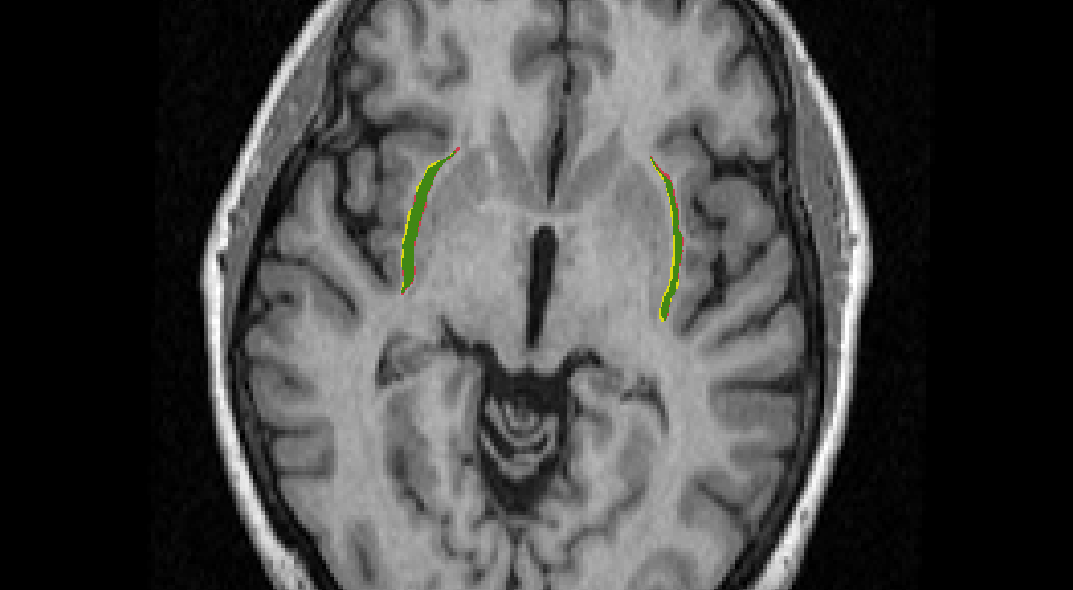}
      \caption{Axial view (slice 143), healthy subject (ID 199).}
      \end{subfigure} &
      \hspace{\mySpace}
     \begin{subfigure}{0.49\linewidth}
      \includegraphics[trim={14cm 8cm 12cm 6.2cm}, clip=true,width=\linewidth]
      {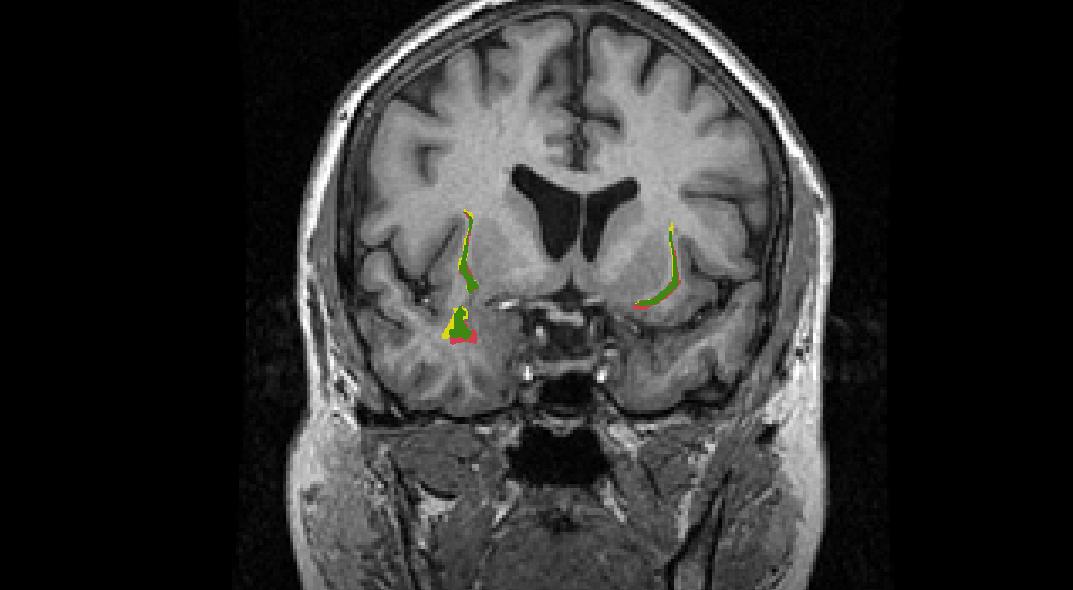}
       \caption{Coronal view (slice 88), healthy subject (ID 199).}
      \end{subfigure}
      \\
       \hspace{\mySpace}
      \begin{subfigure}{\myWidth}
      \includegraphics[trim={11cm 7.5cm 10cm 3cm}, clip=true,width=\linewidth]{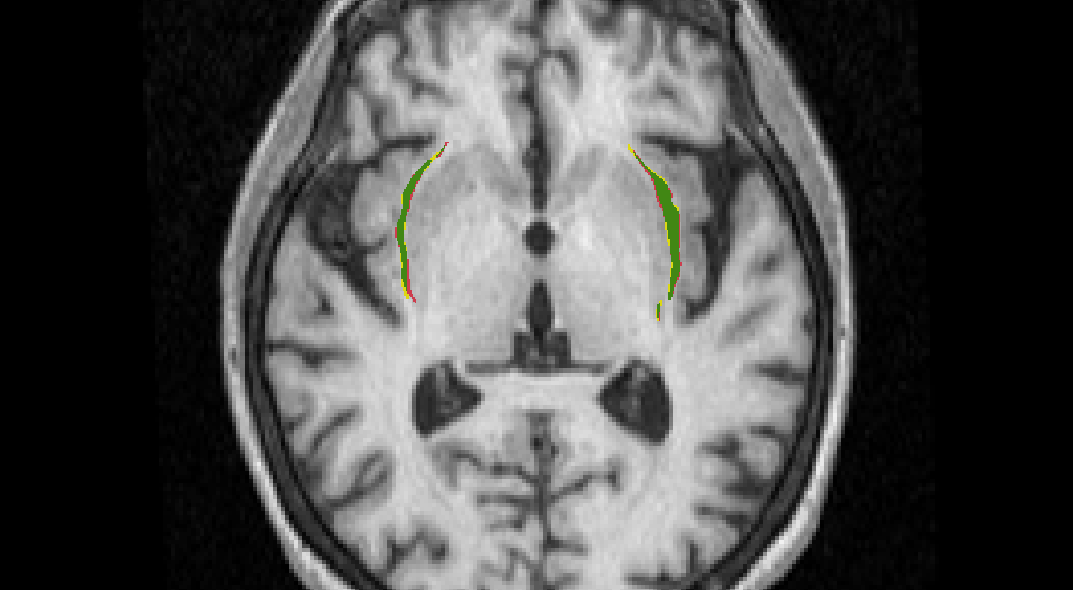}
       \caption{Axial view (slice 150), AD subject (ID 250).}
      \end{subfigure} &
     \hspace{\mySpace}
     \begin{subfigure}{0.495\linewidth}
      \includegraphics[trim={12cm 10cm 12cm 3cm}, clip=true,width=\linewidth]
      {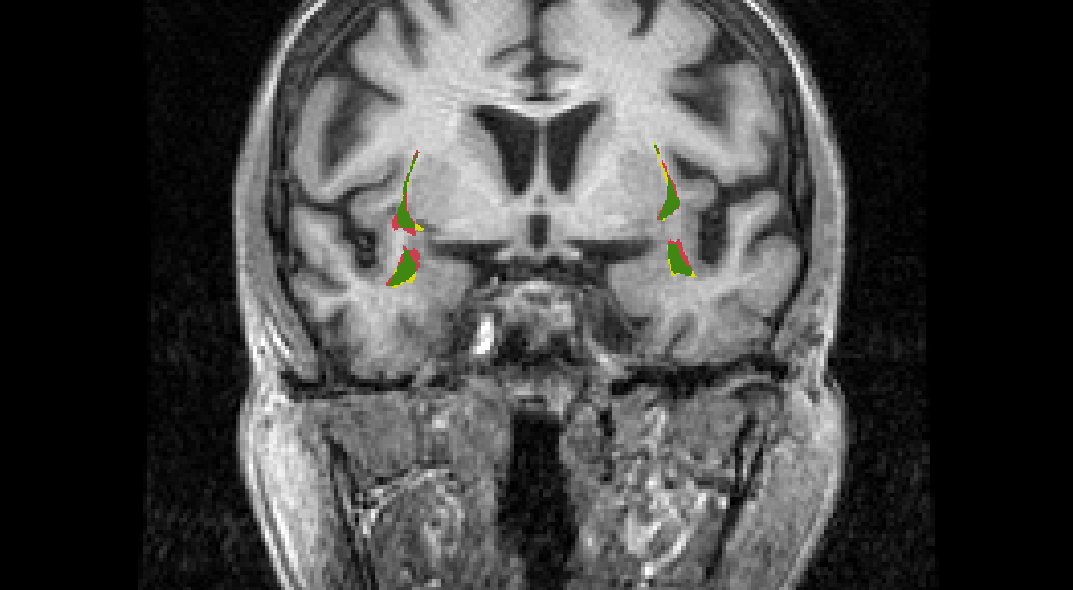}
      \caption{Coronal view (slice 79), AD subject (ID 250).}
      \end{subfigure}
\end{tabular}
\caption{
Claustrum segmentations obtained on repeated T1-weighted scans of two subjects from the Miriad dataset. In green we show the overlap between the two time points segmentations, in red the voxels in the first time point segmentation that are not included in the second one, and in yellow the voxels in the second time point segmentation that are not included in the first one. Top row: Healthy subject with Dice score between segmentations of 0.796 (averaged across hemispheres). Bottom row: AD subject with Dice score between segmentations of 0.778 (averaged across hemispheres). These subjects were chosen for their representative test-retest Dice scores.
}
\label{fig:miriad}
\end{figure}

On the Miriad test-retest dataset, we discarded 3 subjects for which the first two sessions scans were unavailable, and other 5 subjects where the rigid registration was inaccurate (for a spatial scaling factor between the two images), obtaining 40 AD subjects and 21 healthy controls. On these remaining subjects, we obtained an average Dice score between claustrum segmentations computed on two subsequent scans of 0.781 $\pm$ 0.068 (0.774 $\pm$ 0.068 on AD subjects and 0.796 $\pm$ 0.065 on healthy controls), denoting a good robustness of the method. The average claustrum volume for healthy controls and AD subjects was 1,757.74 $\pm$ 188.67 mm$^3$ and 1,593.54 $\pm$ 318.88 mm$^3$, respectively 
(one-sided pvalue = 0.0158 with linear regression adjusted for age). Fig.~\ref{fig:miriad} shows an example of the agreement between the segmentations on repeated scans for one AD subject and one healthy control. Repeated segmentations show good overlap, with most differences occurring at the boundaries and in the fragmented ventral region. A 3D rendering of claustrum segmentations on repeated scans for a healthy subject is displayed in \suppfigref{supp-sup_fig:Miriad3D}.

Table~\ref{tab:fsm} shows the average Dice score between segmentations obtained on different modalities of the same subject in the FSM dataset. We observe that the highest average Dice is obtained between quantitative T1 and T1-weighted (0.809 $\pm$ 0.044) and lowest obtained between Proton Density and T1-weighted scans (0.696 $\pm$ 0.085). Fig.~\ref{fig:fsm} shows images from different modalities of one subject, together with the overlap between the segmentation obtained on these modalities and the one computed on the T1-weighted scan. We note that there is a good agreement, with a few differences at the boundaries.
A 3D rendering of claustrum segmentations from T1-weighted and T2-weighted scans of a FSM subject is also shown in \suppfigref{supp-sup_fig:fsm3D}.

{
\renewcommand\arraystretch{1.3}
\addtolength{\tabcolsep}{+0.4mm}
\begin{table}[t!]
\tiny
\begin{center}
\resizebox{1\linewidth}{!}{%
\begin{tabular}{|c|c|c|}
\hline
T2w vs T1w (n=31) &  PD vs T1w (n=36) & qT1 vs T1w (n=36)  \\  \hline
0.720 $\pm$ 0.083
& 0.696 $\pm$ 0.085
&  0.809 $\pm$ 0.044 \\ \hline
\end{tabular}
}
\caption{
Mean and standard deviation of Dice score between claustrum segmentations computed on T1-weighted (T1w) scans and other modalities (T2-weighted (T2w), Proton Density (PD) and quantitative T1 (qT1)) of the same subject, across all subjects in the FSM dataset. In brackets, the number of subjects that are used to compute the average Dice score.
}
\label{tab:fsm}
\end{center}
\end{table}
\begin{figure}[t!]
\centering

\begin{tabular}{ccc}
 \hspace{-2 mm}
  \textbf{T1w} 
     \vspace{4 mm}
      \hspace{-5 mm}
      &
\begin{tabular}{c}
      \includegraphics[align=c,trim={3.5cm 6cm 3.5cm 5cm}, clip=true,width=0.42\linewidth]{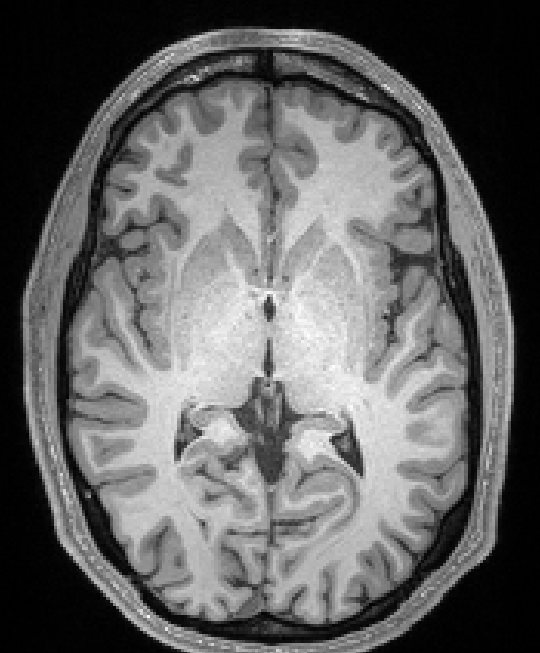} \end{tabular} &
      \hspace{-11mm}
      \begin{tabular}{c}
       \includegraphics[align=c,trim={3.5cm 6cm 3.5cm 5cm}, clip=true,width=0.42\linewidth]{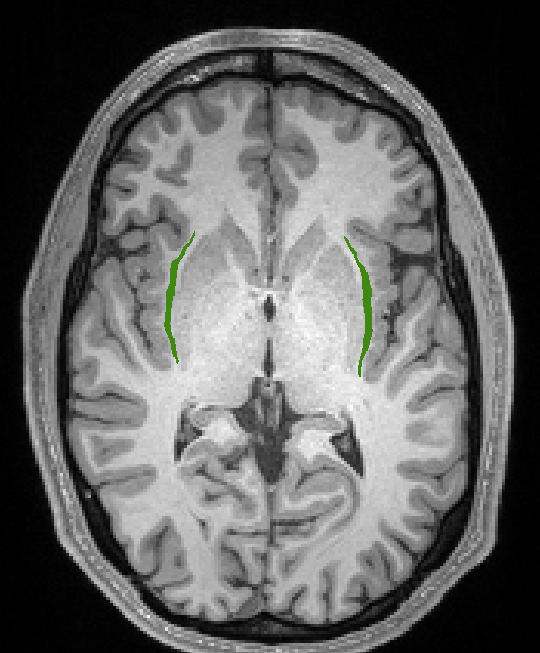}
       \end{tabular} \\
       \hspace{-2 mm}
       \textbf{T2w}
       \vspace{4 mm}
        \hspace{-5 mm} 
        &
       \begin{tabular}{c}
      \includegraphics[align=c,trim={3.5cm 6cm 3.5cm 5cm}, clip=true,width=0.42\linewidth]
      {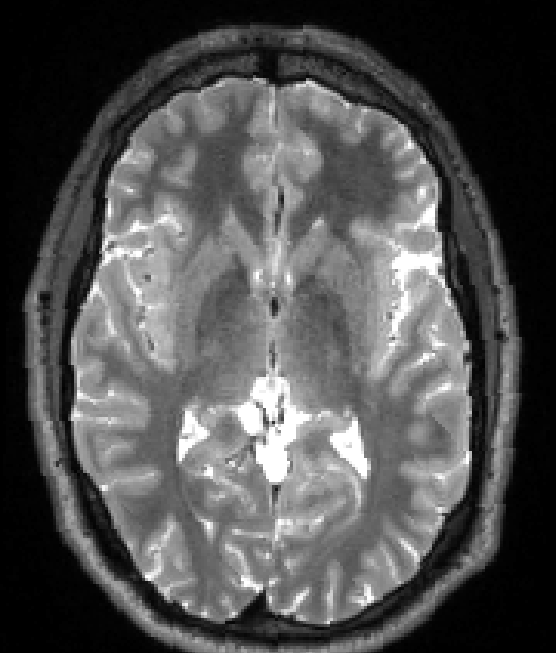}
      \end{tabular} & 
       \hspace{-11mm}
      \begin{tabular}{c}
        {\setlength{\fboxrule}{2pt}
      \includegraphics[align=c,trim={3.5cm 6cm 3.5cm 5cm}, clip=true,width=0.42\linewidth]
      {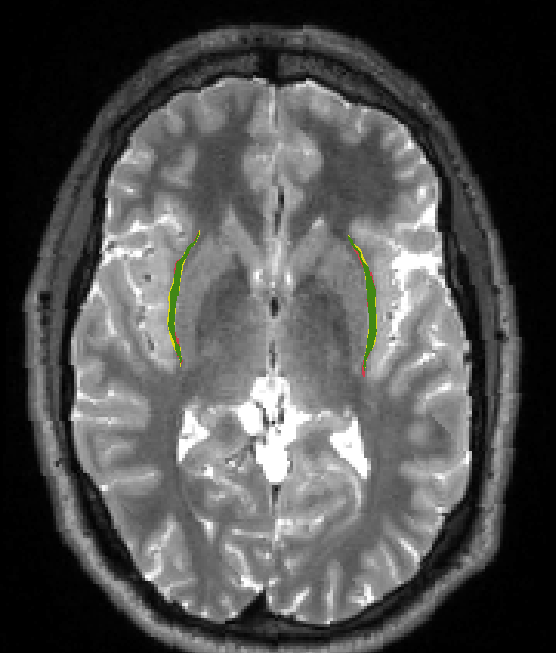}
      } 
      \end{tabular} \\
      \hspace{-2 mm}
      \textbf{PD}
         \vspace{4 mm}
          \hspace{-5 mm}
     &
      \begin{tabular}{c}
      \includegraphics[align=c,trim={3.5cm 6cm 3.5cm 5cm}, clip=true,width=0.42\linewidth]{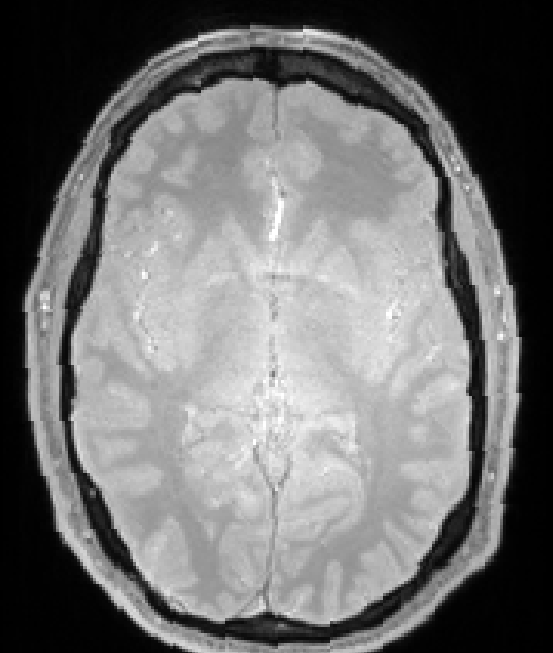} 
      \end{tabular} &
       \hspace{-11mm}
      \begin{tabular}{c}
    \includegraphics[align=c,trim={3.5cm 6cm 3.5cm 5cm}, clip=true,width=0.42\linewidth]{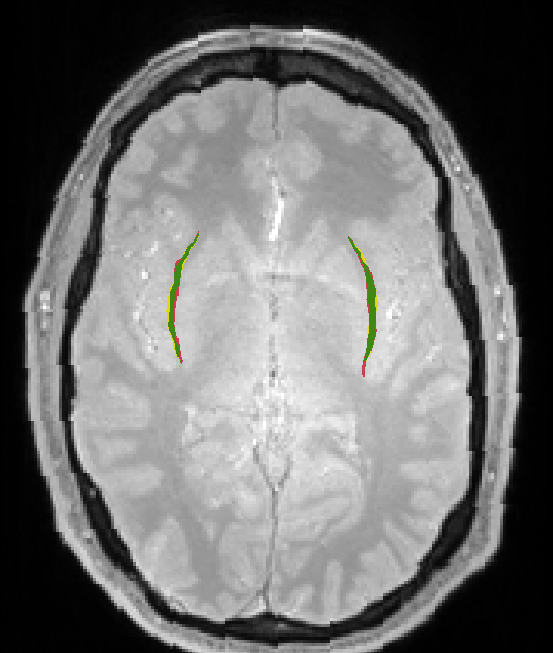}
    \end{tabular} \\
    \hspace{-2 mm}
     \textbf{qT1}
      \vspace{4 mm}
          \hspace{-5 mm}
          &
    \begin{tabular}{c}
      \includegraphics[align=c,trim={3.5cm 6cm 3.5cm 5cm}, clip=true,width=0.42\linewidth]
      {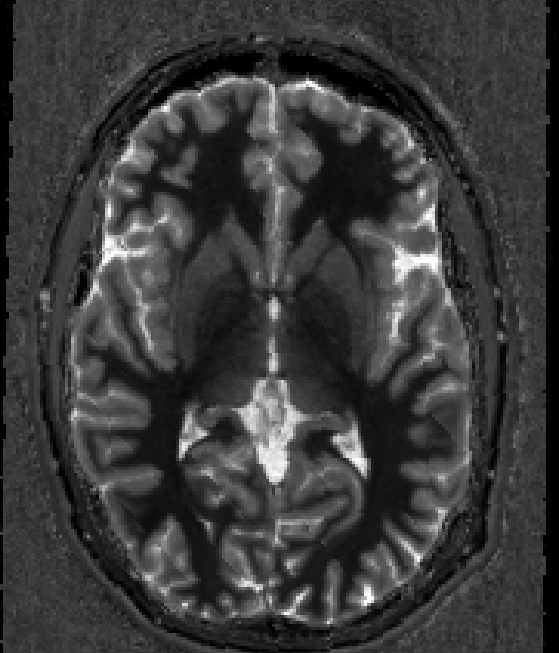}
      \end{tabular}
      &
       \hspace{-11mm}
       \begin{tabular}{c}
      \includegraphics[align=c,trim={3.5cm 6cm 3.5cm 5cm}, clip=true,width=0.42\linewidth]
      {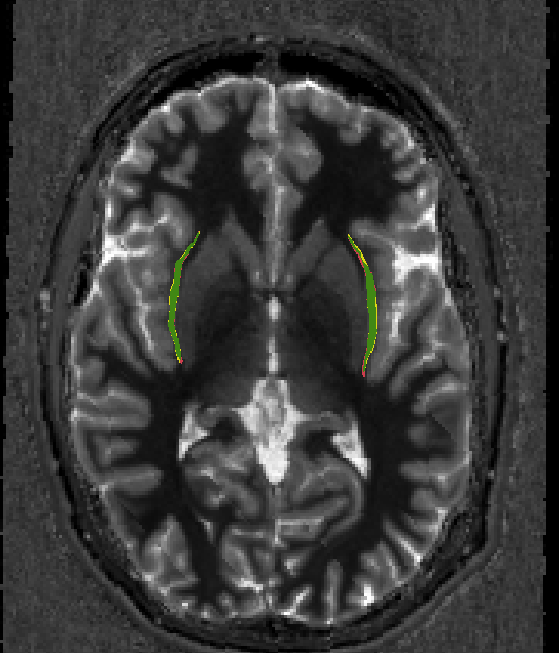}
      \end{tabular}
      \\
\end{tabular}
\caption{
Left: Axial view of different modalities of the same subject from the FSM dataset (subject ID: 011). Right: Overlap between the segmentation obtained on the given modality and the one computed on the T1w scan. In green, we show the overlap between the two segmentations, in red the voxels that are in the T1w segmentation and that are not included in the other modality one, and in yellow the voxels in the other modality segmentation that are not included in the T1w one. 
To facilitate comparison, we show the same axial slice (slice 100) across all modalities. For consistency, the slice was chosen from a similar location as in previous axial-view figures.
For this subject, Dice scores (vs. T1w segmentation, averaged across hemispheres) are 0.800 for T2w, 
0.730 for PD 
and 0.818 for qT1.
This subject was selected for its representative segmentation quality.
}
\label{fig:fsm}
\end{figure}

Fig.~\ref{fig:fsm_it} shows the Dice score between claustrum segmentations computed on synthesized images with varying TI and the one obtained on TI=1,000, averaged across subjects of the FSM dataset. As the TI increases, the Dice score initially decreases on average, reaching its minimum at TI=380 with an average value of 0.676 $\pm$ 0.094, and then it increases towards a complete overlap. When averaged across all TI and subjects, the method obtains a Dice score of 0.873 $\pm$ 0.109.
\begin{figure}[t!]
\centering
\includegraphics[trim={0cm 0 0cm 0}, clip=true,width=0.95\linewidth]{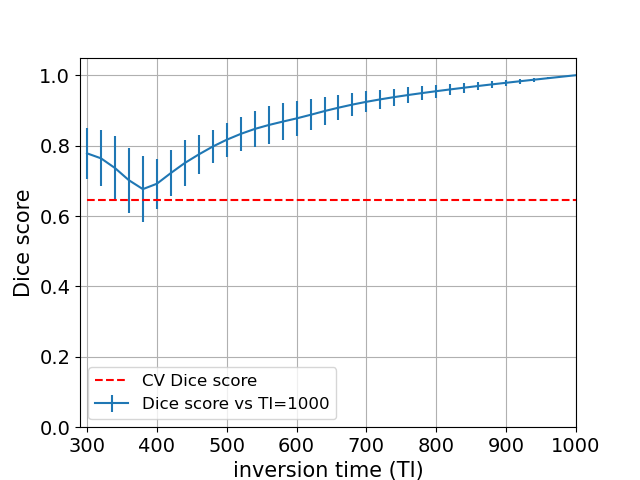} 
\caption{
 Dice score between claustrum segmentations computed on synthesized images with different TI and the one obtained on TI=1,000, on the FSM dataset. The full line shows the average Dice across all the subjects, and the bars extend to one standard deviation away from the average. The Dice score obtained on average using CV on the high resolution cases is also shown as reference (0.632, see Table~\ref{tab:CV_results}).
}
\label{fig:fsm_it}
\end{figure}
\begin{figure}[t!]
\centering
\begin{tabular}{ccc}
 \hspace{-2 mm}
  \textbf{TI 300} 
     \vspace{4 mm}
      \hspace{-5 mm}
      &
\begin{tabular}{c}
      \includegraphics[align=c,trim={9.5cm 7cm 9.5cm 5cm}, clip=true,width=0.38\linewidth]{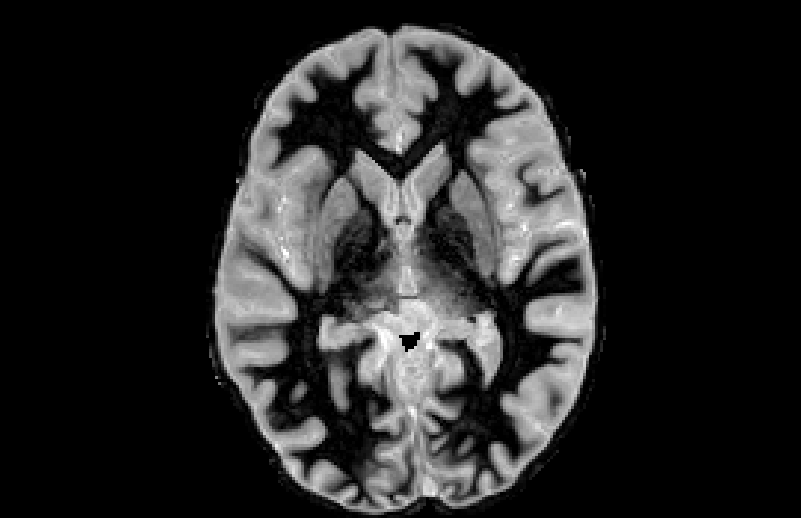} \end{tabular} &
      \hspace{-11mm}
      \begin{tabular}{c}
       \includegraphics[align=c,trim={9.5cm 7cm 9.5cm 5cm}, clip=true,width=0.38\linewidth]{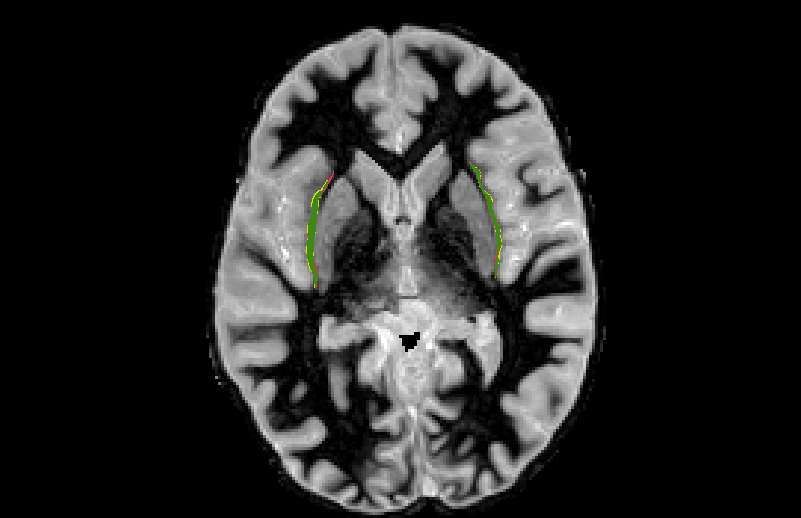}
       \end{tabular} \\
       \hspace{-2 mm}
       \textbf{TI 400}
       \vspace{4 mm}
        \hspace{-5 mm} 
        &
       \begin{tabular}{c}
      \includegraphics[align=c,trim={9.5cm 7cm 9.5cm 5cm}, clip=true,width=0.38\linewidth]
      {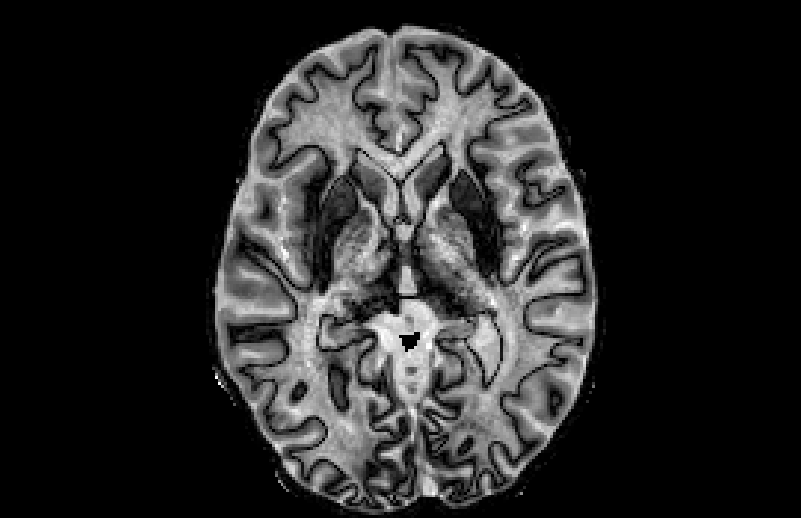}
      \end{tabular} & 
       \hspace{-11mm}
      \begin{tabular}{c}
        {\setlength{\fboxrule}{2pt}
      \includegraphics[align=c,trim={9.5cm 7cm 9.5cm 5cm}, clip=true,width=0.38\linewidth]
      {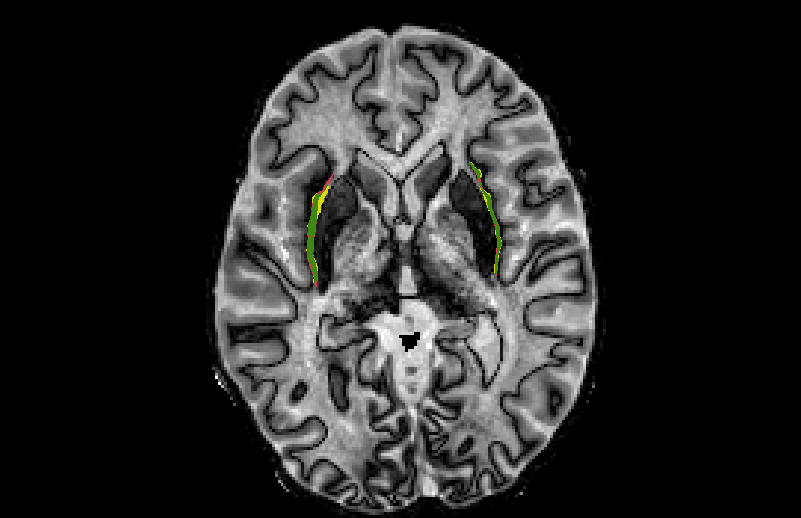}
      } 
      \end{tabular} \\
      \hspace{-2 mm}
      \textbf{TI 500}
         \vspace{4 mm}
          \hspace{-5 mm}
     &
      \begin{tabular}{c}
      \includegraphics[align=c,trim={9.5cm 7cm 9.5cm 5cm}, clip=true,width=0.38\linewidth]
      {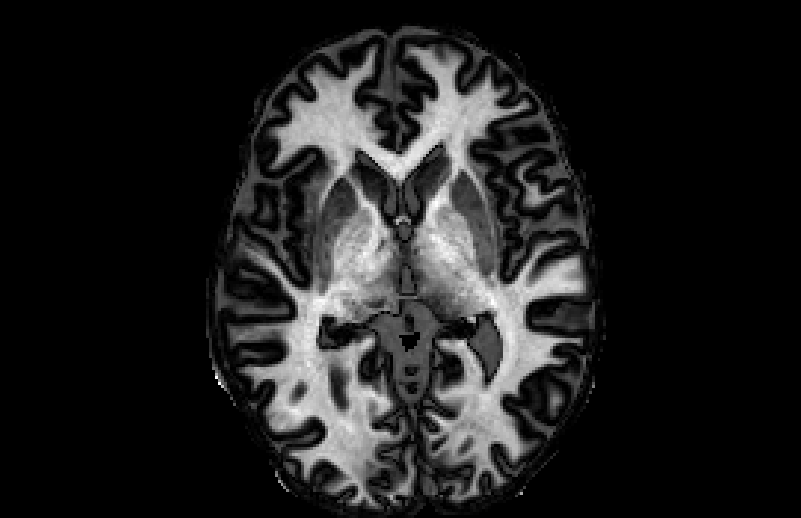}
      \end{tabular} &
       \hspace{-11mm}
      \begin{tabular}{c}
     \includegraphics[align=c,trim={9.5cm 7cm 9.5cm 5cm}, clip=true,width=0.38\linewidth]
      {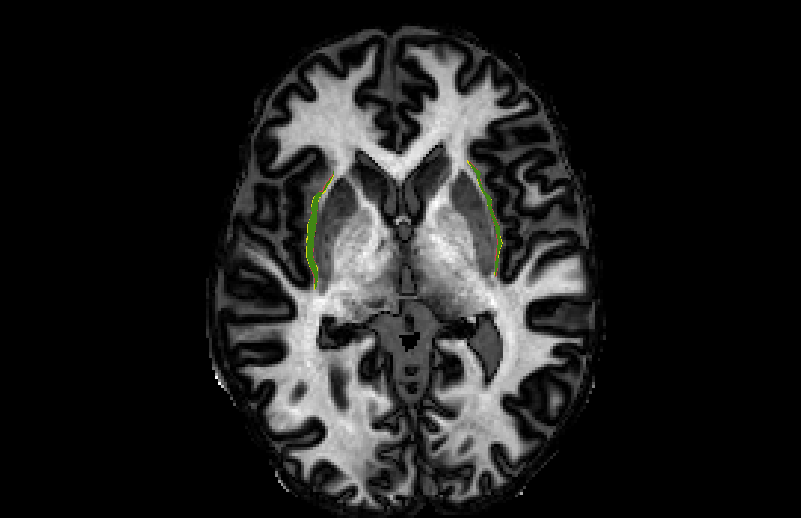}
    \end{tabular} \\
    \hspace{-2 mm}
     \textbf{TI 600}
      \vspace{4 mm}
          \hspace{-5 mm}
          &
    \begin{tabular}{c}
      \includegraphics[align=c,trim={9.5cm 7cm 9.5cm 5cm}, clip=true,width=0.38\linewidth]
      {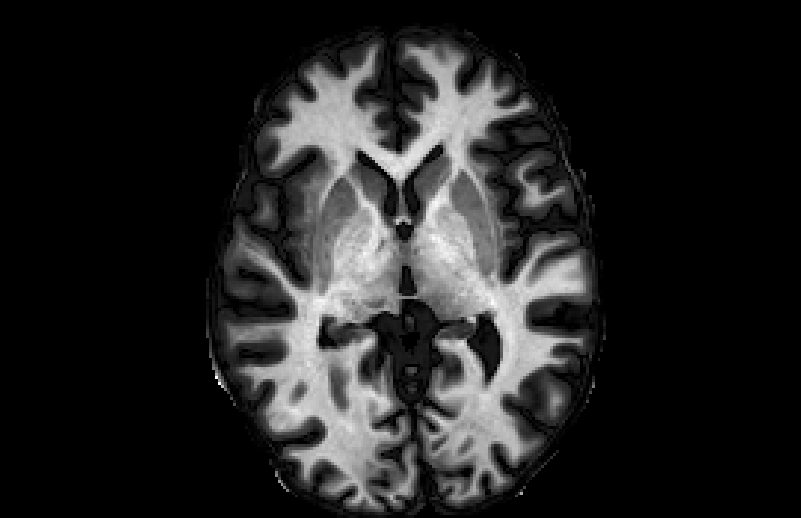}
      \end{tabular}
      &
       \hspace{-11mm}
       \begin{tabular}{c}
      \includegraphics[align=c,trim={9.5cm 7cm 9.5cm 5cm}, clip=true,width=0.38\linewidth]
      {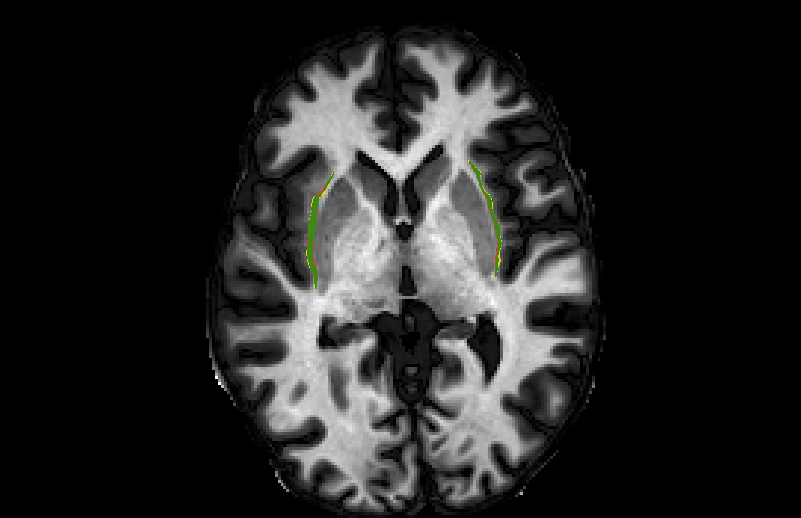}
      \end{tabular}
      \\
       \hspace{-2 mm}
     \textbf{TI 800}
      \vspace{4 mm}
          \hspace{-5 mm}
          &
    \begin{tabular}{c}
      \includegraphics[align=c,trim={9.5cm 7cm 9.5cm 5cm}, clip=true,width=0.38\linewidth]
      {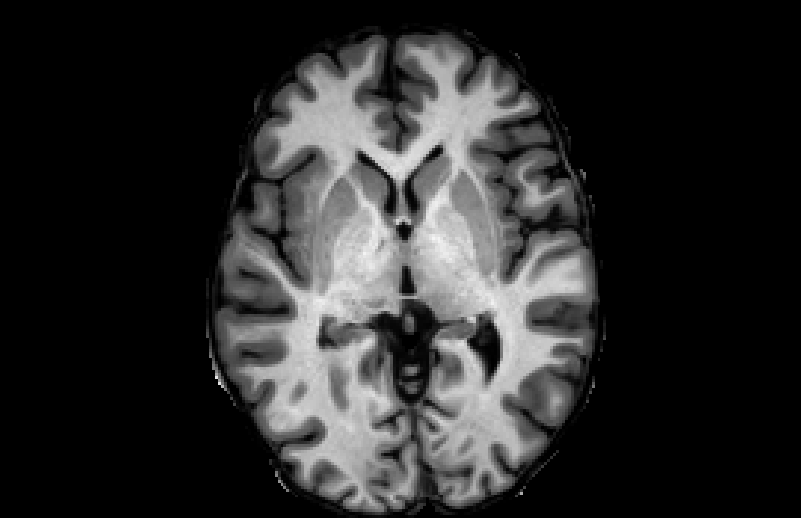}
      \end{tabular}
      &
       \hspace{-11mm}
       \begin{tabular}{c}
      \includegraphics[align=c,trim={9.5cm 7cm 9.5cm 5cm}, clip=true,width=0.38\linewidth]
      {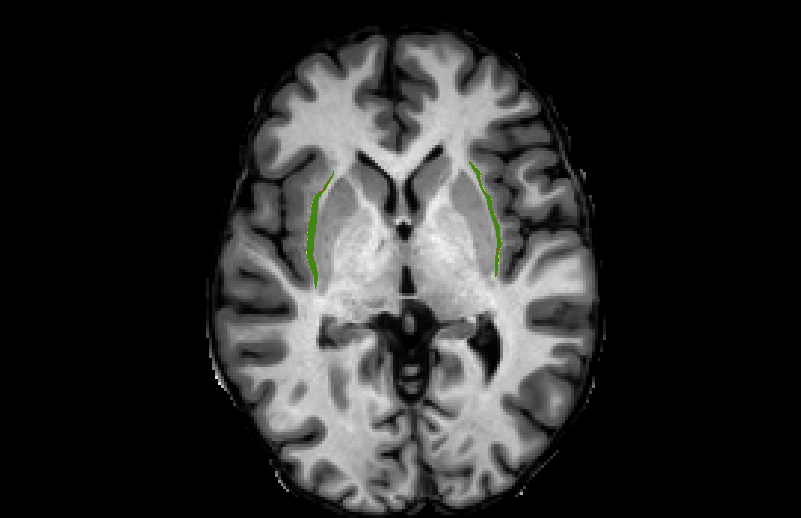}
      \end{tabular}
      \\
       \hspace{-2 mm}
     \textbf{TI 1,000}
      \vspace{4 mm}
          \hspace{-5 mm}
          &
    \begin{tabular}{c}
      \includegraphics[align=c,trim={9.5cm 7cm 9.5cm 5cm}, clip=true,width=0.38\linewidth]
      {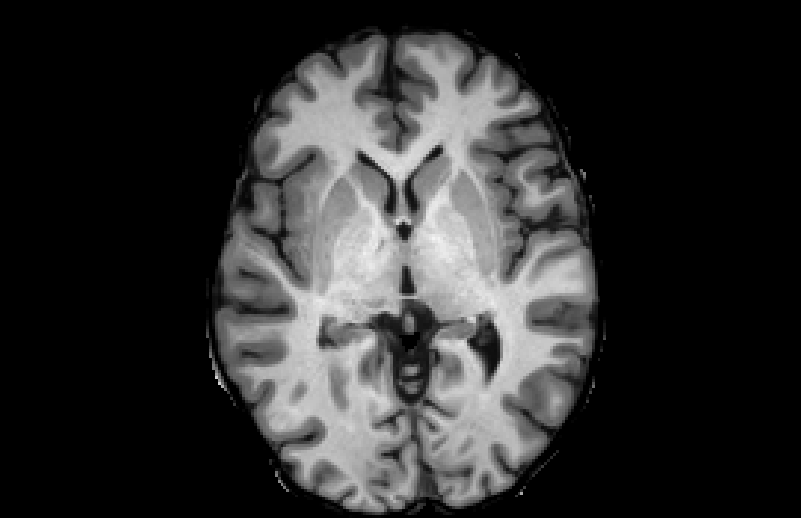}
      \end{tabular}
      &
       \hspace{-11mm}
       \begin{tabular}{c}
      \includegraphics[align=c,trim={9.5cm 7cm 9.5cm 5cm}, clip=true,width=0.38\linewidth]
      {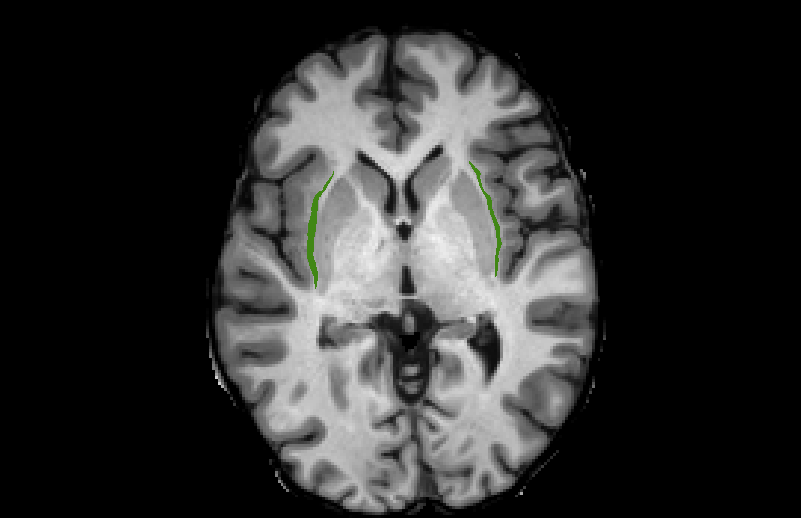}
      \end{tabular}
      \\
\end{tabular}
\caption{
Left: Axial view of synthetic images of one subject from the FSM dataset, obtained with different TI (subject ID: 010, slice: 99). Right: Overlap between the segmentation obtained on the image with the given TI and the one computed on TI=1,000. In green, we show the overlap between the two segmentations, in red the voxels that are in the segmentation on TI = 1,000 and not in the other TI one, and in yellow the vice versa.}
\label{fig:fsm010_it}
\end{figure}
Fig.~\ref{fig:fsm010_it} shows synthetic images with varying TI for one subject in the dataset, together with the agreement between the segmentations obtained on those images and the one computed on TI=1,000. We observe that the contrast changes rapidly from TI=300 to TI=500, and then it varies more smoothly, in line with the Dice changing shown in Fig.~\ref{fig:fsm_it}. In general the subject displays a very good overlap between the segmentations, with the most differences obtained at TI=400, in line with the lowest Dice score achieved on average at TI=380 in Fig.~\ref{fig:fsm_it}.

In all these experiments on \textit{in vivo} data, it took around 3 minutes to apply the method on a subject using basic configurations (only the affine registration to MNI152 space to determine the FoV) using a machine with 16 CPUs. The version with the nonlinear registration to MNI152 space, used to compute the QC score for quality assessment, took around 10 minutes per subject on the same machine.

\section{Discussion and conclusion}
\noindent
In this paper we proposed a method based on the SynthSeg framework for ultra-high resolution claustrum segmentation, which is robust against changes in contrast and resolution. To the best of our knowledge, this is the first method for automatic claustrum segmentation with these features. The method was designed to exploit ultra-high resolution \textit{ex vivo} images, which display details that are not visible in typical \textit{in vivo} scans, to learn to segment claustrum in all its components. Then, since it is contrast- and resolution-insensitive, the method can be applied to \textit{in vivo} images (which typically differ in contrast from \textit{ex vivo} scans), across a wide range of contrasts and resolutions.

We assessed performance of the method using CV on the 18 high-resolution scans, where we obtained a moderate Dice score (0.632 $\pm$ 0.061), which might be due to the challenging nature of claustrum segmentation. We also quantified CV performance in terms of other metrics, achieving for instance a mean surface distance of 0.458 $\pm$ 0.124 mm, a volumetric similarity of 0.867 $\pm$ 0.092, and an intersection over union of 0.465 $\pm$ 0.066. The latter penalizes segmentation errors more strongly than Dice score and therefore provides valuable insight for a thin, small structure like the claustrum.

Comparing our Dice score with performance of other available segmentation methods, we observe that \citet{berman2020automatic} achieved an average Dice score of 0.55, while \citet{brun2022automatic} and \citet{coates2024high} did not quantify their claustrum segmentation accuracy - \citet{coates2024high} reported that their own method can be used to \textit{approximate} the claustrum location in typical MRI scans. Only \citet{li2021automated} and \citet{albishri2022unet} achieved higher Dice scores (0.718 and 0.82  respectively). In this regard, it is worth noting that our results are obtained at much higher resolution, where it might be more challenging to get all the details and nuances of the claustrum. 
Additionally, a contrast-agnostic method is generally expected to perform worse than a supervised CNN that is tested on its own training domain, but it gains in ability to generalize. For instance, \citet{albishri2022unet}'s Dice score of 0.82 was assessed with 5-fold CV within one homogeneous dataset. When we applied their pre-trained method to our \textit{in vivo} labeled case, the predictions were poor, with a Dice score of 0.299. Besides the change of dataset (with different scanner and image acquisition protocol), differences in image resolution and labeling protocol may also account for the significant decrease in performance. 
A segmentation method that instead aims to generalize across contrasts and resolutions was recently proposed in \citet{casamitjana2024next}. 
 It performs segmentation of 333 ROIs across the whole brain - including the claustrum - using a probabilistic atlas derived from histology, integrated into a Bayesian framework. We tested this method on our high-resolution dataset (0.35 mm isotropic), achieving an average Dice score of 0.370 $\pm$ 0.112 against our manual labels. Several factors may contribute to this low Dice score, such as differences in labeling protocols, the method not being optimized for single-hemisphere segmentation, and the presence of a strong bias field, which is particularly challenging for Bayesian methods. Overall, these results highlight a difference in the intended application: Our method targets accurate claustrum segmentation, while their method appears to be more suitable for whole-brain application, where the claustrum is not the primary focus.

In the context of inter-rater variability, we observed that our CV Dice score was exceeded by inter-rater Dice scores, both before and after applying SmartInterpol: They ranged between 0.80 and 0.83, suggesting that the high inter-rater agreement is not merely an outcome of using SmartInterpol.
Other studies assessing inter-rater variability for claustrum segmentation reported Dice values between 0.67 and 0.83 \citep{berman2020automatic,coates2023high,coates2024high}, while \citet{li2021automated} computed an intra-rater variability Dice score of 0.667. Part of the gap between our CV and inter-rater Dice scores may stem from SynthSeg’s contrast-agnostic nature: While human raters are able to perform optimally on these high-resolution hemispheres with good contrast around the claustrum, the automatic method may trade performance on this specific domain for generalizability, which is however essential for its broader use.
We note that SynthSeg could be tuned on a specific domain, by adjusting intensity priors in the GMM. 
However, \citet{billot2023synthseg} showed that doing so consistently lowered performance, highlighting the benefits of augmenting data beyond realistic boundaries.

In addition to CV on the 18 high-resolution cases, we demonstrated that our method also works on scans at typical \textit{in vivo} resolutions ($\approx$1 mm isotropic).
When tested on the 18 hemispheres downsampled at different lower resolutions, the method exhibited excellent robustness, maintaining good performance up to approximately 1.5 mm isotropic. Additionally, it proved effective on \textit{in vivo} T1-weighted scans at typical resolutions from the IXI, Miriad, and FSM datasets, including subjects with AD. On the IXI data, we evaluated segmentation accuracy in absence of ground truth using the QC score and claustrum volume to identify the most suspicious cases for detailed visual inspection, and we concluded that the method did not severely fail on any subject. 
Furthermore, the QC scores on the IXI data were, on average, similar to the corresponding reference distribution computed on the high-resolution manual labels (0.570 $\pm$ 0.060 vs. 0.539 $\pm$ 0.078, respectively).
We remind that the QC score is not a direct measure of segmentation performance. This comparison rather aims to contextualize the QC scores for IXI data in relation to a reference distribution of satisfactory QC values, which reflects natural inter-subject variability of the claustrum.
We point out here that IXI is a challenging data set with a very wide subject age range, large voxel size, 1.5T and 3T, and two different MRI scanner manufactures. The success of the claustrum segmentation on 581 subjects from this dataset demonstrates its robustness. The lack of significant differences in results obtained on 1.5T and 3T scans further proves the method as robust against variations in contrast.

Regarding claustrum volumes, CV segmentations on high-resolution cases yielded values comparable to the manual annotations (1,415.35 $\pm$ 260.66 mm$^3$ vs. 1,253.05 $\pm$ 283.79 mm$^3$ respectively; $\approx$13\% difference), while segmentations on the IXI dataset resulted in substantially larger volumes (1,793.92 $\pm$ 259.16 vs. 1,253.05 $\pm$ 283.79 mm$^3$ respectively; $\approx$43\% difference). It is not clear why the IXI volumes are so much higher, but it is likely due to partial volume effects at 1 mm when trying to segment a structure sometimes only 0.5 mm wide. In contrast, the standard deviations of volume distributions for IXI-based segmentations and high-resolution cases (both manual and automatic) are comparable, suggesting that inter-subject variability in claustrum volume is preserved across different data types.
The volumes derived from our manual labels are also broadly consistent with those reported in other high-resolution claustrum studies (1,253.05 $\pm$ 283.79 mm$^3$ vs. 1,405 $\pm$ 216.15 mm$^3$ respectively; $\approx$12\% difference). However, cross-study comparisons should be interpreted with caution, as they are influenced by differences in resolution, imaging modalities, labeling protocols, and sample sizes (see Fig.~\ref{subfig:literature_volumes}): \citet{calarco2023establishing}, using histology at a resolution similar to our labels (0.1 mm isotropic), reported a comparable volume (1,243 mm$^3$), while \citet{kang2020comprehensive}'s measurement was higher (1,715 mm$^3$ on average per hemisphere), possibly due to partial volume effects at lower resolution (0.7 mm isotropic). \citet{coates2024high} labeled a case included in our study, allowing for direct volume comparison with our manual label: We found that the two volumes of the right claustrum were consistent after spatial alignment (1,468.9 mm$^3$ vs. 1,497 mm$^3$; 1.9\% difference), while we observed a larger discrepancy in the left hemisphere (1438.1 mm$^3$ vs. 1,168.2 mm$^3$; 18.8\% difference). This discrepancy may arise from left-right inconsistencies that we observed in their labeling, where the right ventral claustrum appeared more extensively labeled than the left, potentially contributing to their significant volume asymmetry.

The robustness of the method was also evaluated in a test-retest setting (average Dice score between two repeated scans of 0.781 $\pm$ 0.068) and across different contrasts. It showed a good generalization ability to different modalities, with the lowest overlap obtained between Proton Density and T1-weighted scans (0.696 $\pm$ 0.085), probably because of the poor contrast around claustrum in Proton Density images. While the method is in fact robust towards changes in contrast, performance still depends on the ability of a contrast to resolve the claustrum.
When using synthetic images to simulate different MPRAGES over a range of TIs, the method showed very good robustness as well (average Dice score across all TIs and subjects of 0.873 $\pm$ 0.109), with the most differences observed on average at TI=380 (Dice score of 0.676 $\pm$ 0.094). In practice, the impact of the performance drop at TI = 380 ms may be irrelevant as the typical TI range encountered in 3T MPRAGES is narrower, usually between 800 and 1000 ms. The broader TI range in our experiment served as a stress test for the contrast-insensitive performance of the method, since it produces extreme contrast variations (see Fig.~\ref{fig:fsm010_it}). Remarkably, despite these extreme changes, the method maintained strong performance.

We emphasize that contrast- and resolution-insensitivity of the method is contingent on adequate signal quality: For reliable segmentation, images should have a voxel size of at least 1–1.5 mm and good contrast around the claustrum. If poor resolution, contrast, or image quality renders the claustrum indistinguishable, the segmentation will be largely guessed, requiring cautious interpretation.
Also, for optimal generalizability, model selection should ideally use a diverse validation set with different contrasts and resolutions. In the absence of such a dataset, we adopted different criteria: For CV, we maximized the Dice score on labeled images (primarily \textit{ex vivo}), while for the final model, we trained on all available labeled data and optimized the average QC score on a subset of \textit{in vivo} T1-weighted IXI scans. In the latter case, T1-weighted scans were selected as they represent the most practical and commonly used modality for claustrum segmentation. Although this non-heterogeneous validation set may bias the model toward its contrast and resolution, our results still demonstrate strong generalization across diverse imaging conditions. 
This broad range of tested scenarios also enables practical recommendations for designing \textit{in vivo} claustrum segmentation studies. T1-weighted images seem to provide the best contrast, while resolution should balance accuracy and acquisition time, depending on the specific application: Whereas our results show good performance up to 1.4 mm voxel size, higher resolutions inevitably capture finer structural details. As mentioned earlier, images should be high-quality to yield reliable outcomes.

Beyond the need for high image quality, the proposed method is subject to other limitations. In terms of training data, the high-resolution \textit{ex vivo} images were crucial for capturing fine claustrum details and developing a high-resolution segmentation model. However, the small sample size and high average age of the dataset (61.94 $\pm$ 13.66 years) may limit generalizability and introduce age-related bias. These constraints - typical of \textit{ex vivo} datasets - are partially mitigated through SynthSeg extensive label augmentation and synthetic intensities generation, which alter morphological features well beyond typical age effects and expose the model to a high degree of variability. SynthSeg has in fact been shown to require fewer training examples than supervised CNNs to achieve optimal performance \citep{billot2023synthseg}. Our robust results on larger, diverse datasets - spanning different ages, scanner types, and contrasts - seem to be in agreement with this finding, suggesting that these data limitations are (at least partially) addressed by the method (see also \suppsecref{supp-sup_sec:volume_age_ixi} for insights into the method's ability to detect age effects in the IXI dataset).

Although we focused on claustrum segmentation in the paper, for context the model was trained to segment all structures within the ROI. Training labels for other structures were generated by the SynthSeg whole-brain framework, which is suboptimal for high-resolution hemispheres due to its 1 mm resolution segmentations and whole-brain design. However, these training labels were sufficient for our objective and provided valuable contextual information for claustrum segmentation. Future research could focus on refining these labels, potentially leading to improved automatic segmentation of these regions within the ROI.
Similarly, SynthMorph was designed for whole-brain image registration and may not work optimally on high-resolution \textit{ex vivo} hemispheres. However, visual inspection confirmed accurate alignment of the manual labels to MNI space, before proceeding with the construction of the probabilistic atlas and QC analysis. Both SynthSeg and SynthMorph were chosen for their contrast-independent properties.

After using whole-brain SynthSeg to generate the training labels, the data were cropped with the claustrum centered in the FoV — a common but suboptimal practice in image segmentation. Although this can introduce shortcut learning, which prevents detection of claustrum at image borders \citep{lin2024shortcut}, our consistent FoV centering around claustrum during testing ensured it did not hamper our results. 
We also mirrored all training labels to the right hemisphere and trained the model exclusively on right-sided data, processing left hemispheres at test time by reversing the input ROI. This approach ensures the model remained agnostic to the hemisphere, avoiding the introduction of laterality bias. This is crucial to enable future unbiased assessment of hemispheric asymmetries in the claustrum, which have been suggested by prior research \citep{kapakin2011claustrum}. Finally, despite the extensive hyperparameter tuning that we conducted to optimize the model for claustrum segmentation, CV performance on high-resolution data was still moderate. Future work could explore alternative loss functions, such as distance transform or topology-preserving losses \citep{kirchhoff2024skeleton}, to potentially improve segmentation accuracy for a thin structure like the claustrum.

Besides these considerations, the proposed method was able to address shortcomings of existing methods for claustrum segmentation, such as segmenting only the dorsal part of claustrum \citep{berman2020automatic} achieving inaccurate or unquantified results \citep{berman2020automatic,brun2022automatic,coates2024high}, and being difficult to generalize to unseen datasets \citep{albishri2022unet,li2021automated}. Thanks to the use of synthetic images during training - with augmented labels and random contrast intensities - it was able to successfully segment dorsal and ventral portions of claustrum in T1-weighted scans across multiple datasets, as well as multimodal images, yielding exceptional generalization. Given the scarcity of neuroimaging tools for the study of claustrum (including methods for its automatic segmentation), the shortcomings of existing methods, and the variety of unresolved questions about the structure and its functions, we believe that the proposed method will be a valuable resource for further investigations of the claustrum. It may also be useful as a \textit{post hoc} correction tool for automatic segmentations of surrounding regions, such as addressing putamen overlabeling errors \citep{perlaki2017comparison,dewey2010reliability}.

\section*{Acknowledgment}
\noindent
Much of the computation resources required for this research was performed on computational hardware generously provided by the Massachusetts Life Sciences Center (https://www.masslifesciences.com/). We would also like to acknowledge support from NIH grants R01NS105820, R01EB023281, R01NS083534, R01AG057672, and R01NS112161, and grants 1RF1MH123195, 1R01AG070988, 1UM1MH130981, and 1RF1AG080371.
This project was also supported by the Innovative Health Initiative Joint Undertaking (JU) under grant agreement No 101112153. The JU receives support from the European Union’s Horizon Europe research and innovation programme and COCIR, EFPIA, EuropaBio, MedTech Europe, Vaccines Europe, AB Science SA and Icometrix NV.

\section*{Data availability}
\noindent
\textbf{\textit{In vivo} datasets}:
\begin{list}{-}{\leftmargin=0.7em \itemindent=1em}
    \item The IXI dataset is publicly available at the link \url{http://brain-development.org/ixi-dataset/}
    \item The Miriad dataset is publicly available at the link \url{https://www.ucl.ac.uk/drc/research-clinical-trials/minimal-interval-resonance-imaging-alzheimers-disease-miriad}
    \item The FSM dataset is publicly available in the OpenNeuro platform at \url{https://doi.org/10.18112/openneuro.ds004958.v1.0.0}
    \item High-resolution case 16 \citep{lusebrink2017t1} is publicly available in the Dryad Digital Repository at \url{http://dx.doi.org/10.5061/dryad.38s74}
\end{list}
\textbf{\textit{Ex vivo} datasets}:
\begin{list}{-}{\leftmargin=0.7em \itemindent=1em}
\item Case 3 \citep{costantini2023cellular} is publicly available on the DANDI platform at \url{https://dandiarchive.org/dandiset/000026}
    \item Case 13 \citep{edlow20197} is publicly available on the OpenNeuro platform at \url{https://doi.org/10.18112/openneuro.ds002179.v1.1.0} and in the Dryad Digital Repository at \url{https://doi.org/10.5061/dryad.119f80q}.
    \item All other \textit{ex vivo} hemispheres are not open access.
\end{list}
\textbf{Claustrum segmentations}:
Our high-resolution manual labels in native space, the probabilistic atlas in MNI152 space, and the automatic segmentations on \textit{in vivo} data can be downloaded at \url{https://surfer.nmr.mgh.harvard.edu/pub/data/claustrumData.tar.gz}.

\appendix
\section{Test metrics}\label{sec:test_metrics}
\noindent
To assess CV performances of the segmentation method, we used the following metrics:
\begin{list}{-}{\leftmargin=1em \itemindent=0em}
\item Dice score: it is a measure of volumetric overlap between two structures A and B, defined as $\frac{2v(A \cap B)}{v(A)+v(B)}$ where $v(\cdot)$ represents the volume of a structure.
\item Intersection over union: it is a measure of volumetric overlap between two structures A and B, defined as $\frac{v(A \cap B)}{v(A \cup B)}$ where $v(\cdot)$ represents the volume of a structure.
\item True positive rate: it is defined as TP / (TP + FN), where TP = number of true positives, FN = number of false negatives, FP = number of false positives.
\item False discovery rate: it is defined as FP /(FP + TP), where TP = number of true positives, FN = number of false negatives, FP = number of false positives.
\item Robust Hausdorff distance: we used Hausdorff distance \citep{huttenlocher1993comparing} with 95$\%$ percentile.
\item Mean surface distance: mean surface distance between two surfaces $S$ and $S'$ is defined as $\frac{1}{n_S+n_{S'}}\left(\sum_{p\in S} d(p,S')+\sum_{p'\in S'} d(p',S)\right)$ where $n_S, n_{S'}$ are the number of points on surfaces $S$ and $S'$ respectively, and for a given point $p\in S$ we define the distance from $S'$ as $d(p,S')=min_{p' \in S'}\|p-p'\|_2$.
\item Volumetric similarity: volumetric similarity between two structures A and B is defined as $1-\frac{|v(A)-v(B)|}{v(A)+v(B)}$ where $v(\cdot)$ represents the volume of a structure.
\end{list}

\section{Increasing the robustness of the method by adding voxel-wise noise}\label{sec:appendix}
\noindent
We compared the model with and without adding random voxel-wise Gaussian noise in the synthetic intensity images during training. The noise is added as part of the intensity image augmentation right after the bias field simulation, in the following way: We sample a standard deviation $\sigma$ from a uniform distribution $\sigma \in \mathcal{U}(0,100)$, then sample independent voxel-wise Gaussian noise from $\mathcal{N}(0,\sigma^2)$, and add it to the intensity image with probability $95\%$.

For the case with noise, we used the model trained on the first fold of the CV described in Sec.~\ref{sec:training}, following the same procedure for selecting the best epoch (i.e., maximizing the Dice score on the validation set of that fold). We then applied the resulting model to all IXI subjects and computed the QC score as detailed in Sec.~\ref{sec:epoch_selection}. We then trained a model on the same CV fold \textit{without} adding noise to the synthetic images, and repeated the same procedure. We obtained an average QC score on the IXI dataset of 0.555 $\pm$ 0.050 for the model with noise, and 0.490 $\pm$ 0.094 for the one without noise, which compare to the average QC metric on the manual labels of 0.539 $\pm$ 0.078, showing that the model with noise achieves more robust results.

\bibliography{bibliography}

\end{document}